\documentclass[11pt]{article}

\usepackage[preprint]{acl}

\usepackage{times}
\usepackage[T1]{fontenc}
\usepackage[utf8]{inputenc}
\usepackage{microtype}
\usepackage{inconsolata}
\usepackage{graphicx}

\usepackage{makecell}
\usepackage{booktabs}
\usepackage{multirow}

\usepackage{bm}
\usepackage{amsmath}
\usepackage{amssymb}
\DeclareMathOperator*{\argmax}{arg\,max}
\DeclareMathOperator*{\mean}{mean}

\usepackage{cleveref}
\Crefname{section}{\S}{\S\S}

\title{
Signals Are Present, Predictions Fail:\\
Probing Diagram Representations in Large Vision-Language Models
}

\author{
    \textbf{Haruto Yoshida\textsuperscript{1}},
    \textbf{Keito Kudo\textsuperscript{1}},
    \textbf{Yoichi Aoki\textsuperscript{1}},
    \textbf{Ryota Tanaka\textsuperscript{2}},
\\
    \textbf{Itsumi Saito\textsuperscript{1}},
    \textbf{Keisuke Sakaguchi\textsuperscript{1}},
    \textbf{Kentaro Inui\textsuperscript{1}}
\\
\\
    \textsuperscript{1}Tohoku University,
    \textsuperscript{2}Human Informatics Labs., NTT, Inc.
\\
    \texttt{\{yoshida.haruto.p1, keito.kudo.q4, youichi.aoki.p2\}@dc.tohoku.ac.jp, } \\
    \texttt{ryota.tanaka@ntt.com, }
    \texttt{\{itsumi.saito, keisuke.sakaguchi\}@tohoku.ac.jp, } \\
    \texttt{kentaro.inui@mbzuai.ac.ae} \\
}

\begin{document}
\maketitle
\begin{abstract}
Large vision-language models (LVLMs) demonstrate strong performance on diagram understanding benchmarks, yet they still struggle to understand relationships between elements, particularly those represented by directed edges between nodes.
To investigate the causes of this limitation, we probe the internal representations of LVLMs using a synthetic diagram dataset based on directed graphs.
Our linear probing experiments reveal that signals about nodes and edges are linearly decodable from the hidden states of LVLMs, even for aspects that LVLMs struggle with in VQA, such as edge direction and connectivity between nodes.
These results reveal a gap between internal signals and model outputs.
We further perform causal interventions on the model's hidden states, showing that, for some aspects, the signals identified through probing causally contribute to the model's final predictions.
These findings also hold on flowcharts, not only on synthetic diagrams.
\end{abstract}

\section{Introduction}
Diagrams convey complex information through visual layouts that people can grasp quickly, making them an important communication medium across various domains, including academic research~\cite{Hsu2021-xf, Kohn2006-oe} and business~\cite{Jost2017-xl, Kernbach2010-ju}.
Accordingly, various systems for diagram understanding have been studied~\cite{Kembhavi2016-eh, Li2025-uk}.

Large vision-language models (LVLMs)~\cite{Bai2025-wu, Bai2025-zl, Liu2024-xj, Team2025-fs} have shown strong performance on diagram understanding benchmarks~\cite{Zhang2025-bh, Lu2024-hj}.
However, prior studies report that LVLMs still struggle to understand certain visual elements in diagrams, such as relational information indicated by arrows and lines~\cite{Hou2025-mf, Zhu2025-mw, Pan2024-ls}.
To better understand the causes of these limitations, it is important to examine how LVLMs internally represent diagram information, yet how they represent basic visual elements such as nodes, edges, and global structure remains underexplored.

\begin{figure}[t]
    \centering
    \includegraphics[width=\linewidth]{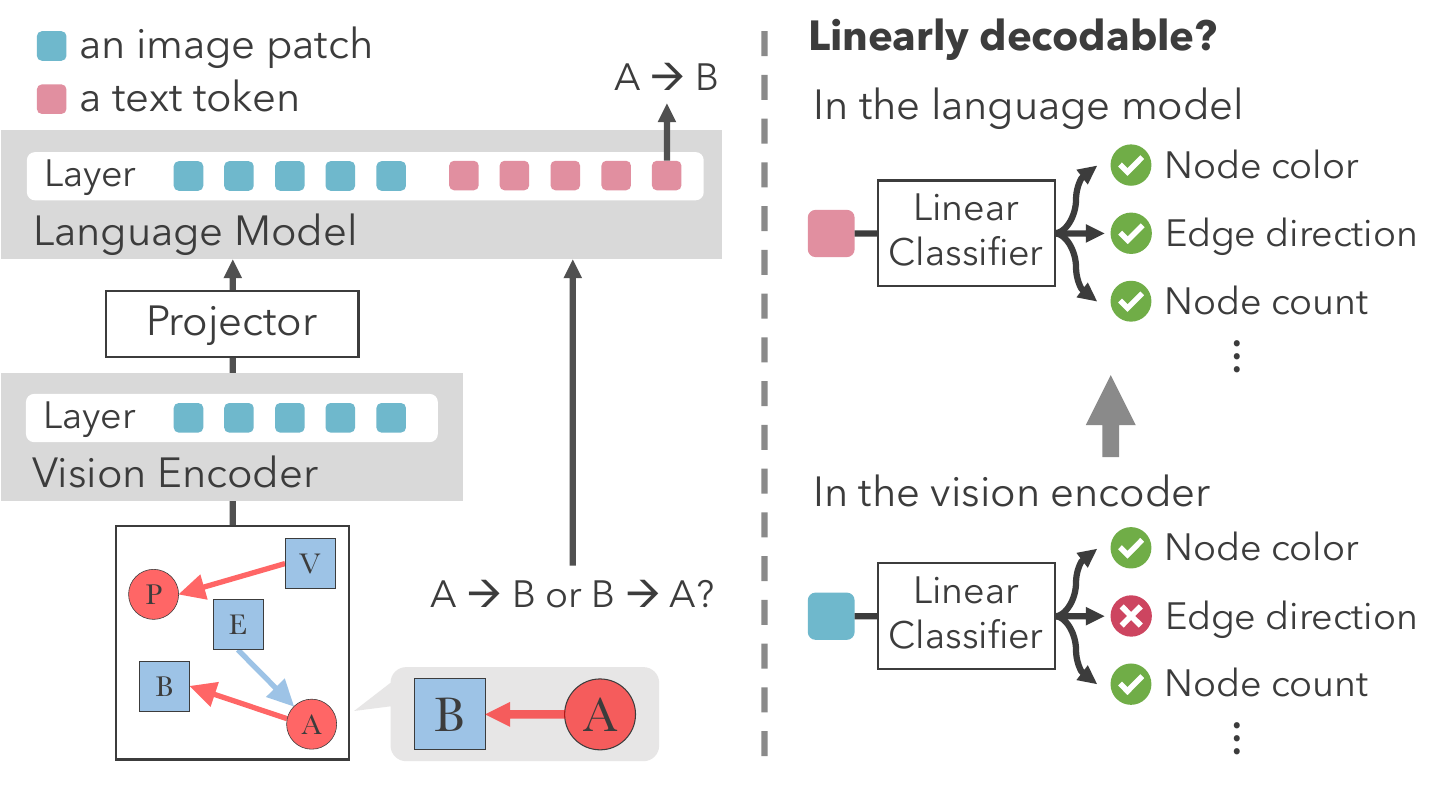}
    \caption{
        Overview of this study.
        We analyze the internal representations of LVLMs using linear probing and causal interventions on a synthetic diagram dataset.
        Even for aspects that LVLMs fail to answer in the VQA setting, the corresponding signals remain linearly decodable from their hidden states, revealing a gap between internal signals and model outputs.
    }
    \label{fig:overview}
\end{figure}

\begin{table}[t]
    \tabcolsep 1mm
    \small
    \centering
    \begin{tabular}{l l l}
        \toprule
        Category & Aspect $a$ & Label set $\mathcal{Y}$ \\
        \midrule
        \multirow[c]{4}{*}{Single}
        & \texttt{Node Color} & \makecell[l]{\{red, green, yellow, blue,\\brown, orange, pink, purple\}} \\
        & \texttt{Node Shape} & \makecell[l]{\{circle, square, pentagon,\\hexagon, septagon\}} \\
        & \texttt{In-deg.} & \{0, 1, 2, 3, 4\} \\
        & \texttt{Out-deg.} & \{0, 1, 2, 3, 4\} \\
        \midrule
        \multirow[c]{5}{*}{Multiple}
        & \texttt{Edge Color} & \makecell[l]{\{red, green, yellow, blue,\\brown, orange, pink, purple\}} \\
        & \texttt{Edge Style} & \{solid, dashed\} \\
        & \texttt{Edge Exist.} & \{exist, not exist\} \\
        & \texttt{Edge Direct.} & \{forward, backward\} \\
        & \texttt{Multi-hop Path} & \{exist, not exist\} \\
        \midrule
        \multirow[c]{2}{*}{Global}
        & \texttt{Node Count} & \{1, 2, 3, 4, 5\} \\
        & \texttt{Edge Count} & \{1, 2, 3, 4, 5\} \\
        \bottomrule
    \end{tabular}
    \caption{
        The 3 categories, 11 aspects, and their corresponding label sets $\mathcal{Y}$.
    }
    \label{tab:dataset_aspects_and_labels}
\end{table}

To fill this gap, we comprehensively investigate how diagram-specific visual elements are internally distinguished in LVLMs.
We focus on whether information about nodes, edges, and global structure is represented internally in the vision encoder and the language model.
In contrast to prior analyses that focus on natural images~\cite{Gandelsman2024-rb, Tao2024-qf}, we target diagrams, which rely on symbolic representations to illustrate structured relationships.
To enable controlled analysis, we construct a synthetic diagram dataset.
We apply probing~\cite{Alain2017-us} to examine the linear decodability of node, edge, and global information, and causal interventions~\cite{Zhang2024-cr} to assess whether the detected signals contribute to model predictions.

Our contributions are threefold.
First, we introduce a synthetic diagram dataset with controllable visual elements, enabling systematic analysis while reducing the influence of domain-specific knowledge.
Second, our probing analysis shows that signals about nodes and edges are linearly decodable from the hidden states of either the vision encoder or the language model.
Importantly, these signals remain linearly decodable even for aspects that the model struggles with in VQA, revealing a gap between internal signals and model outputs.
This suggests that the difficulty may lie not in the absence of relevant signals, but in the model's ability to use them.
Third, our causal interventions show that, for several aspects, the signals identified through probing causally contribute to the model's predictions.
We further confirm that these findings hold on flowcharts, indicating that they are not specific to our synthetic diagrams.
These findings provide insight into how diagram information is represented and used within LVLMs.

\section{Task and Model Setup}
\label{sec:general_setting}
As a testbed for our analysis, we evaluate the VQA task on synthetically created simple diagram images (graphs).
This VQA task can be solved as a classification task.
As shown in Figure~\ref{fig:overview}, the LVLM takes as input a synthetically created simple diagram image $x$ and a question $q$, and outputs the answer $y \in \mathcal{Y}$ (where $\mathcal{Y}$ is the set of answer labels) to the question as text.

In our experiments, we primarily focus on Qwen3-VL-8B-Instruct~\cite{Bai2025-wu} as the main target of analysis.
In addition, we conduct the same experiments on Qwen2.5-VL 7B~\cite{Bai2025-zl}, LLaVA1.5 7B~\cite{Liu2024-xj}, and Gemma3-4B-IT~\cite{Team2025-fs}.
These results are reported in Appendix~\ref{sec:appendix_probing} and~\ref{sec:appendix_intervention}.

\section{Dataset Construction}
\label{sec:dataset_construction}
To investigate how LVLMs perceive basic visual elements in diagrams, we construct a synthetic dataset of directed-graph diagrams.
The diagrams consist of nodes and edges and serve as a simplified approximation of real-world diagrams, retaining only their core visual elements.
This setup allows precise control over evaluation factors such as node color and edge direction.
Moreover, synthetic data helps reduce biases and shortcuts inherited from natural-image and natural-language datasets~\cite{Kafle2017-yu, Gururangan2018-qn, Dancette2021-jn, Healey2024-rp}, enabling fine-grained analysis and evaluation.

\begin{figure}[t]
    \centering
    \includegraphics[width=\linewidth]{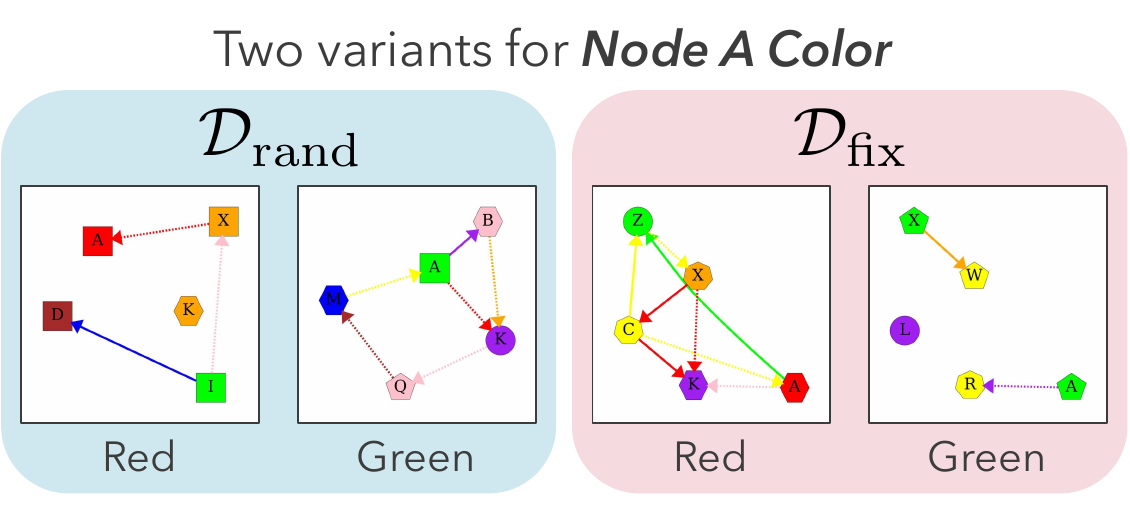}
    \caption{
        Examples of synthetic diagrams.
        Each diagram contains five nodes, and we control evaluation aspects such as node color, shape, and edge connectivity.
        We provide two variants: $\mathcal{D}_{\mathrm{rand}}$, which uses random node layouts (left part), and $\mathcal{D}_{\mathrm{fix}}$, which uses fixed layouts (right part).
    }
    \label{fig:sample_diagrams}
\end{figure}

\subsection{Diagram Specification}
\label{subsec:diagram_specification}

Following prior work on diagram understanding~\cite{Zhu2025-mw, Sartori2025-ec}, each diagram satisfies the following properties:
(1) Nodes and edges are colored using one of eight colors: red, green, yellow, blue, brown, orange, pink, or purple.
(2) Each diagram contains 5 nodes.
(3) Node shapes are one of circle, square, pentagon, hexagon, or septagon.
(4) Each node is labeled with a single alphabetic character as an identifier.
(5) Edge styles are either solid or dashed.
Further details are provided in Appendix~\ref{subsec:appendix_dataset_specification}.

\subsection{Evaluation Aspects}
\label{subsec:aspect_definition}
We define 11 aspects as basic visual information in diagrams, as shown in Table~\ref{tab:dataset_aspects_and_labels}.
These aspects are based on visual information emphasized in prior works~\cite{Silva2011-ec, Pan2024-ls, Sterzik2024-mo, Tseng2025-uv, Hou2025-mf, Zhu2025-mw}.

Diagram understanding fundamentally requires composing multiple visual components, primarily through binding visual features with objects and aggregating information across components.
These operations apply at different spatial scopes, based on which we classify the 11 aspects into the following three categories:
\begin{itemize}
    \item \textbf{Single}: Visual information localized around a single node.
    \item \textbf{Multiple}: Combinations of multiple localized pieces of visual information.
        Since edges are defined between two nodes, their recognition requires a multi-step process: first identifying the endpoint nodes, then recognizing the edge attributes between them.
    \item \textbf{Global}: Visual information that cannot be determined without considering all nodes and edges in the diagram.
\end{itemize}
In analyzing each aspect, we focus on attributes of the node with identifier A (node A) and the edge connecting nodes A and B (edge AB) as representatives\footnote{The choice of identifiers is arbitrary, and the same analysis can be applied to others such as node C.}.
For example, in the \texttt{Node Color} aspect, we evaluate whether the color of node A is represented (i.e., linearly decodable) in the model.
The \texttt{Node Count} aspect is an exception: multiple nodes share the identifier A and the task is to count them, since the total number of nodes is fixed at five.

\subsection{Dataset Definition}
For each aspect $a$, we define a dataset $\mathcal{D}$ to verify whether it can be recognized as follows:
\begin{align}
    \mathcal{D} = \{(x_i, y_i)\}_{i=1}^N.
\end{align}
Here, $x_i$ is the $i$-th diagram image and $y_i$ is its label.
Specifically, this is a dataset containing labels $y_i$ for each aspect as shown in Table~\ref{tab:dataset_aspects_and_labels}.

In this study, we train probes to analyze the internal representations of an LVLM.
To train and evaluate these probes, for each aspect, we construct two variants with different node layouts:
\begin{itemize}
    \item $\mathcal{D}_{\mathrm{rand}}$: For each diagram in the dataset, the node layout is determined independently at random.
    \item $\mathcal{D}_{\mathrm{fix}}$: For all diagrams in the dataset, nodes with each identifier are placed at the same spatial position.
\end{itemize}
Figure~\ref{fig:sample_diagrams} shows examples of each variant.
For each aspect, both $\mathcal{D}_{\mathrm{rand}}$ and $\mathcal{D}_{\mathrm{fix}}$ consist of 100 data points per classification class.

\paragraph{Training dataset for probing.}
\label{subsec:train_dataset}
We construct the diagram dataset with random node layouts, $\mathcal{D}_{\mathrm{train}}$, as the base for the probe training dataset.
In pilot experiments, we observed that the probe achieved near-perfect accuracy on certain patches by exploiting shortcuts, such as predicting based on colors present in the input patch without identifying the target node or edge (see Appendix~\ref{subsec:appendix_probe_shortcut_details} for details).
To prevent this, we prepare an additional dataset $\mathcal{D}^{\bot}_{\mathrm{rand}}$, which consists of diagrams in which the target node or edge does not exist, with the gold label $y = \bot$ (e.g., the question asks about node A, but node A does not exist in the image).
This forces the probe to verify target existence before predicting its attributes, which resolved the shortcut issue in our experiments.
$\mathcal{D}^{\bot}_{\mathrm{rand}}$ contains the same number of samples as $\mathcal{D}_{\mathrm{rand}}$.
Using $\mathcal{D}^{\bot}_{\mathrm{rand}}$, we define the probe training dataset as $\mathcal{D}_{\mathrm{train}} = \mathcal{D}_{\mathrm{rand}} \cup \mathcal{D}^{\bot}_{\mathrm{rand}}$.

\paragraph{Evaluation dataset for probing.}
We prepare the evaluation dataset by constructing $M$ variants of $\mathcal{D}_{\mathrm{fix}}$.
Formally, denoting the $j$-th evaluation subset as $\mathcal{D}^{j}_{\mathrm{fix}}$, the entire evaluation dataset is expressed as $\mathcal{D}_{\mathrm{test}} = \{ \mathcal{D}^{0}_{\mathrm{fix}}, \mathcal{D}^{1}_{\mathrm{fix}}, \cdots, \mathcal{D}^{M-1}_{\mathrm{fix}} \}$.
The fixed node layout, which is not seen during probe training, allows us to analyze the model's internal representations in a controlled manner, and evaluating across $M$ variants ensures that the observed behaviors are not artifacts of a specific spatial configuration.

\begin{table*}[t]
    \tabcolsep 1mm
    \small
    \centering
    \begin{tabular}{lrrrrrrrrrrr}
        \toprule
        & \makecell{\texttt{Node}\\\texttt{Color}}
        & \makecell{\texttt{Node}\\\texttt{Shape}}
        & \makecell{\texttt{In-deg.}}
        & \makecell{\texttt{Out-deg.}}
        & \makecell{\texttt{Edge}\\\texttt{Color}}
        & \makecell{\texttt{Edge}\\\texttt{Style}}
        & \makecell{\texttt{Edge}\\\texttt{Exist.}}
        & \makecell{\texttt{Edge}\\\texttt{Direct.}}
        & \makecell{\texttt{Multi-hop}\\\texttt{Path}}
        & \makecell{\texttt{Node}\\\texttt{Count}}
        & \makecell{\texttt{Edge}\\\texttt{Count}} \\
        \midrule
        Chance level
        & 12.5
        & 20.0
        & 20.0
        & 20.0
        & 12.5
        & 50.0
        & 50.0
        & 50.0
        & 50.0
        & 20.0
        & 20.0 \\
        Qwen3-VL 8B
        & \textbf{95.5}
        & \textbf{78.4}
        & \textbf{53.7}
        & \textbf{40.6}
        & \textbf{80.6}
        & \textbf{85.7}
        & \textbf{81.2}
        & \textbf{53.9}
        & \textbf{55.9}
        & \textbf{78.7}
        & \textbf{83.8} \\
        \bottomrule
    \end{tabular}
    \caption{
        $\mathrm{Acc}_{\mathrm{VQA}}$ for Qwen3-VL 8B.
        Accuracy exceeded chance level for all aspects, but \texttt{Edge Direction} and \texttt{Multi-hop Path} remained near chance, while other aspects achieved substantially higher accuracy.
    }
    \label{tab:vqa_qwen3}
\end{table*}

\section{Preliminary Experiment: VQA}
\label{sec:experiment_vqa}
As a preliminary experiment, we evaluate the VQA performance of LVLMs on synthetic diagrams to confirm above-chance-level performance.
Specifically, we assess whether the models can correctly predict the gold label $y$ given a diagram $x$ and question $q$ in a simple VQA setting.
The question $q$ includes the label set $\mathcal{Y}$ as answer choices.
For example, for the \texttt{Node Color} aspect, the question is ``What color is node A?'' with choices $\mathcal{Y} = \{\text{red}, \text{green}, \ldots\}$.
The same input format is used in the probing and causal intervention experiments.
The full list of questions for each aspect is provided in Table~\ref{tab:questions_for_aspects}.

\subsection{Evaluation Metric}
\label{subsec:eval_metric}
We define the final score as the mean accuracy across all evaluation data subsets.
Formally, it is defined as:
\begin{align}
    \mathrm{Acc}_{\mathrm{VQA}}(q, \mathcal{D}_{\mathrm{test}}\!)\!=\!\!\!\mean_{d_{\mathrm{fix}} \in \mathcal{D}_{\mathrm{test}}}\!\!\mathrm{acc}_{\mathrm{VQA}}(q, d_{\mathrm{fix}}).
\end{align}
Here, $\mathrm{acc}_{\mathrm{VQA}}$ is the accuracy of the LVLM on the VQA task for dataset $d_{\mathrm{fix}} \in \mathcal{D}_{\mathrm{test}}$, and $\mean_{s \in \mathcal{S}}(\cdot)$ denotes the element-wise arithmetic mean of the values associated with each $s \in \mathcal{S}$.
We judge a prediction as correct if the LVLM's output text contains the label $y$ as a substring.

\subsection{Experimental Results}
\label{subsec:vqa_experimental_results}
Table~\ref{tab:vqa_qwen3} summarizes the $\mathrm{Acc}_{\mathrm{VQA}}$ of Qwen3-VL 8B.
The model achieved high accuracy on aspects such as \texttt{Node Color} and \texttt{Edge Style}, while accuracy on \texttt{Edge Direction} and \texttt{Multi-hop Path} remained near chance level.
This suggests that Qwen3-VL recognizes basic visual attributes but struggles with relational properties that require identifying edge directions or tracing multi-step paths.
See Appendix~\ref{sec:details_of_vqa} for additional discussion.

\section{Probing}
\label{sec:probing}
We investigate whether visual information in diagrams is linearly encoded within LVLMs.
To this end, we perform linear probing on the hidden states of each layer and patch in both the vision encoder and the language model, and examine whether there are hidden states from which such signals can be captured.
Linear probing is an effective method for identifying which parts of a model contain signals corresponding to specific information~\cite{Alain2017-us}.

\begin{figure}[t]
    \centering
    \includegraphics[width=\linewidth]{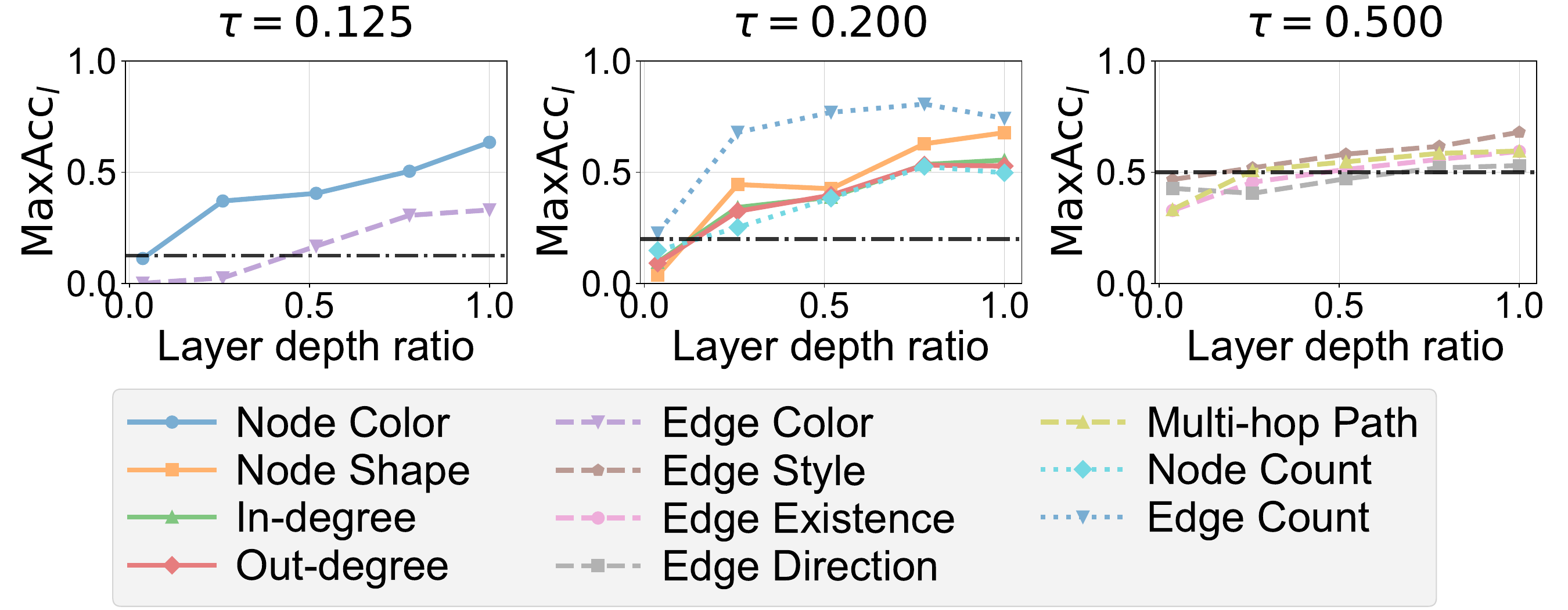}
    \caption{
        $\mathrm{MaxAcc}_{l}$ per layer in the vision encoder of Qwen3-VL 8B.
        The x-axis shows relative layer position (0 is the input layer, 1 is the final layer), and the y-axis shows accuracy.
        The black dash-dot line in each panel indicates the threshold $\tau$.
        Accuracy for Single- and Global-category aspects rose with depth, whereas Multiple-category aspects stayed near the threshold throughout.
    }
    \label{fig:layerwise_image_probing_qwen3_vision_encoder}
\end{figure}

\begin{figure*}[t]
    \centering
    \includegraphics[width=\linewidth]{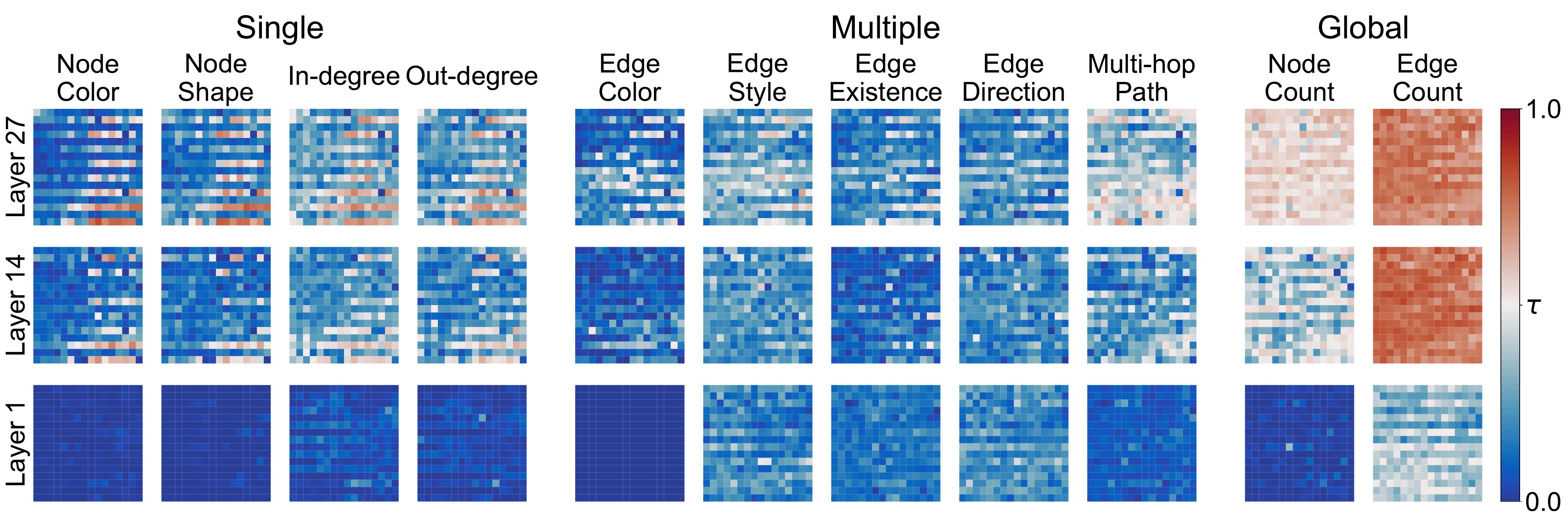}
    \caption{
        Position-wise accuracy in the vision encoder of Qwen3-VL 8B.
        Each heatmap shows accuracy by patch position for a specific layer and aspect.
        The evaluation diagrams use the node layout shown in the right part of Figure~\ref{fig:sample_diagrams}.
        Single-category aspects localized near node positions (bottom-right), Global-category aspects spread across background patches, and Multiple-category aspects showed few high-accuracy patches.
    }
    \label{fig:patchwise_image_probing_qwen3_vision_encoder}
\end{figure*}

\subsection{Probe Definition}
Let $\bm h_{l, t}$ denote the hidden state at layer $l$ and position $t$ of the vision encoder or language model given diagram $x$ and question $q$.
The probe $f$ that predicts the VQA task answer or $\bot$ from the hidden state is defined as a single-layer linear classifier:
\begin{align}
    \hat{y} = f(\bm h_{l, t}) = \argmax_{\mathcal{Y} \cup \{\bot\}} \bm W\bm h_{l, t} + \bm b,
\end{align}
where $\hat{y} \in \mathcal{Y} \cup \{\bot\}$ denotes the predicted answer label, $\bm W \in \mathbb{R}^{(|\mathcal{Y}| + 1)\times |\bm h_{l, t}|}$ is the weight matrix and $\bm b \in \mathbb{R}^{(|\mathcal{Y}| + 1)}$ is the bias term.

We train probes for three components using the training dataset $\mathcal{D}_{\mathrm{train}}$ described in \cref{subsec:train_dataset}: (i) the vision encoder and (ii) the image-input part of the language model, for which we train a per-layer probe $f_l$ from image-patch hidden states $\bm h_{l, t}$ at all patch positions $t$; and (iii) the text-input part of the language model, for which we train a separate probe $f_{l, t}$ for each layer $l$ and each token position $t$ in question $q$.
Training details, including hyperparameters and how we address shortcut learning of probes, are provided in Appendix~\ref{sec:appendix_probing}.

\subsection{Evaluation Metric}
Let $\mathrm{acc}_{l, t}(d_{\mathrm{fix}})$ denote the accuracy of probe $f$ (the match rate between $\hat{y}$ and $y$) when using the hidden state at layer $l$ and position $t$ on evaluation data subset $d_{\mathrm{fix}} \in \mathcal{D}_{\mathrm{test}}$.
The final evaluation score is the mean accuracy across all subsets:
\begin{align}
    \mathrm{Acc}_{l, t}(\mathcal{D}_{\mathrm{test}}) = \mean_{d_{\mathrm{fix}} \in \mathcal{D}_{\mathrm{test}}}\mathrm{acc}_{l, t}(d_{\mathrm{fix}}).
\end{align}
To obtain a per-layer score of how linearly encoded the target information is, we define the maximum probing accuracy $\mathrm{MaxAcc}_{l}$ at layer $l$:
\begin{align}
    \mathrm{MaxAcc}_{l}(\mathcal{D}_{\mathrm{test}})\!=\!\!\!\mean_{d_{\mathrm{fix}} \in \mathcal{D}_{\mathrm{test}}}\!\!(
    \max_{t}{\mathrm{acc}_{l, t}(d_{\mathrm{fix}})}).
\end{align}
That is, $\mathrm{MaxAcc}_{l}$ takes the maximum over positions and averages across subsets.
We introduce a threshold $\tau = 1/|\mathcal{Y}|$ (chance-level accuracy excluding $\bot$), and consider probing successful when $\mathrm{Acc}_{l, t} > \tau$.

\subsection{Experimental Results}
\label{subsec:probing_results}

\begin{figure}[t]
    \centering
    \includegraphics[width=\linewidth]{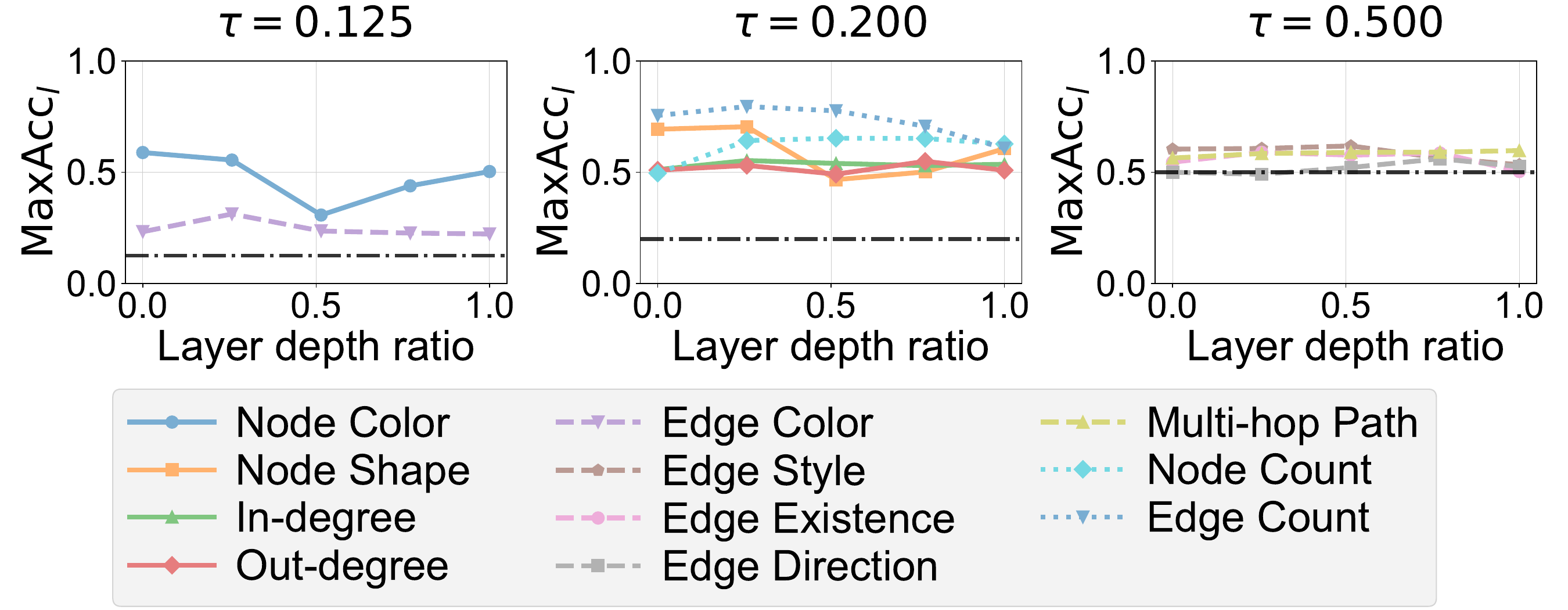}
    \caption{
        $\mathrm{MaxAcc}_{l}$ per layer in the image-input part of the language model in Qwen3-VL 8B.
        The x-axis shows relative layer position (0 is the input layer, 1 is the final layer), and the y-axis shows accuracy.
        The black dash-dot line in each panel indicates the threshold $\tau$.
        Compared with the final vision encoder layer, accuracy showed little change overall.
    }
    \label{fig:layerwise_image_probing_qwen3_language_model}
\end{figure}

\begin{figure*}[t]
    \centering
    \includegraphics[width=\linewidth]{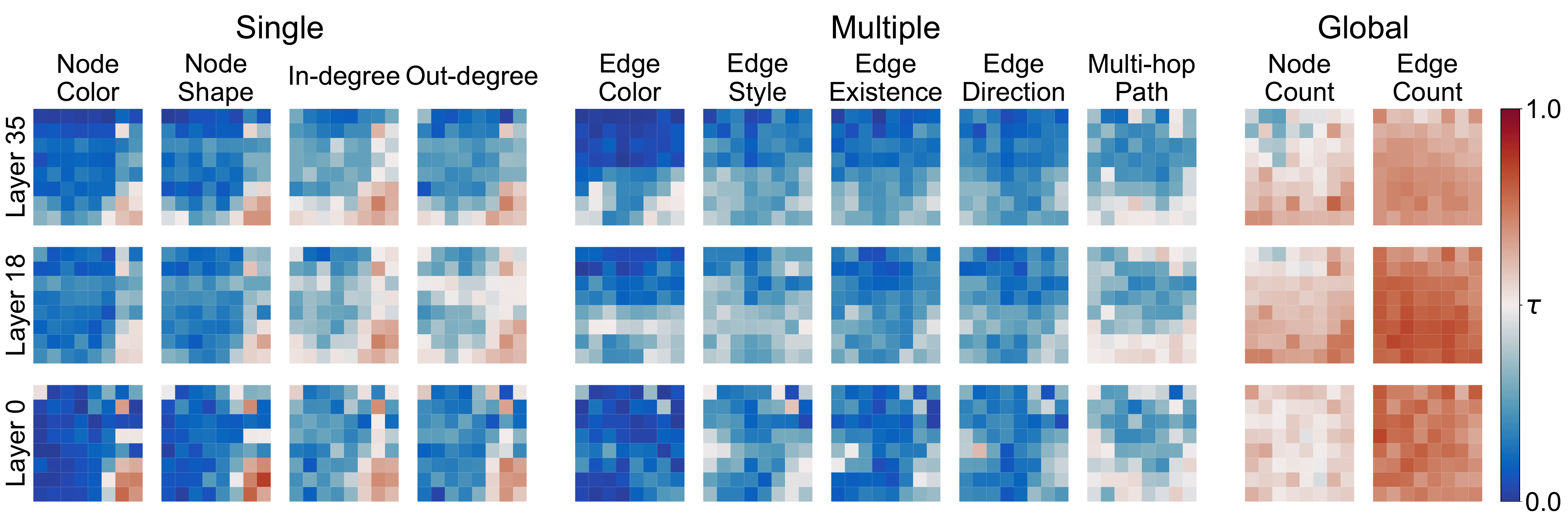}
    \caption{
        $\mathrm{acc}_{l, t}(d_{\mathrm{fix}})$ in the language model of Qwen3-VL 8B for a representative subset $d_{\mathrm{fix}}$.
        The node layout of the evaluation diagrams is the same as that of the diagrams in the right part of Figure~\ref{fig:sample_diagrams}.
        The spatial distribution of high-accuracy patches remained largely unchanged across language model layers.
    }
    \label{fig:patchwise_image_probing_qwen3_language_model}
\end{figure*}

\begin{figure*}[t]
    \centering
    \includegraphics[width=\linewidth]{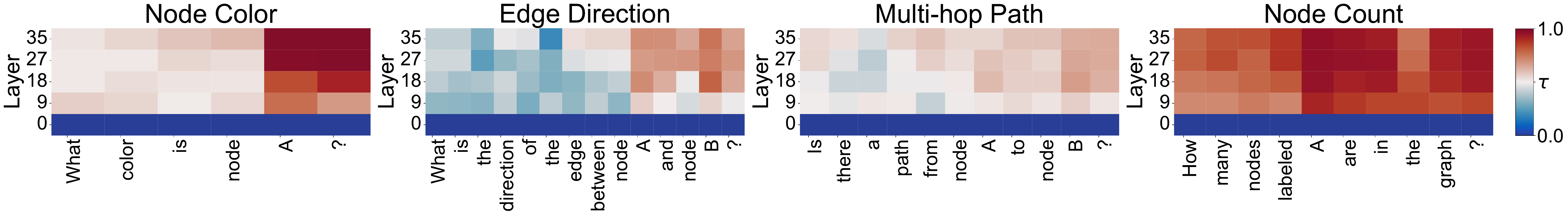}
    \caption{
        Probing results for text tokens in Qwen3-VL 8B.
        The x- and y-axes indicate token position in the question and model layer, respectively.
        Accuracy increased sharply at tokens specifying the target node or edge.
    }
    \label{fig:text_probing_qwen3_body}
\end{figure*}

\paragraph{Signals for Single (node) and Global (counting) aspects become linearly decodable in the vision encoder.}
Figure~\ref{fig:layerwise_image_probing_qwen3_vision_encoder} reports $\mathrm{MaxAcc}_{l}$ at each layer of the vision encoder.
For aspects in the Single and Global categories, accuracy increased with depth, suggesting that the linear decodability of signals for these aspects increases in deeper layers.
In contrast, accuracies for aspects in the Multiple category remained lower than those for the other categories even in the final layers, suggesting that signals for these aspects are more difficult to decode linearly from individual hidden states of the vision encoder.

\paragraph{Signals for Single (node) aspects are localized near node positions, while those for Global (counting) aspects are broadly distributed across patch positions.}
Figure~\ref{fig:patchwise_image_probing_qwen3_vision_encoder} visualizes $\mathrm{acc}_{l, t}(d_{\mathrm{fix}})$ as heatmaps across vision encoder layers and patch positions.
We use the subset $d_{\mathrm{fix}}$ as the examples shown in the right part of Figure~\ref{fig:sample_diagrams} (i.e., diagrams sharing the same node layout).
For aspects in the Single category, high-accuracy patches were concentrated around the target node A in the bottom-right region.
For aspects in the Global category, accuracy exceeded the threshold at almost all positions in later layers, suggesting that signals for these aspects are broadly distributed across patch positions, including background regions.
This high accuracy on background patches suggests that hidden states corresponding to background positions may serve as storage or aggregation points for global information.
Prior work on register tokens~\cite{Darcet2024-ba} in Vision Transformers~\cite{Dosovitskiy2021-ob} reports that global information is concentrated in a small subset (around 2\%) of background tokens.
In contrast, our results suggest that signals for aspects in the Global category are distributed across nearly all background patches in our diagram setting.

\paragraph{Signals remain linearly decodable in the image-input part of the language model.}
Figure~\ref{fig:layerwise_image_probing_qwen3_language_model} shows $\mathrm{MaxAcc}_{l}$ across language model layers.
$\mathrm{MaxAcc}_{l}$ at layer 0 was nearly identical to that at the final vision encoder layer, suggesting that linearly decodable signals are largely preserved through the adapter.
$\mathrm{MaxAcc}_{l}$ then changed only slightly across language model layers for all aspects, indicating that these signals remain linearly decodable throughout image-input processing.
The position-wise analysis (Figure~\ref{fig:patchwise_image_probing_qwen3_language_model}) also showed little change across layers in the number or spatial distribution of high-accuracy positions.

\paragraph{Aspect-specific signals become linearly decodable through text conditioning.}
Figure~\ref{fig:text_probing_qwen3_body} shows $\mathrm{acc}_{l, t}$ across layers and token positions in the text-input part of the language model.
Results for aspects other than \texttt{Node Color}, \texttt{Edge Direction}, \texttt{Multi-hop Path}, and \texttt{Node Count} are provided in Appendix~\ref{subsec:appendix_additional_results}.
For most aspects, accuracy increased sharply at tokens specifying the target nodes or edges.
This suggests that the language model selectively aggregates information from image-position hidden states into text-position hidden states according to the target specified by the text.
In contrast, aspects in the Global category showed relatively high accuracy from the first text token, suggesting that signals for these aspects are already linearly decodable before explicit target specification.
For \texttt{Multi-hop Path}, although accuracy was relatively low compared with other aspects, it slightly exceeded the threshold across a wide range of layers in the text-input part, suggesting that signals for this aspect are linearly decodable, albeit weakly.

In summary, these results show that signals related to node and edge information are linearly decodable from hidden states in either the vision encoder or the language model.
The layers and positions at which these signals are most linearly decodable vary across aspects.
In particular, for \texttt{Edge Direction} and \texttt{Multi-hop Path}, the corresponding signals remain linearly decodable despite low VQA accuracy, revealing a gap between internal signals and model outputs.

\section{Causal Intervention}
\label{sec:causal_intervention}

In \cref{sec:probing}, we found that signals for various diagram aspects are linearly decodable from the hidden states of the vision encoder and the language model.
However, it remains unclear whether these signals causally contribute to the model's predictions.
To test this, we selectively corrupt hidden states with high probing accuracy in both the vision encoder and the language model.
We then measure the resulting change in VQA performance.

\begin{figure}[t]
    \centering
    \includegraphics[width=\linewidth]{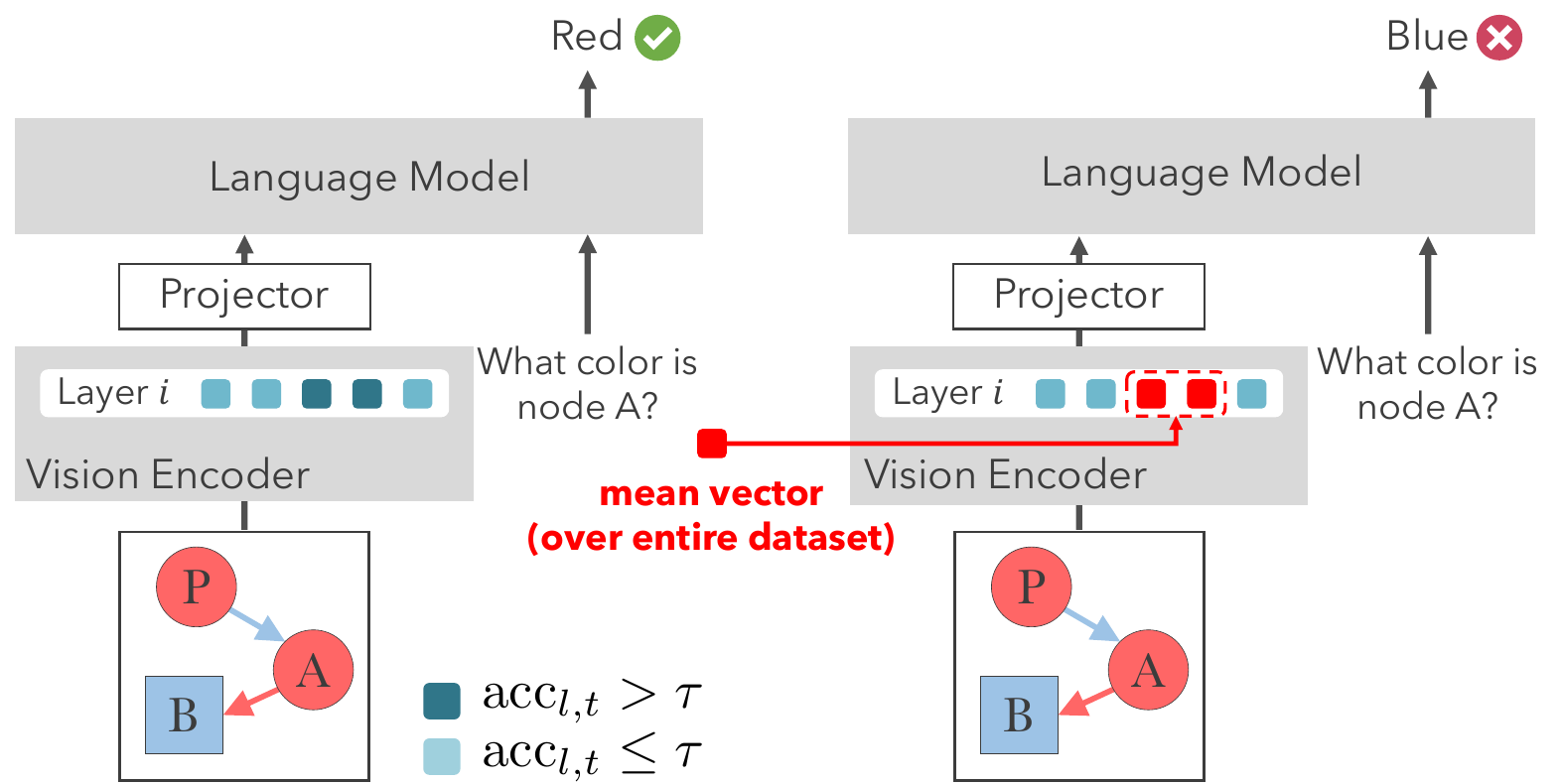}
    \caption{
        Overview of the causal intervention.
        Hidden states at patch positions whose probing accuracy exceeds the threshold are replaced with the uninformative hidden state computed over all images and positions in the evaluation dataset.
    }
    \label{fig:causal_intervention}
\end{figure}

\subsection{Method}
\label{subsec:intervention_method}
As shown in Figure~\ref{fig:causal_intervention}, our procedure is to: (1) identify the target layers and positions, (2) replace the selected hidden states with an uninformative hidden state averaged over the evaluation dataset, and (3) compare VQA accuracy (\cref{subsec:eval_metric}) between interventions on high-probing-accuracy positions and random positions.

\begin{table*}[t]
    \tabcolsep 1mm
    \small
    \centering
    \begin{tabular}{lcrrrrrrrrrrr}
        \toprule
        & Layer
        & \makecell{\texttt{Node}\\\texttt{Color}}
        & \makecell{\texttt{Node}\\\texttt{Shape}}
        & \makecell{\texttt{In-deg.}}
        & \makecell{\texttt{Out-deg.}}
        & \makecell{\texttt{Edge}\\\texttt{Color}}
        & \makecell{\texttt{Edge}\\\texttt{Style}}
        & \makecell{\texttt{Edge}\\\texttt{Exist.}}
        & \makecell{\texttt{Edge}\\\texttt{Direct.}}
        & \makecell{\texttt{Multi-hop}\\\texttt{Path}}
        & \makecell{\texttt{Node}\\\texttt{Count}}
        & \makecell{\texttt{Edge}\\\texttt{Count}} \\
        \midrule
        \multirow{3}{*}{\makecell[l]{Vision\\encoder}}
        & 1
        & \textbf{$-$6.0}
        & $\pm$0.0
        & $\pm$0.0
        & $\pm$0.0
        & $\pm$0.0
        & \textbf{$-$1.1}
        & $\pm$0.0
        & $\pm$0.0
        & $\pm$0.0
        & \textbf{$-$5.5}
        & \textbf{$-$6.0} \\
        & 14
        & \textbf{$-$77.4}
        & \textbf{$-$46.7}
        & \textbf{$-$24.0}
        & \textbf{$-$3.6}
        & \textbf{$-$4.8}
        & \textbf{$-$0.8}
        & \textbf{$-$0.8}
        & $+$0.2
        & \textbf{$-$0.8}
        & \textbf{$-$24.4}
        & $\pm$0.0 \\
        & 27
        & \textbf{$-$78.2}
        & \textbf{$-$55.7}
        & \textbf{$-$17.5}
        & \textbf{$-$11.1}
        & \textbf{$-$32.9}
        & \textbf{$-$8.3}
        & $+$0.1
        & \textbf{$-$0.6}
        & $+$0.6
        & \textbf{$-$5.8}
        & $\pm$0.0 \\
        \midrule
        \multirow{3}{*}{\makecell[l]{Language\\model}}
        & 0
        & \textbf{$-$74.5}
        & \textbf{$-$52.0}
        & \textbf{$-$19.7}
        & \textbf{$-$8.5}
        & \textbf{$-$28.8}
        & \textbf{$-$9.1}
        & \textbf{$-$2.5}
        & \textbf{$-$0.5}
        & \textbf{$-$0.3}
        & \textbf{$-$3.7}
        & $\pm$0.0 \\
        & 18
        & \textbf{$-$16.5}
        & \textbf{$-$10.4}
        & \textbf{$-$3.9}
        & $+$0.1
        & \textbf{$-$30.7}
        & \textbf{$-$4.5}
        & \textbf{$-$0.5}
        & $+$1.8
        & \textbf{$-$1.8}
        & \textbf{$-$0.8}
        & $\pm$0.0 \\
        & 35
        & \textbf{$-$0.1}
        & $+$0.0
        & $\pm$0.0
        & $+$0.1
        & $+$0.0
        & $\pm$0.0
        & $\pm$0.0
        & \textbf{$-$0.2}
        & $+$0.1
        & $+$0.1
        & $\pm$0.0 \\
        \bottomrule
    \end{tabular}
    \caption{
        Accuracy difference between the \textit{accuracy-based} and \textit{controlled} settings (\textit{accuracy-based} $-$ \textit{controlled}).
        More negative values (\textbf{bold}) indicate a larger accuracy decrease under the \textit{accuracy-based} intervention than under the \textit{controlled} intervention.
        For \texttt{Edge Count}, probing accuracy exceeded $\tau$ at all positions in most layers, making the \textit{accuracy-based} and \textit{controlled} settings equivalent ($\pm 0.0$), precluding causal assessment for this aspect.
    }
    \label{tab:causal_intervention_diff}
\end{table*}

\paragraph{Target layers.}
We apply interventions at early, middle, and late layers of both the vision encoder (layers 1, 14, and 27) and the language model (layers 0, 18, and 35)\footnote{We exclude vision encoder layer 0, which contains only patch embeddings, and language model layer 36, whose hidden states, except at the final position, do not affect the output.}.

\paragraph{Target positions.}
We intervene at positions where the probing accuracy exceeds the threshold $\tau$.
We replace all such target positions simultaneously in a single intervention.
For dataset $d_{\mathrm{fix}} \in \mathcal{D}_{\mathrm{test}}$ and layer $l$, the set of target positions for intervention is defined as:
\begin{align}
    \mathcal{S}^{l}_{d_{\mathrm{fix}}} = \{ t \in \{1,\ldots,T\} \mid \mathrm{acc}_{l, t}(d_{\mathrm{fix}}) > \tau \},
\end{align}
where $T$ is the total number of positions.

\paragraph{Intervention operation.}
Following prior work~\cite{Neo2025-ko}, we use the dataset-mean hidden state as an uninformative reference.
For each $d_{\mathrm{fix}} \in \mathcal{D}_{\mathrm{test}}$, we compute $\bm \mu_{l, d_{\mathrm{fix}}}$ at layer $l$ as:
\begin{align}
    \bm \mu_{l, d_{\mathrm{fix}}} = \mean_{x_i \in d_{\mathrm{fix}},\; t \in \{1,\ldots,T\}}\bm h_{i,l,t}.
\end{align}
By averaging over all images and positions, the mean hidden state retains the general activation statistics of the layer but removes information specific to any particular image or position.
For each input $x_i$, we replace the hidden states at intervention target positions $t \in \mathcal{S}^{l}_{d_{\mathrm{fix}}}$ with this vector:
\begin{align}
    \tilde{\bm h}_{i,l,t} =
    \begin{cases}
        \bm \mu_{l, d_{\mathrm{fix}}} & (t \in \mathcal{S}^{l}_{d_{\mathrm{fix}}}),\\
        \bm h_{i,l,t} & (t \not\in \mathcal{S}^{l}_{d_{\mathrm{fix}}})
    \end{cases}.
\end{align}
We compute $\mathrm{Acc}_{\mathrm{VQA}}$ under this intervention setting.
We refer to this setting as \textit{accuracy-based}.

\paragraph{Control experiment.}
To verify that the intervention effect truly comes from targeting high-probing-accuracy positions, we conduct a control experiment that intervenes at randomly selected positions instead.
Specifically, we sample a set $\mathcal{R}^{l}_{d_{\mathrm{fix}}} \subset \{1,\ldots,T\}$ with $|\mathcal{R}^{l}_{d_{\mathrm{fix}}}|=|\mathcal{S}^{l}_{d_{\mathrm{fix}}}|$ and perform replacement as:
\begin{align}
    \tilde{\bm h}_{i,l,t} =
    \begin{cases}
        \bm \mu_{l, d_{\mathrm{fix}}} & (t \in \mathcal{R}^{l}_{d_{\mathrm{fix}}}),\\
        \bm h_{i,l,t} & (t \not\in \mathcal{R}^{l}_{d_{\mathrm{fix}}})
    \end{cases}.
\end{align}
We compute $\mathrm{Acc}_{\mathrm{VQA}}$ under this intervention setting.
We refer to this setting as \textit{controlled}.

\paragraph{Evaluation criteria.}
We compute the accuracy difference as \textit{accuracy-based} $-$ \textit{controlled}.
Because both settings use the same mean-vector replacement procedure, taking the difference controls for generic effects of replacement, allowing us to assess the contribution of signals at the targeted positions.
Therefore, a negative difference indicates that these signals causally contribute to the model's predictions.

\subsection{Results}
Table~\ref{tab:causal_intervention_diff} reports the accuracy difference between the \textit{accuracy-based} and \textit{controlled} settings at each layer of Qwen3-VL 8B.
For the Single category and the \texttt{Edge Color}, \texttt{Edge Style}, and \texttt{Node Count} aspects, we observed a relatively large drop in accuracy compared with the \textit{controlled} setting.
This indicates that, for these aspects, the hidden states of both the vision encoder and the language model contain representations that contribute to the model's predictions.
For \texttt{Edge Direction} and \texttt{Multi-hop Path}, the model's VQA accuracy was already close to chance level (\cref{subsec:vqa_experimental_results}), and thus, little change is expected regardless of the intervention.
Exceptionally, for \texttt{Edge Count}, probing accuracy exceeded $\tau$ at all positions in most layers, so all patches are replaced with $\bm \mu_{l, d_{\mathrm{fix}}}$ in both settings, making the two conditions equivalent ($\pm 0.0$); thus, we cannot determine its causal contribution under our intervention design.

\section{Generalization to Flowcharts}
\label{sec:flowchart_main}

To examine whether our findings generalize beyond synthetic graphs, we conduct VQA, probing, and causal intervention experiments with Qwen3-VL 8B on flowcharts.
We construct a dataset for each aspect in \cref{subsec:aspect_definition} by editing and extending Mermaid code sampled from FlowVQA~\cite{Singh2024-qm}.
We exclude \texttt{Node Shape} because varying node shapes changes the image size, preventing consistent patch-wise evaluation across images.
Unlike the synthetic graphs, these flowcharts contain characteristics commonly found in real-world flowcharts, such as textual labels and conventional layouts.
We follow the experimental setup used for the synthetic graphs, with modifications where necessary for the flowchart setting.
Detailed setup and results are provided in Appendix~\ref{sec:appendix_real_world}.

\paragraph{VQA trends are largely consistent with those observed on synthetic graphs.}
Table~\ref{tab:vqa_flowchart_main} reports VQA accuracy for all aspects.
Accuracy on \texttt{Out-degree}, \texttt{Edge Existence}, \texttt{Edge Direction}, and \texttt{Multi-hop Path} remained near chance level, while accuracy on the other six aspects was clearly above chance level.
This is largely consistent with the results on synthetic graphs, except that \texttt{Out-degree} and \texttt{Edge Existence} exceeded chance level there.

\begin{table*}[t]
    \tabcolsep 1mm
    \small
    \centering
    \begin{tabular}{lrrrrrrrrrr}
        \toprule
        & \makecell{\texttt{Node}\\\texttt{Color}}
        & \makecell{\texttt{In-deg.}}
        & \makecell{\texttt{Out-deg.}}
        & \makecell{\texttt{Edge}\\\texttt{Color}}
        & \makecell{\texttt{Edge}\\\texttt{Style}}
        & \makecell{\texttt{Edge}\\\texttt{Exist.}}
        & \makecell{\texttt{Edge}\\\texttt{Direct.}}
        & \makecell{\texttt{Multi-hop}\\\texttt{Path}}
        & \makecell{\texttt{Node}\\\texttt{Count}}
        & \makecell{\texttt{Edge}\\\texttt{Count}} \\
        \midrule
        Chance level
        & 12.5
        & 25.0
        & \textbf{25.0}
        & 12.5
        & 50.0
        & 50.0
        & \textbf{50.0}
        & \textbf{50.0}
        & 25.0
        & 25.0 \\
        Qwen3-VL 8B
        & \textbf{95.3}
        & \textbf{46.8}
        & \textbf{25.0}
        & \textbf{78.6}
        & \textbf{98.5}
        & \textbf{53.0}
        & \textbf{50.0}
        & \textbf{50.0}
        & \textbf{45.8}
        & \textbf{45.5} \\
        \bottomrule
    \end{tabular}
    \caption{
        $\mathrm{Acc}_{\mathrm{VQA}}$ for Qwen3-VL 8B on flowcharts.
    }
    \label{tab:vqa_flowchart_main}
\end{table*}

\begin{figure}[t]
    \centering
    \includegraphics[width=\linewidth]{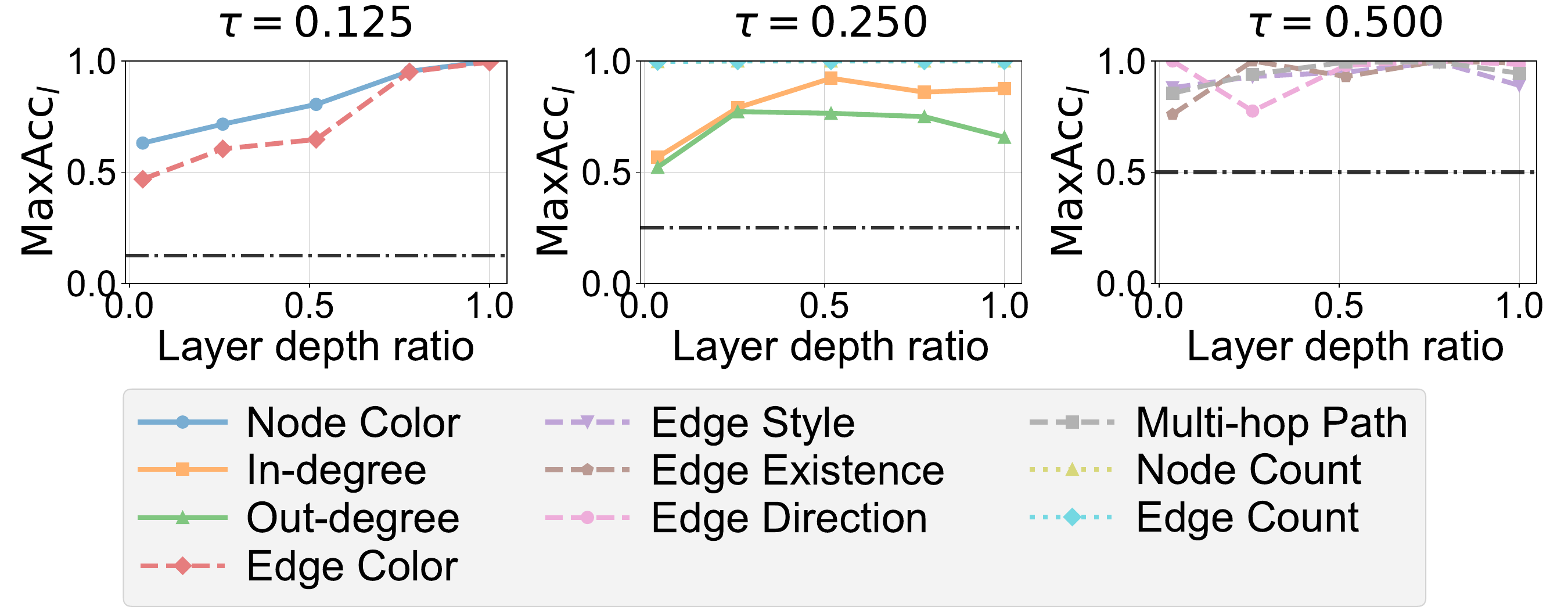}
    \caption{
        $\mathrm{MaxAcc}_{l}$ per layer in the vision encoder of Qwen3-VL 8B on flowcharts.
        Axes and the threshold line follow Figure~\ref{fig:layerwise_image_probing_qwen3_vision_encoder}.
        All aspects exceeded the threshold, including Multiple-category aspects.
    }
    \label{fig:layerwise_real_ve_main}
\end{figure}

\begin{figure}[t]
    \centering
    \includegraphics[width=\linewidth]{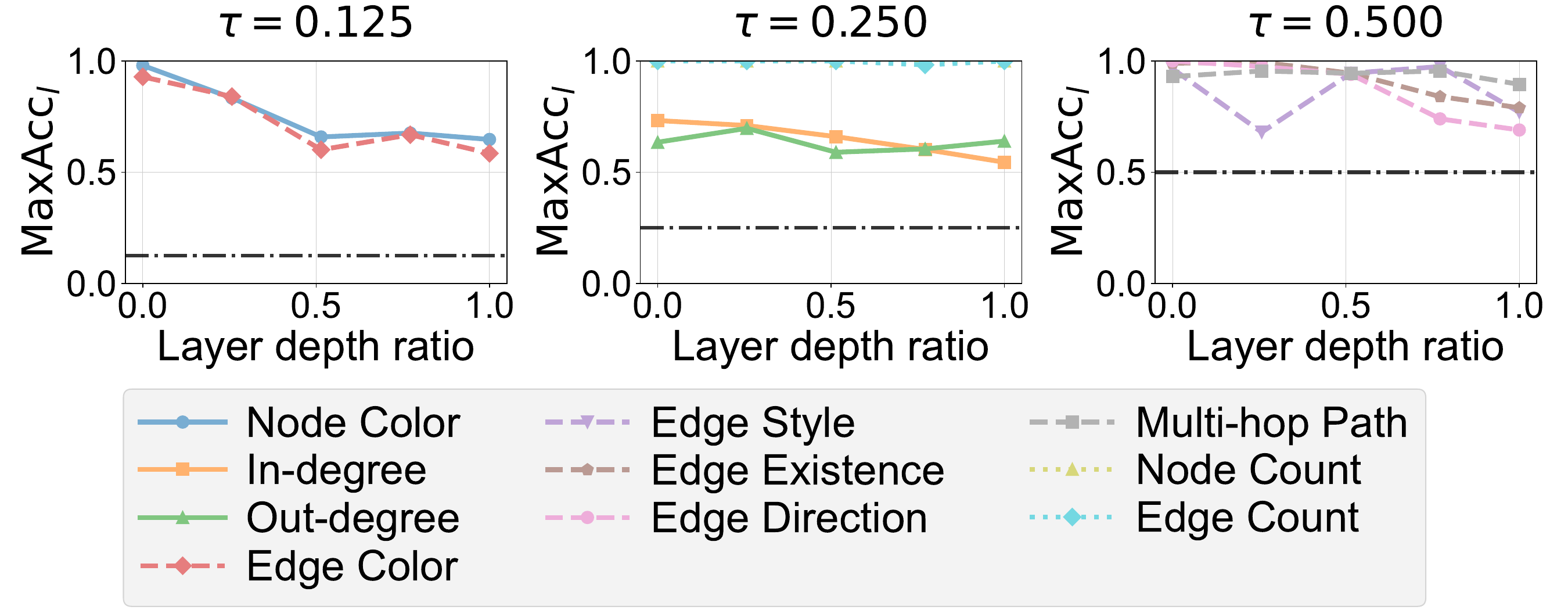}
    \caption{
        $\mathrm{MaxAcc}_{l}$ per layer in the image-input part of the language model in Qwen3-VL 8B on flowcharts.
        Axes and the threshold line follow Figure~\ref{fig:layerwise_image_probing_qwen3_vision_encoder}.
        All aspects maintained accuracy above the threshold.
    }
    \label{fig:layerwise_real_lm_main}
\end{figure}

\paragraph{Signals for all aspects are linearly decodable within the model.}
Figures~\ref{fig:layerwise_real_ve_main} and~\ref{fig:layerwise_real_lm_main} show $\mathrm{MaxAcc}_{l}$ per layer in the vision encoder and the language model, respectively.
Signals for all aspects, including aspects in the Multiple category that the model struggles with in VQA, were linearly decodable from these hidden states.
Consistent with the results on synthetic graphs, this reveals a similar gap between internal signals and model outputs on flowcharts.

\paragraph{Signals identified through probing causally contribute to the model's predictions on flowcharts.}
Table~\ref{tab:causal_intervention_diff_flowchart} reports the difference in VQA accuracy between the \textit{accuracy-based} and \textit{controlled} settings.
For the six aspects with above-chance VQA accuracy, intervening on patches with high probing accuracy decreased VQA accuracy for at least one layer.
This indicates that the signals identified through probing causally contribute to the model's predictions on flowcharts, showing that our finding is not limited to synthetic graphs.
For the remaining aspects, VQA accuracy was already close to chance level, leaving little room for further decreases.

\begin{table*}[t]
    \tabcolsep 1mm
    \small
    \centering
    \begin{tabular}{lcrrrrrrrrrr}
        \toprule
        & Layer
        & \makecell{\texttt{Node}\\\texttt{Color}}
        & \makecell{\texttt{In-deg.}}
        & \makecell{\texttt{Out-deg.}}
        & \makecell{\texttt{Edge}\\\texttt{Color}}
        & \makecell{\texttt{Edge}\\\texttt{Style}}
        & \makecell{\texttt{Edge}\\\texttt{Exist.}}
        & \makecell{\texttt{Edge}\\\texttt{Direct.}}
        & \makecell{\texttt{Multi-hop}\\\texttt{Path}}
        & \makecell{\texttt{Node}\\\texttt{Count}}
        & \makecell{\texttt{Edge}\\\texttt{Count}} \\
        \midrule
        \multirow{3}{*}{\makecell[l]{Vision\\encoder}}
        & 1 & $+$0.6 & $\pm$0.0 & $\pm$0.0 & \textbf{$-$7.4} & \textbf{$-$23.5} & \textbf{$-$5.0} & \textbf{$-$0.5} & $\pm$0.0 & \textbf{$-$3.0} & \textbf{$-$19.2} \\
        & 14 & \textbf{$-$76.8} & \textbf{$-$0.5} & $\pm$0.0 & \textbf{$-$69.5} & $+$5.0 & \textbf{$-$4.5} & $\pm$0.0 & $\pm$0.0 & \textbf{$-$32.5} & $+$1.3 \\
        & 27 & \textbf{$-$75.4} & \textbf{$-$0.8} & $\pm$0.0 & \textbf{$-$58.6} & \textbf{$-$1.0} & $\pm$0.0 & $\pm$0.0 & \textbf{$-$1.5} & \textbf{$-$23.3} & \textbf{$-$1.5} \\
        \midrule
        \multirow{3}{*}{\makecell[l]{Language\\model}}
        & 0 & \textbf{$-$44.0} & \textbf{$-$17.0} & $\pm$0.0 & $+$2.4 & \textbf{$-$47.5} & \textbf{$-$17.5} & $\pm$0.0 & $\pm$0.0 & \textbf{$-$15.5} & \textbf{$-$2.5} \\
        & 18 & \textbf{$-$29.8} & \textbf{$-$14.8} & $\pm$0.0 & \textbf{$-$26.9} & \textbf{$-$37.0} & $+$11.5 & $\pm$0.0 & $+$0.5 & \textbf{$-$7.3} & \textbf{$-$2.5} \\
        & 35 & \textbf{$-$0.1} & $\pm$0.0 & $\pm$0.0 & $\pm$0.0 & $\pm$0.0 & $\pm$0.0 & $\pm$0.0 & $\pm$0.0 & $+$0.3 & $+$0.3 \\
        \bottomrule
    \end{tabular}
    \caption{
        Causal intervention results on flowcharts, where more negative values indicate a larger decrease in VQA accuracy relative to the \textit{controlled} setting.
    }
    \label{tab:causal_intervention_diff_flowchart}
\end{table*}

\section{Related Work}
\label{sec:related_works}

\paragraph{Diagram understanding.}
Prior work on diagram understanding has addressed tasks such as VQA~\cite{Lu2024-hj, Wang2024-ml} and image captioning~\cite{Hsu2021-xf, Huang2023-ow}.
While classical approaches relied on rule-based methods such as pattern matching and predefined schemas~\cite{Blostein1996-on, Song2017-hn}, LVLM-based approaches have recently become prevalent.
Understanding diagrams requires recognizing visual elements such as symbols and shapes, as well as capturing their relationships.
Studies~\cite{Pan2024-ls, Zhu2025-mw, Hou2025-mf} show that LVLMs adequately recognize individual elements but struggle to capture their relationships.
However, the underlying cause of this limitation remains underexplored.
Our work investigates this limitation from the perspective of internal representations by analyzing how LVLMs represent diagram elements and their relationships.

\paragraph{Internal analysis of LVLMs.}
Prior work has examined the internal representations of LVLMs to understand how they process visual information.
\citet{Gandelsman2024-rb} analyzed how natural images are represented within CLIP~\cite{Radford2021-qb}, whose vision encoder is widely used in LVLMs.
They found that multiple attention heads capture specific properties such as color, shape, and number, and that visual information is localized to patches corresponding to specific image regions.
\citet{Tao2024-qf} examined how global and local information from natural images is represented in the language model of LVLMs.
They found that middle layers encode global information, while later layers encode local information.
In contrast, we analyze internal representations for diagrams, where information is conveyed symbolically through text and shapes and structurally through relationships such as edge connectivity.

\paragraph{Gap between internal signals and outputs.}
Prior work has shown a gap between information detectable within LVLMs and their outputs.
Projecting internal representations of LVLMs into the language model's vocabulary space reveals that objects present in the image receive higher confidence than hallucinated ones, even when the model's outputs contain hallucinations~\cite{Jiang2025-tl}.
Similarly, another study found that linear probes on hidden states substantially outperform generation-based predictions of object presence, with this gap primarily originating from the language model's output head~\cite{Li2026-rd}.
Both studies focus on object presence in natural images.
We show that a similar gap arises for relational aspects of diagrams, such as edge direction and the existence of multi-hop paths.

\section{Conclusion}
We constructed a synthetic diagram dataset and applied probing and causal interventions to investigate how LVLMs represent visual information in diagrams.
Our probing analysis shows that signals about nodes and edges are linearly decodable from the hidden states of either the vision encoder or the language model.
Notably, these signals remain linearly decodable even for aspects that the model struggles with in VQA, revealing a gap between internal signals and model outputs.
Causal interventions further show that, for several aspects, the signals identified through probing causally contribute to the model's predictions.
These findings also hold on flowcharts, indicating that they are not specific to our synthetic diagrams.
These findings provide insight into how LVLMs represent and use diagram information.

\clearpage
\section*{Limitations}
\paragraph{Coverage of real-world diagrams.}
We use synthetic diagrams to minimize the influence of dataset biases and shortcut learning that can arise from natural-language and natural-image data.
Although using a synthetic dataset enables well-controlled and fine-grained analysis, our setup does not necessarily capture all complex situations found in real-world diagrams.
To partly address this, we additionally evaluate on flowcharts derived from FlowVQA~\cite{Singh2024-qm} (\cref{sec:flowchart_main}), which contain features characteristic of real-world flowcharts, such as textual labels and conventional layouts.
This evaluation is nevertheless limited in two respects.
First, the flowcharts are still produced by controlled editing and augmentation of node and edge attributes, and are therefore not fully natural.
Second, flowcharts are only one of many diagram formats, and diagrams in other domains may rely on different visual conventions.
We therefore do not claim that our findings extend to real-world diagrams in general, and evaluating a wider range of diagram formats remains future work.

In addition, we focus on basic visual information, such as node color and shape, and edge direction.
Although complex information may be represented by combining these cues, how the model represents composite visual patterns and higher-order structures remains unexamined.

\paragraph{Probing methods.}
Probing has been widely used to analyze internal representations in language models~\cite{Conneau2018-lt, Tenney2019-lh, Campbell2023-gt, Li2023-yp}, but its validity as an interpretability method has also been questioned~\cite{Liu2023-us, Burns2023-fr}.
Our analysis relies on linear probes and therefore measures only linear decodability.
A lack of linear decodability does not imply that the corresponding signal is absent; it may be encoded nonlinearly or distributed across multiple positions.
Future work could examine these possibilities using a broader range of probing methods~\cite{Gurnee2023-bk, Bricken2023-xu}.

\section*{Ethics Statement}
We do not anticipate any significant ethical concerns because (i) we conducted no human-subject experiments, and (ii) our tasks involve only simple, synthetically generated diagrams and do not address ethically sensitive topics.

\section*{Acknowledgments}
This work was supported by AMED Grant Number JP26wm0625405, JST PRESTO Grant Number JPMJFR242S, JST SPRING Grant Number JPMJSP2114, JST BOOST Grant Numbers JPMJBY24D2 and JPMJBS2421, and JSPS KAKENHI Grant Numbers JP24K20829, JP25K03175, and JP25KJ0615, and the Nakajima Foundation.
We thank the members of the Tohoku NLP Group for their advice and support.

\bibliography{main}

\clearpage
\appendix
\setcounter{topnumber}{4}
\setcounter{bottomnumber}{2}
\setcounter{totalnumber}{6}
\setcounter{dbltopnumber}{4}
\renewcommand{\topfraction}{0.95}
\renewcommand{\dbltopfraction}{0.95}
\renewcommand{\bottomfraction}{0.5}
\renewcommand{\textfraction}{0.05}
\renewcommand{\floatpagefraction}{0.6}
\renewcommand{\dblfloatpagefraction}{0.6}

\section{Details of VQA Experiment}
\label{sec:details_of_vqa}

\paragraph{Evaluation setup.}
Table~\ref{tab:questions_for_aspects} shows the questions used to evaluate each aspect in our experiments.

\begin{table*}[t]
    \small
    \centering
    \begin{tabular}{ll}
        \toprule
        Aspect & Question \\
        \midrule
        \texttt{Node Color} & What color is node A? \\
        \texttt{Node Shape} & What shape is node A? \\
        \texttt{In-degree} & What is the in-degree of node A? \\
        \texttt{Out-degree} & What is the out-degree of node A? \\
        \midrule
        \texttt{Edge Color} & What color is the edge between node A and node B? \\
        \texttt{Edge Style} & What style is the edge between node A and node B? \\
        \texttt{Edge Existence} & Does an edge exist between node A and node B? \\
        \texttt{Edge Direction} & What is the direction of the edge between node A and node B? \\
        \texttt{Multi-hop Path} & Is there a path from node A to node B? \\
        \midrule
        \texttt{Node Count} & How many nodes labeled A are in the graph? \\
        \texttt{Edge Count} & How many edges are in the graph? \\
        \bottomrule
    \end{tabular}
    \caption{
        Questions used to evaluate each aspect.
    }
    \label{tab:questions_for_aspects}
\end{table*}

\paragraph{Additional discussion.}
Table~\ref{tab:vqa_qwen3} shows that accuracy varies substantially within the Multiple category: \texttt{Edge Color}, \texttt{Edge Style}, and \texttt{Edge Existence} are answered with high accuracy, whereas \texttt{Edge Direction} and \texttt{Multi-hop Path} remain near chance.
This trend suggests a gradation in the degree and type of integration required.
\texttt{Edge Color} and \texttt{Edge Style} can be determined by inspecting any local part of the edge, and \texttt{Edge Existence} can often be determined by detecting whether a visible connection between the queried nodes is present.
In contrast, \texttt{Edge Direction} requires identifying the two queried endpoint nodes and then locating and interpreting the arrowhead between them.
\texttt{Multi-hop Path} further requires composing directional relations across multiple edges.
These latter aspects impose stronger demands on compositional integration, which may contribute to the observed accuracy gap.

\section{Dataset}

\subsection{Detailed Dataset Specification}
\label{subsec:appendix_dataset_specification}
We select node and edge colors from the 11 Basic Color Terms of \citet{Berlin1999-hp}, excluding white, black, and gray, which leaves eight colors.
For node shapes, we use simple geometric figures without strong semantic associations, including circles and regular polygons.
We exclude equilateral triangles because they resemble arrowheads and may therefore blur the distinction between nodes and edges.
For \texttt{Multi-hop Path}, we exclude direct edges between nodes A and B and place exactly one intermediate node between them, so a positive label always corresponds to a two-edge path.

\section{Preliminary Experiment: VQA}
\label{sec:appendix_vqa}
Table~\ref{tab:vqa_all_models} reports the VQA accuracy of all models evaluated in this study.
Across all models, \texttt{Node Color} achieved relatively high accuracy, while \texttt{In-degree} and \texttt{Out-degree} were consistently lower than the other aspects in the Single category, approaching chance level for LLaVA1.5 7B.
Within the Multiple category, \texttt{Edge Color} and \texttt{Edge Existence} were answered above chance by most models, whereas \texttt{Edge Direction} stayed near chance for all models and \texttt{Multi-hop Path} for most models.
These trends are consistent with the results for Qwen3-VL reported in \cref{subsec:vqa_experimental_results} and suggest that recognizing edge directions and multi-step relational structure is difficult across models.

\begin{table*}[t]
    \tabcolsep 1mm
    \small
    \centering
    \begin{tabular}{lrrrrrrrrrrr}
        \toprule
        & \makecell{\texttt{Node}\\\texttt{Color}}
        & \makecell{\texttt{Node}\\\texttt{Shape}}
        & \makecell{\texttt{In-deg.}}
        & \makecell{\texttt{Out-deg.}}
        & \makecell{\texttt{Edge}\\\texttt{Color}}
        & \makecell{\texttt{Edge}\\\texttt{Style}}
        & \makecell{\texttt{Edge}\\\texttt{Exist.}}
        & \makecell{\texttt{Edge}\\\texttt{Direct.}}
        & \makecell{\texttt{Multi-hop}\\\texttt{Path}}
        & \makecell{\texttt{Node}\\\texttt{Count}}
        & \makecell{\texttt{Edge}\\\texttt{Count}} \\
        \midrule
        Chance level
        & 12.5
        & 20.0
        & 20.0
        & 20.0
        & 12.5
        & 50.0
        & 50.0
        & 50.0
        & 50.0
        & 20.0
        & 20.0 \\
        Qwen3-VL 8B
        & 95.5
        & 78.4
        & 53.7
        & 40.6
        & 80.6
        & 85.7
        & 81.2
        & 53.9
        & 55.9
        & 78.7
        & 83.8 \\
        Qwen2.5-VL 7B
        & 86.6
        & 72.6
        & 40.4
        & 33.8
        & 61.0
        & 73.3
        & 65.5
        & 49.7
        & 53.3
        & 93.8
        & 38.7 \\
        LLaVA1.5 7B
        & 40.1
        & 31.3
        & 22.1
        & 24.9
        & 33.2
        & 47.3
        & 54.4
        & 50.0
        & 50.0
        & 20.0
        & 26.1 \\
        Gemma3-4B-IT
        & 68.7
        & 51.8
        & 33.9
        & 35.9
        & 53.2
        & 67.0
        & 61.8
        & 49.5
        & 50.0
        & 30.9
        & 20.1 \\
        \bottomrule
    \end{tabular}
    \caption{
        VQA accuracy of all models evaluated in this study.
        Across all models, \texttt{Edge Direction} remained near chance and \texttt{Multi-hop Path} was difficult for most models, consistent with the pattern observed for Qwen3-VL in Table~\ref{tab:vqa_qwen3}.
    }
    \label{tab:vqa_all_models}
\end{table*}

\section{Probing}
\label{sec:appendix_probing}

\subsection{Shortcut Learning of Probes}
\label{subsec:appendix_probe_shortcut_details}

In the probing experiments, we observed shortcuts across all aspects where the probe predicted answers based solely on superficial features of the input patches.
These shortcuts resulted in 100\% accuracy on patches in certain regions.
For example, for \texttt{Node Color}, the probe achieved 100\% accuracy on patches containing colored nodes by predicting the local patch color without determining whether the patch corresponds to node A.
Beyond \texttt{Node Color}, we also confirmed shortcuts that exploited information such as patch shape and position.

These shortcuts do not solve the intended probing task.
To correctly classify \texttt{Node Color}, the probe must first identify node A and then predict its color.
However, shortcut-based predictions skip the identification of node A.
These shortcuts arise because superficial patch features are correlated with the target labels under the training distribution.

To address this problem, as described in \cref{subsec:train_dataset}, we add diagrams without node A as dummy data to the training dataset.
This approach ensures that when the input does not contain the information necessary for prediction, the probe outputs the dummy class (N/A), thereby avoiding falsely high accuracy through shortcuts.

\subsection{Probe Training Hyperparameters}
\label{subsec:appendix_probe_training_hyperparameters}

We used linear classifiers for probe training, training separately for each aspect.
The hyperparameters differed only in batch size between image-part probing and text-part probing.
Table~\ref{tab:probe_hyperparameters} shows the hyperparameters used for each type of probing.

\begin{table*}[t]
    \small
    \centering
    \begin{tabular}{lcc}
        \toprule
        Parameter & Image Part & Text Part \\
        \midrule
        Epochs & 1000 & 1000 \\
        Batch size\textsuperscript{*} & \{1000, 2000, 4000, 8000, 16000\} & \{10, 20, 40, 80, 160\} \\
        Learning rate\textsuperscript{*} & [0.0001, 0.01] (log scale) & [0.0001, 0.01] (log scale) \\
        Optimizer & AdamW & AdamW \\
        \quad betas & [0.9, 0.999] & [0.9, 0.999] \\
        \quad eps & 1e-8 & 1e-8 \\
        \quad weight decay & 0.01 & 0.01 \\
        \bottomrule
    \end{tabular}
    \caption{
        Hyperparameters used for probe training.
        Batch size and learning rate were subject to hyperparameter search using Optuna.
        Optimizer parameters use PyTorch's AdamW default values.
        \textsuperscript{*}Subject to hyperparameter search
    }
    \label{tab:probe_hyperparameters}
\end{table*}

For hyperparameter search, we used Optuna~\cite{Akiba2019-wt} with TPESampler to perform three hyperparameter search trials.
The search targets were batch size and learning rate, where batch size was selected categorically from the candidates shown in Table~\ref{tab:probe_hyperparameters}, and learning rate was searched on a log scale ranging from 0.0001 to 0.01.
For the evaluation metric, we split 10\% of the training data as validation data and used accuracy on the validation data.
We did not use early stopping during training; instead, we evaluated performance on the validation data at each epoch and saved the model that achieved the highest accuracy.

\subsection{Additional Results}
\label{subsec:appendix_additional_results}

In this section, we present additional probing analyses beyond the Qwen3-VL 8B results reported in the main text.
These include detailed results for Qwen3-VL 8B, robustness checks with different node layouts, larger visual elements, and 7-node graphs, as well as results for additional models (Qwen2.5-VL 7B, LLaVA1.5 7B, Gemma3-4B-IT).

\begin{figure*}[t]
    \centering
    \includegraphics[width=\linewidth]{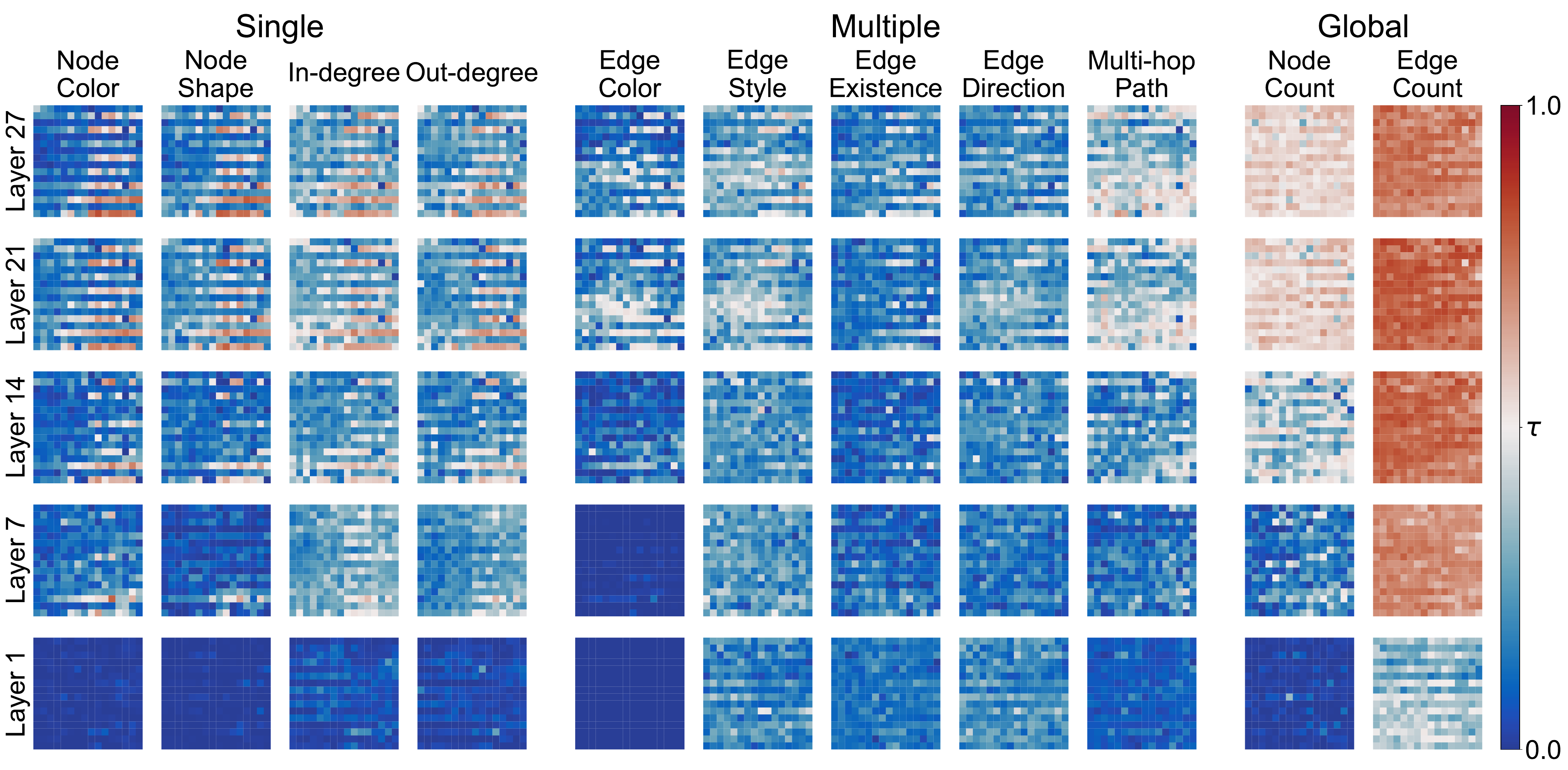}
    \caption{
        Accuracy for each hidden state in the vision encoder of Qwen3-VL 8B.
        Each heatmap shows the accuracy of individual hidden states for a specific layer and aspect.
        Spatial patterns were consistent with Figure~\ref{fig:patchwise_image_probing_qwen3_vision_encoder}.
    }
    \label{fig:patchwise_image_probing_qwen3_vision_encoder_seed_43}
\end{figure*}

\begin{figure*}[t]
    \centering
    \includegraphics[width=\linewidth]{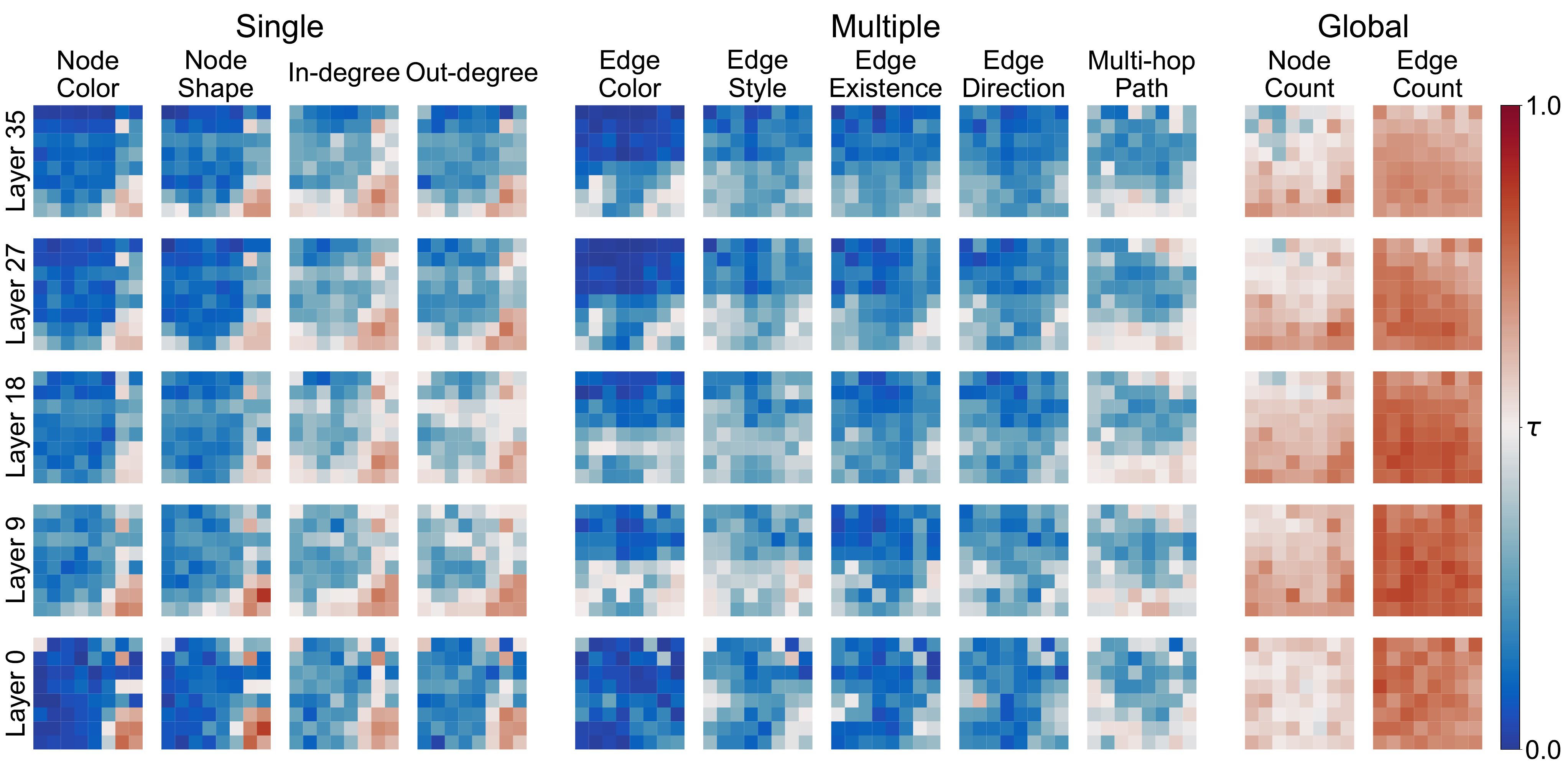}
    \caption{
        Accuracy for each hidden state in the image-input part of the Qwen3-VL 8B language model.
        Each heatmap shows the accuracy of individual hidden states for a specific layer and aspect.
        Spatial patterns were consistent with Figure~\ref{fig:patchwise_image_probing_qwen3_language_model}.
    }
    \label{fig:patchwise_image_probing_qwen3_language_model_seed_43}
\end{figure*}

\begin{figure*}[t]
    \centering
    \includegraphics[width=\linewidth]{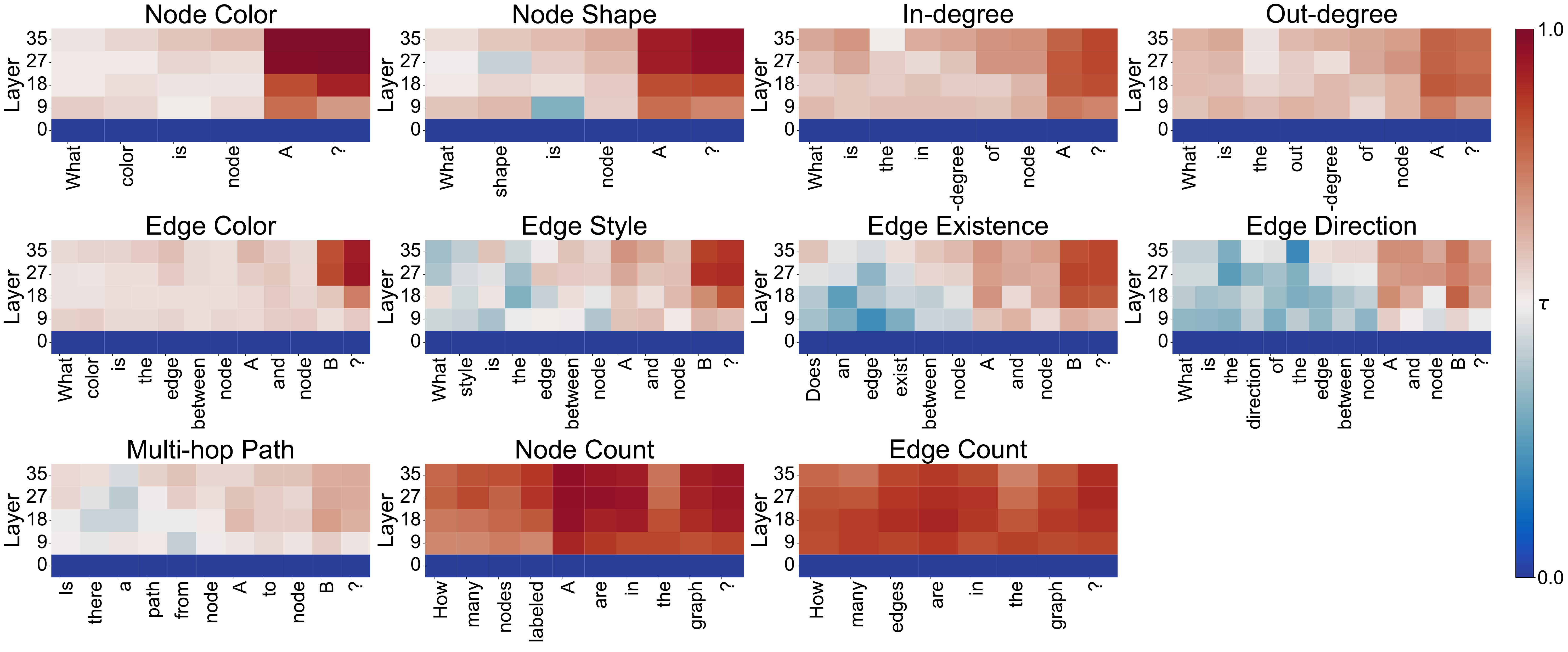}
    \caption{
        Probing results for text tokens in Qwen3-VL 8B.
        The x-axis shows token position in the question, and the y-axis shows layer position.
        Trends were consistent with Figure~\ref{fig:text_probing_qwen3_body}.
    }
    \label{fig:text_probing_qwen3}
\end{figure*}

\paragraph{Qwen3-VL 8B (detailed probing results).}
Figures~\ref{fig:patchwise_image_probing_qwen3_vision_encoder_seed_43} and~\ref{fig:patchwise_image_probing_qwen3_language_model_seed_43} present image probing results with two additional intermediate layers that were omitted in Figures~\ref{fig:patchwise_image_probing_qwen3_vision_encoder} and~\ref{fig:patchwise_image_probing_qwen3_language_model}.
As described in \cref{subsec:probing_results}, Single category information is encoded near node positions, while Global category information is distributed across background regions.
Figure~\ref{fig:text_probing_qwen3} presents text probing results for aspects omitted from Figure~\ref{fig:text_probing_qwen3_body}.
As described in \cref{subsec:probing_results}, accuracy sharply increased at token positions that specify the target nodes or edges.

\begin{figure}[t]
    \centering
    \includegraphics[width=\linewidth]{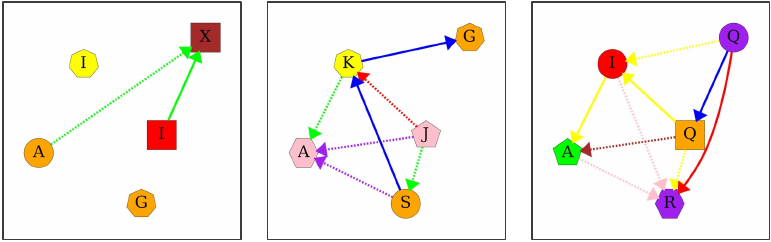}
    \caption{
        Example diagrams with layouts distinct from those in Figure~\ref{fig:sample_diagrams}.
    }
    \label{fig:different_layouts}
\end{figure}

\begin{figure*}[t]
    \centering
    \includegraphics[width=\linewidth]{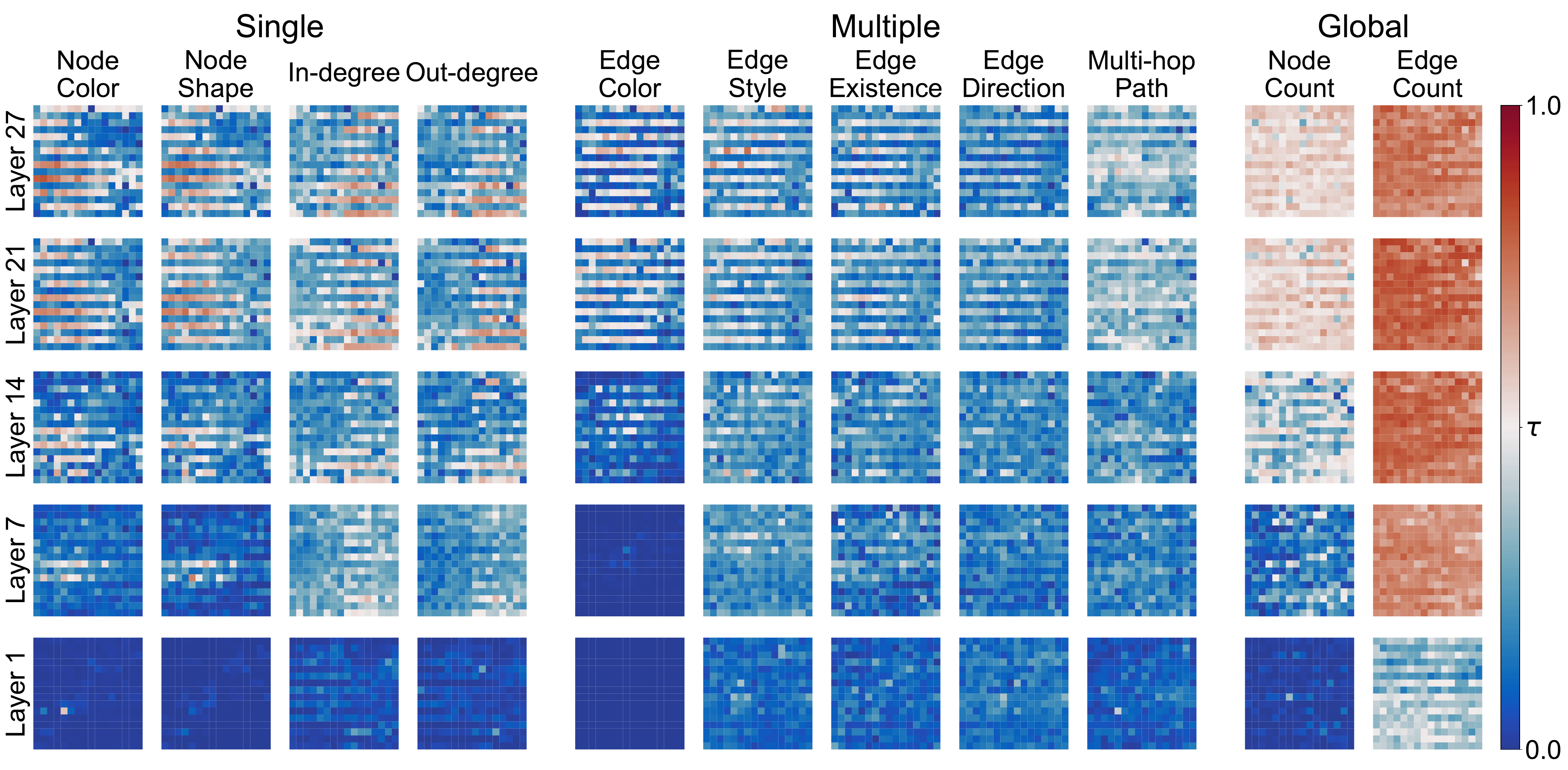}
    \caption{
        Accuracy for each hidden state in the vision encoder of Qwen3-VL 8B (node A placed on the left side).
        Each heatmap shows the accuracy of individual hidden states for a specific layer and aspect.
        High-accuracy patches for Single-category aspects localized on the left side, following node A's position.
    }
    \label{fig:patchwise_image_probing_qwen3_vision_encoder_seed_44}
\end{figure*}

\begin{figure*}[t]
    \centering
    \includegraphics[width=\linewidth]{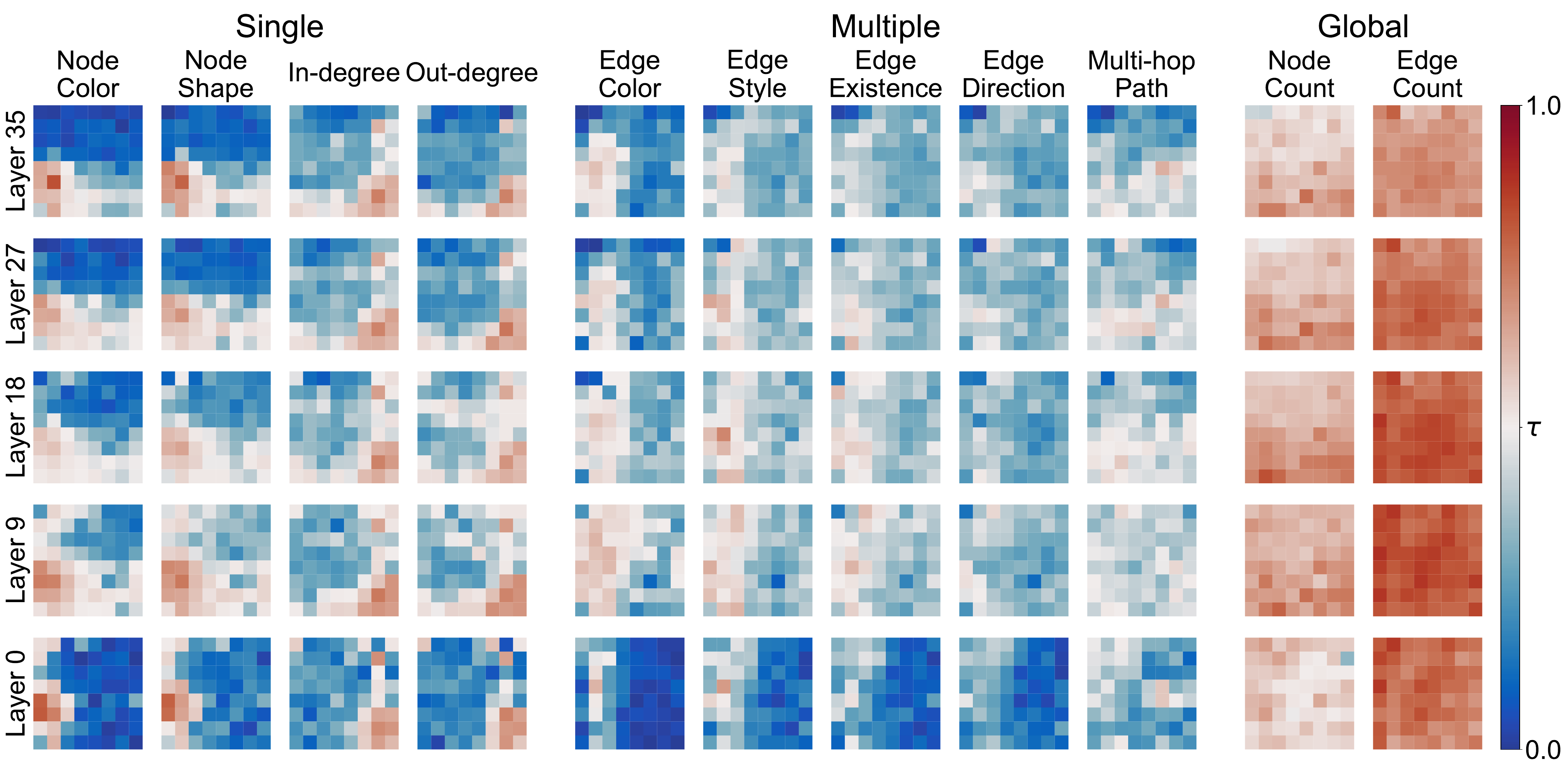}
    \caption{
        Accuracy for each hidden state in the image-input part of the Qwen3-VL 8B language model (node A placed on the left side).
        Each heatmap shows the accuracy of individual hidden states for a specific layer and aspect.
        High-accuracy patches for Single-category aspects localized on the left side, following node A's position.
    }
    \label{fig:patchwise_image_probing_qwen3_language_model_seed_44}
\end{figure*}

\paragraph{Qwen3-VL 8B (different node layouts).}
Figures~\ref{fig:patchwise_image_probing_qwen3_vision_encoder_seed_44} and~\ref{fig:patchwise_image_probing_qwen3_language_model_seed_44} show results with different node layouts (see Figure~\ref{fig:different_layouts}).
Whereas the evaluation diagrams in the main text place the target node (node A) at the bottom right, here node A is placed on the left side.
The distribution of patches with high accuracy for node information changed according to the position of node A.
This positional shift indicates that the decoded signals track the location of the target node rather than a fixed image region.

\begin{figure}[t]
    \centering
    \includegraphics[width=\linewidth]{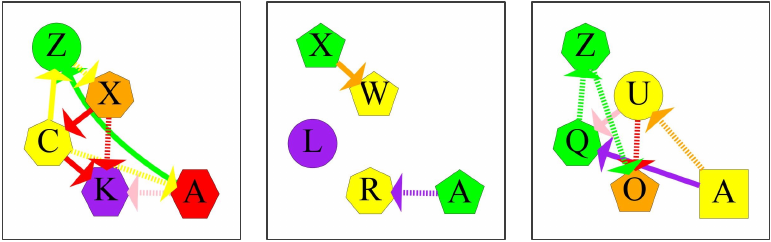}
    \caption{
        Example diagrams with larger nodes and edges compared to those in Figure~\ref{fig:sample_diagrams}.
    }
    \label{fig:large_node_edge_examples}
\end{figure}

\begin{figure*}[t]
    \centering
    \includegraphics[width=\linewidth]{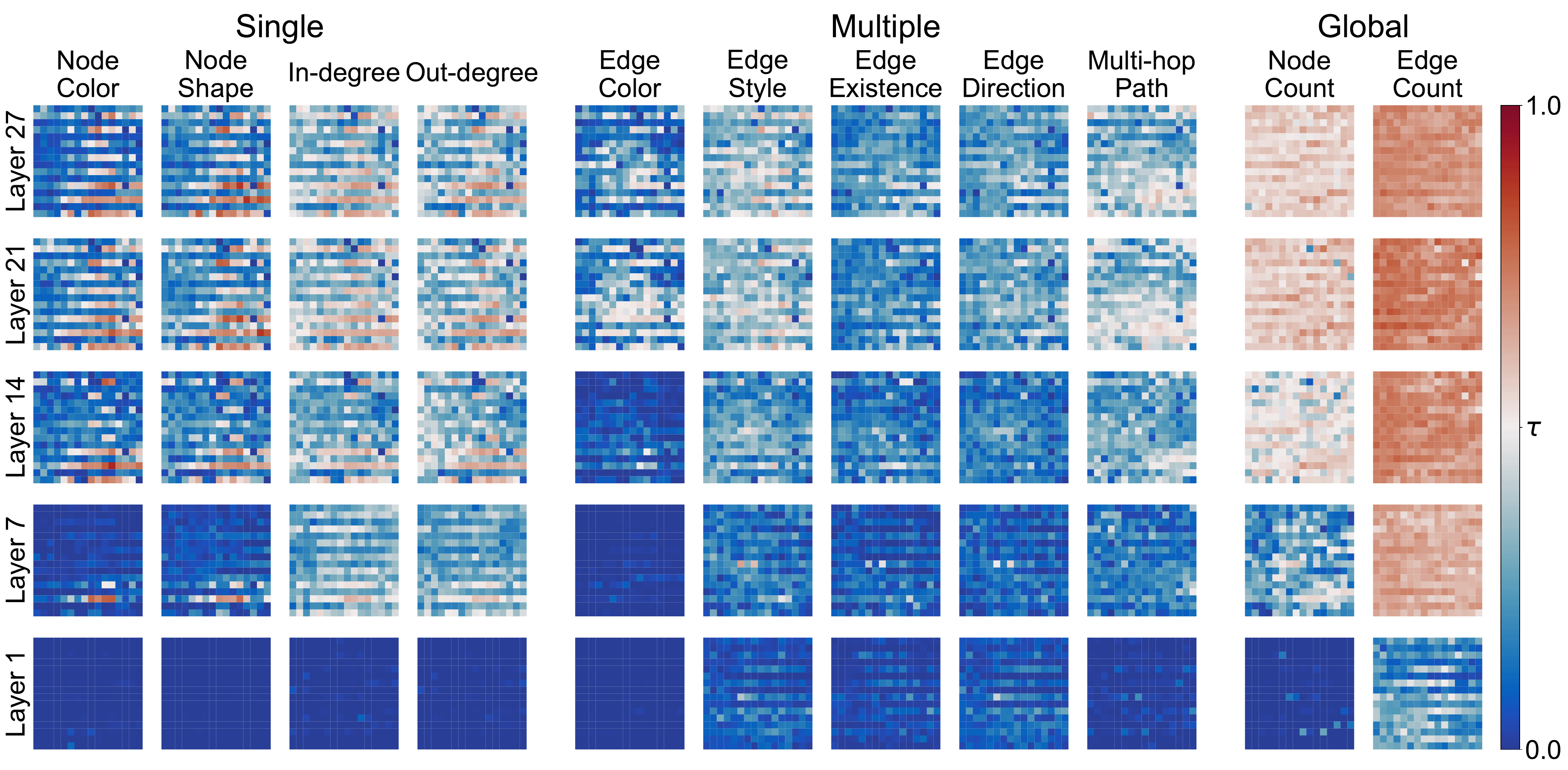}
    \caption{
        Accuracy for each hidden state in the vision encoder of Qwen3-VL 8B (larger nodes and edges).
        Each heatmap shows the accuracy of individual hidden states for a specific layer and aspect.
        Spatial patterns were consistent with Figure~\ref{fig:patchwise_image_probing_qwen3_vision_encoder}.
    }
    \label{fig:patchwise_image_probing_qwen3_vision_encoder_large}
\end{figure*}

\begin{figure*}[t]
    \centering
    \includegraphics[width=\linewidth]{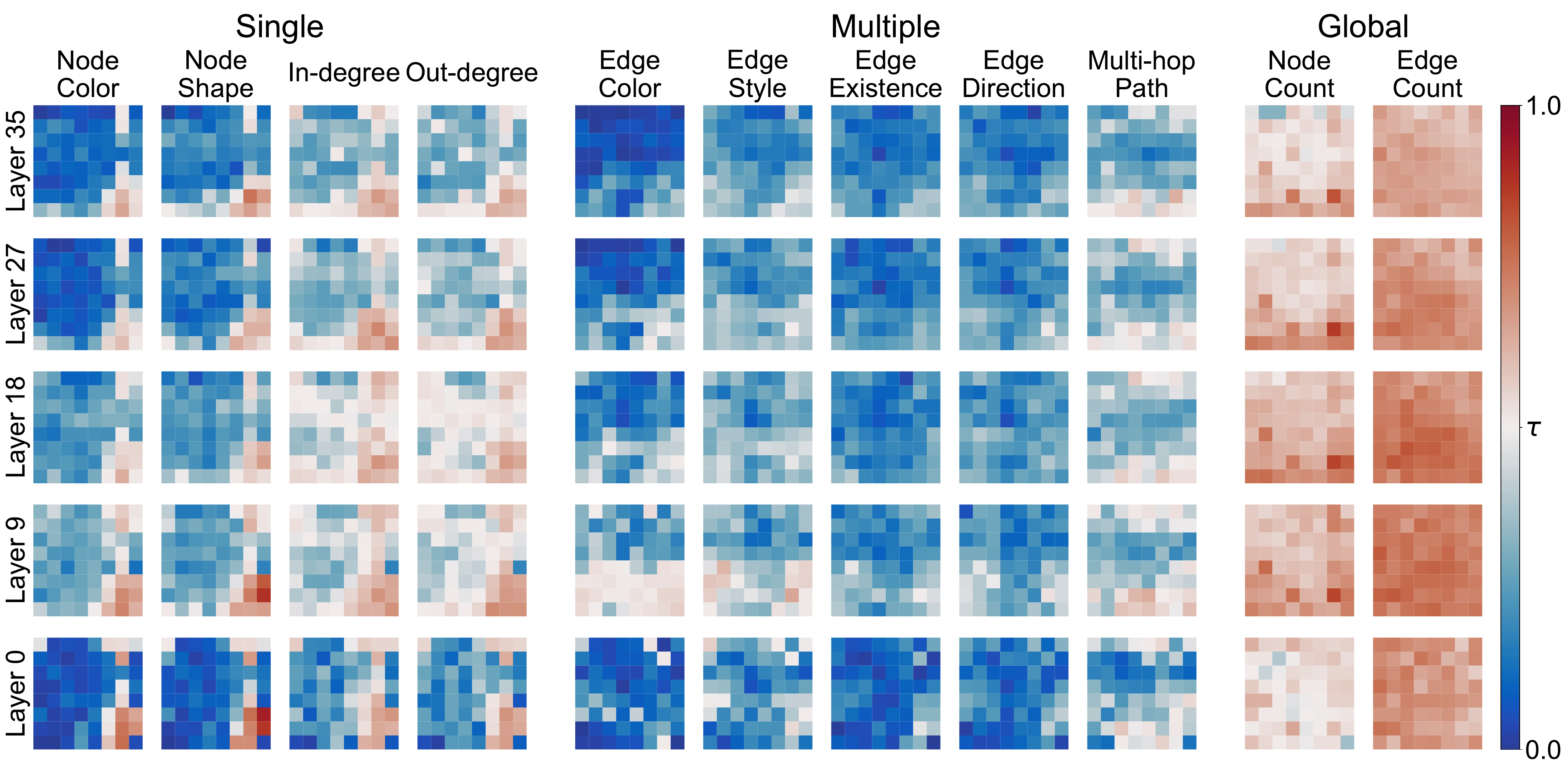}
    \caption{
        Accuracy for each hidden state in the image-input part of the Qwen3-VL 8B language model (larger nodes and edges).
        Each heatmap shows the accuracy of individual hidden states for a specific layer and aspect.
        Spatial patterns were consistent with Figure~\ref{fig:patchwise_image_probing_qwen3_language_model}.
    }
    \label{fig:patchwise_image_probing_qwen3_language_model_large}
\end{figure*}

\paragraph{Qwen3-VL 8B (larger nodes and edges).}
Figures~\ref{fig:patchwise_image_probing_qwen3_vision_encoder_large} and~\ref{fig:patchwise_image_probing_qwen3_language_model_large} show probing results when the sizes of nodes and edges are increased (see Figure~\ref{fig:large_node_edge_examples} for example diagrams).
The results exhibit the same trends as the default-size setting reported in \cref{subsec:probing_results}, indicating that the observed patterns are robust to this change in visual-element size.

\begin{figure}[t]
    \centering
    \includegraphics[width=\linewidth]{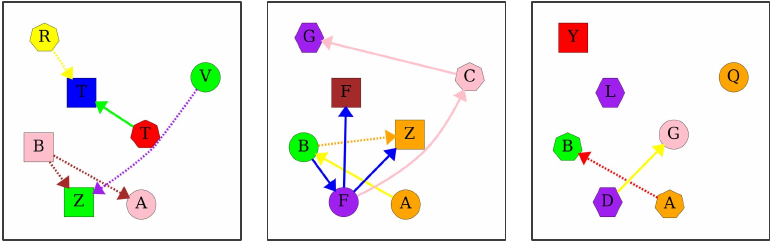}
    \caption{
        Example diagrams with 7 nodes compared to those in Figure~\ref{fig:sample_diagrams}.
    }
    \label{fig:sample_diagrams_7nodes}
\end{figure}

\begin{figure}[t]
    \centering
    \includegraphics[width=\linewidth]{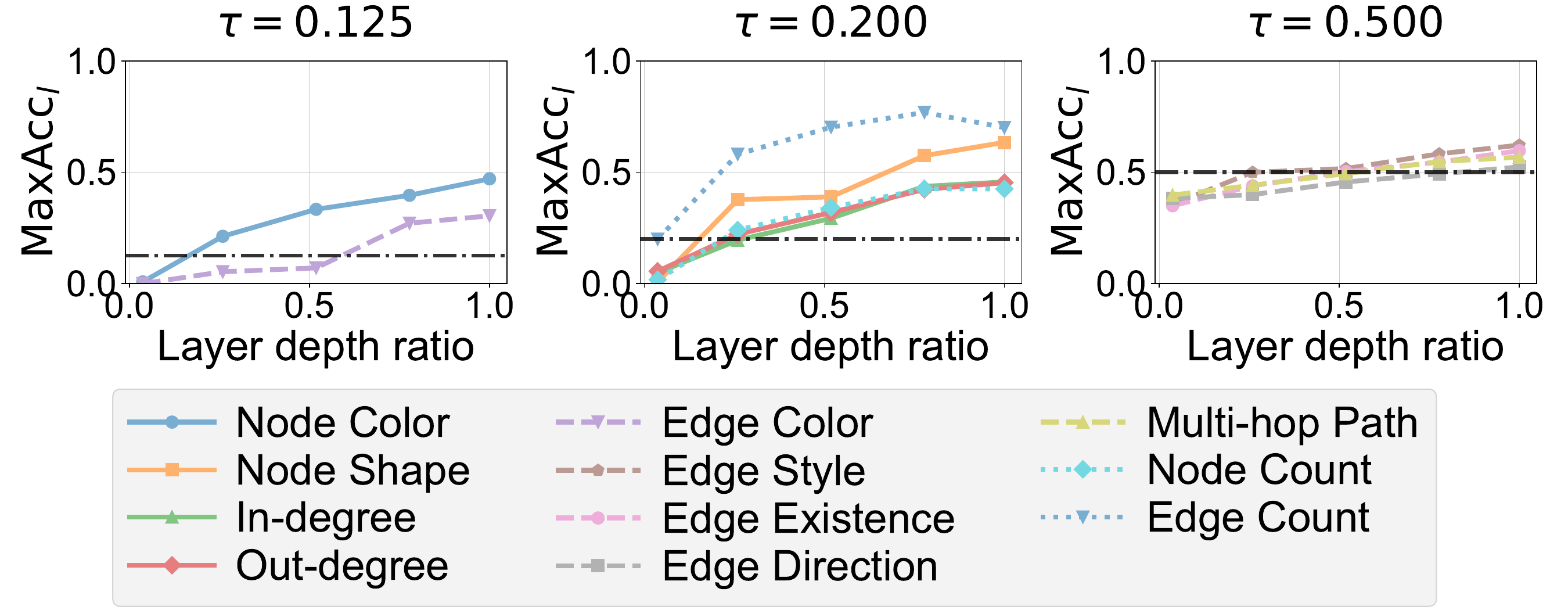}
    \caption{
        $\mathrm{MaxAcc}_{l}$ per layer in the vision encoder of Qwen3-VL 8B (7-node graphs).
        Axes and the threshold line follow Figure~\ref{fig:layerwise_image_probing_qwen3_vision_encoder}.
        Trends were similar to the 5-node setting (Figure~\ref{fig:layerwise_image_probing_qwen3_vision_encoder}), although the overall accuracy was slightly lower.
    }
    \label{fig:layerwise_image_probing_qwen3_vision_encoder_7nodes}
\end{figure}

\begin{figure}[t]
    \centering
    \includegraphics[width=\linewidth]{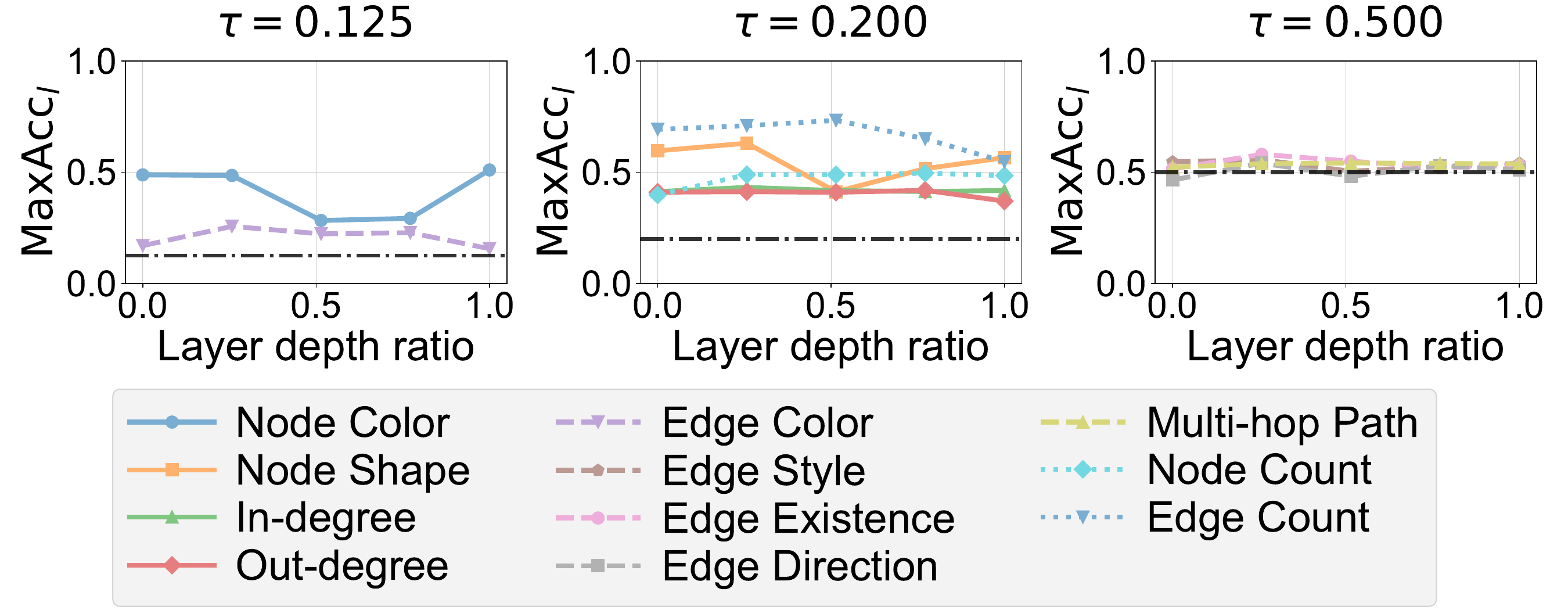}
    \caption{
        $\mathrm{MaxAcc}_{l}$ per layer in the language model of Qwen3-VL 8B (7-node graphs).
        Axes and the threshold line follow Figure~\ref{fig:layerwise_image_probing_qwen3_language_model}.
        Trends were similar to the 5-node setting (Figure~\ref{fig:layerwise_image_probing_qwen3_language_model}), although the overall accuracy was slightly lower.
    }
    \label{fig:layerwise_image_probing_qwen3_language_model_7nodes}
\end{figure}

\begin{figure*}[t]
    \centering
    \includegraphics[width=\linewidth]{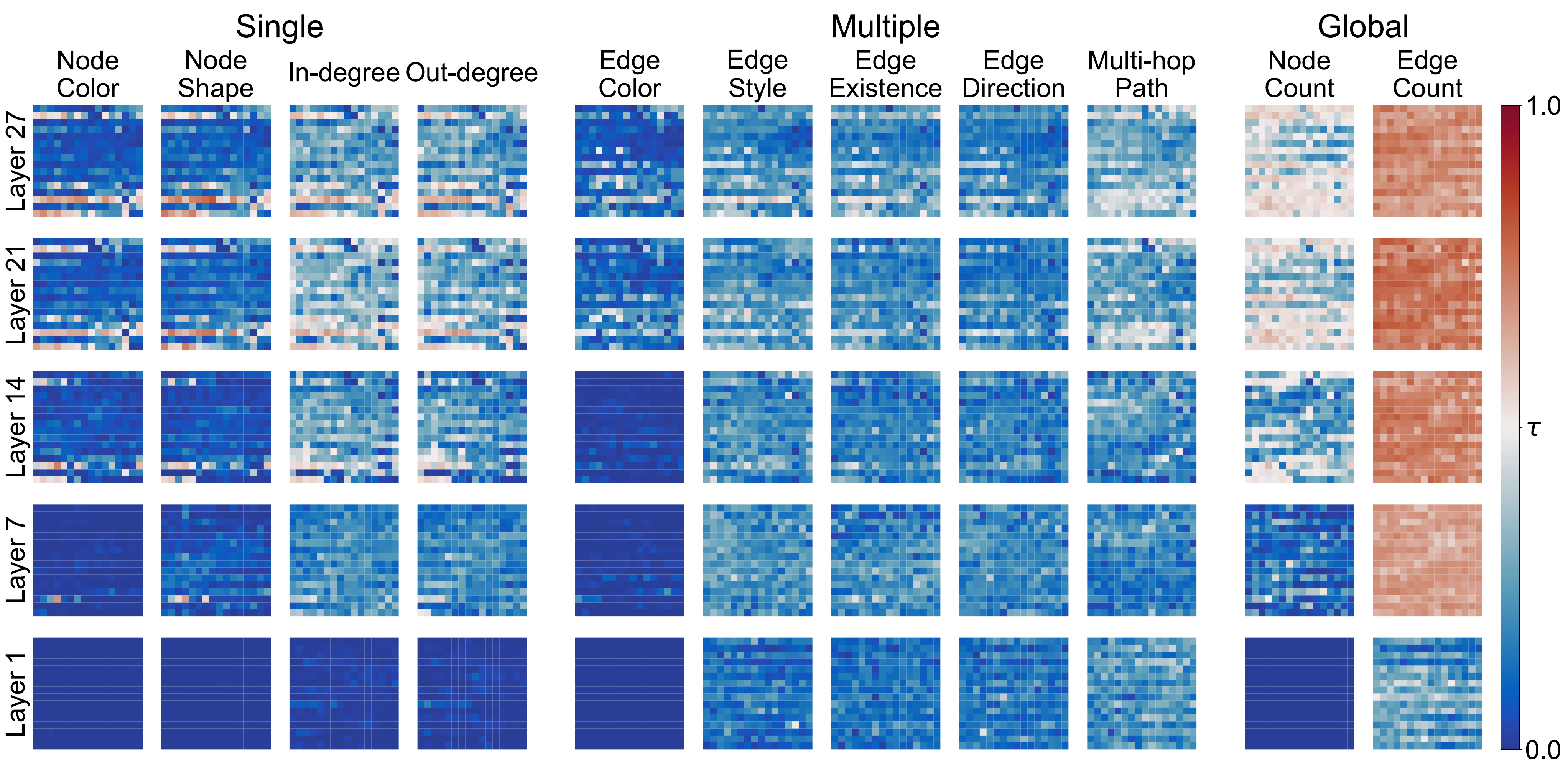}
    \caption{
        Accuracy for each hidden state in the vision encoder of Qwen3-VL 8B (7-node graphs).
        Each heatmap shows the accuracy of individual hidden states for a specific layer and aspect.
        Spatial patterns were consistent with the 5-node setting (Figure~\ref{fig:patchwise_image_probing_qwen3_vision_encoder}), although fewer patches exceeded the threshold.
    }
    \label{fig:patchwise_image_probing_qwen3_vision_encoder_7nodes}
\end{figure*}

\begin{figure*}[t]
    \centering
    \includegraphics[width=\linewidth]{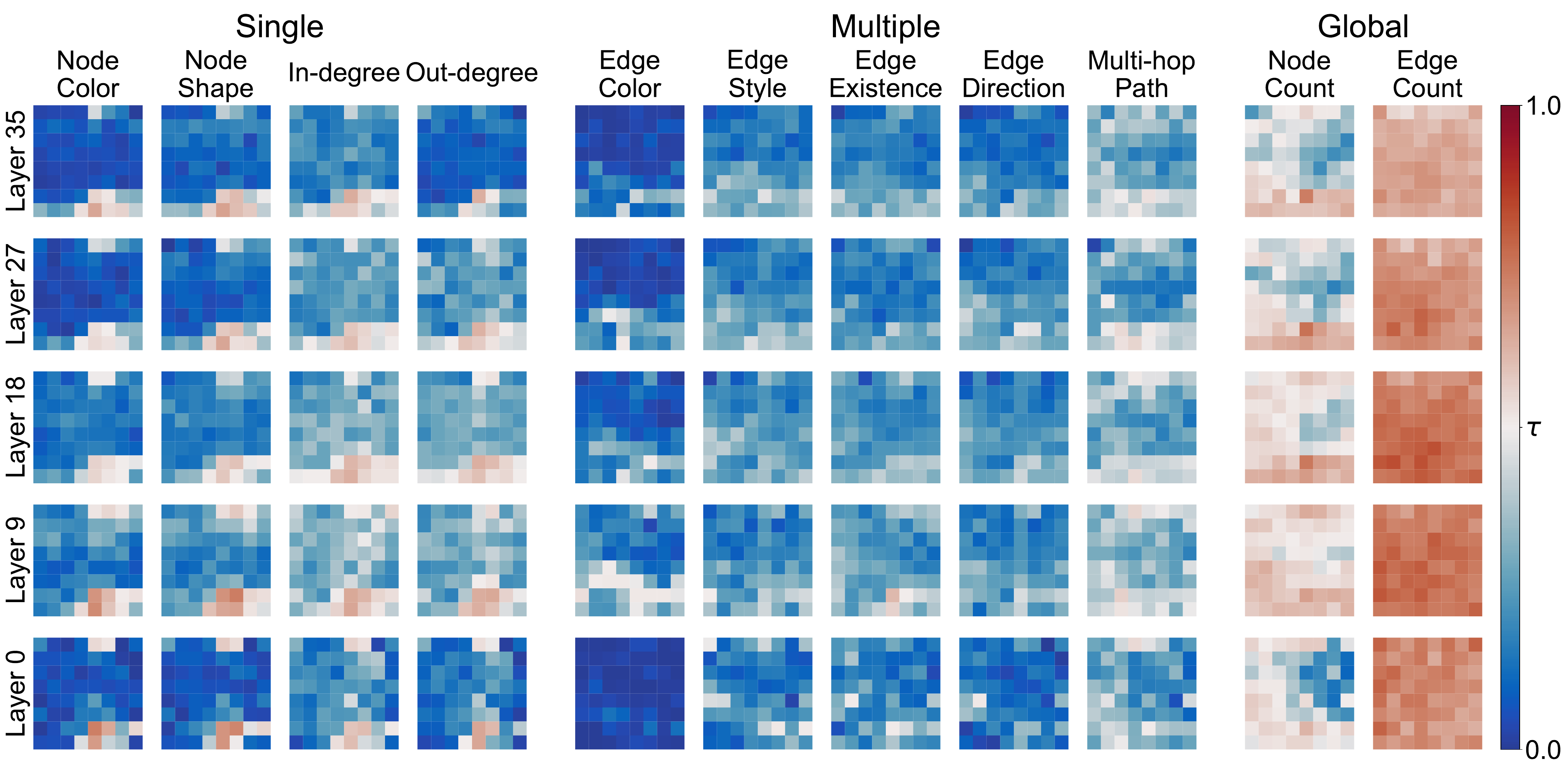}
    \caption{
        Accuracy for each hidden state in the image-input part of the Qwen3-VL 8B language model (7-node graphs).
        Each heatmap shows the accuracy of individual hidden states for a specific layer and aspect.
        Spatial patterns were consistent with the 5-node setting (Figure~\ref{fig:patchwise_image_probing_qwen3_language_model}), although fewer patches exceeded the threshold.
    }
    \label{fig:patchwise_image_probing_qwen3_language_model_7nodes}
\end{figure*}

\begin{figure*}[t]
    \centering
    \includegraphics[width=\linewidth]{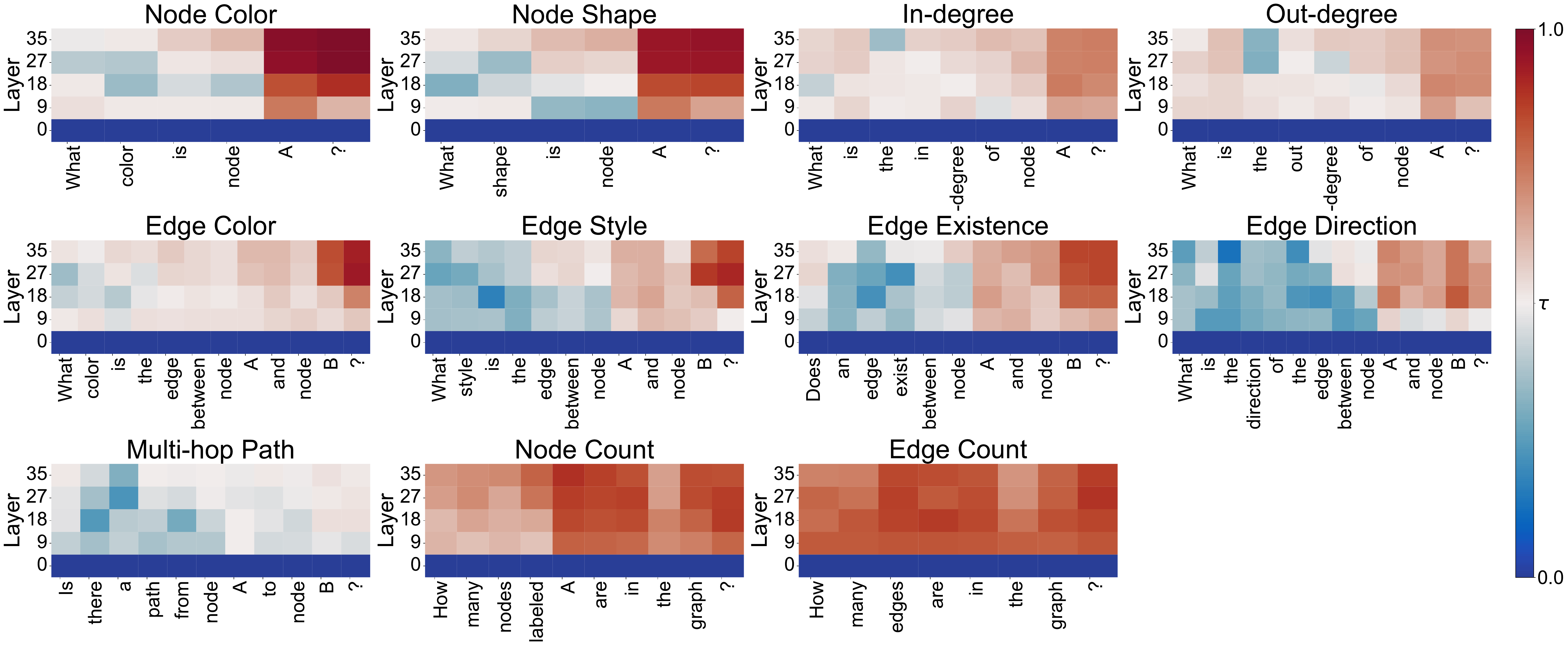}
    \caption{
        Probing results for text tokens in Qwen3-VL 8B (7-node graphs).
        The x-axis shows token position in the question, and the y-axis shows layer position.
        Trends were consistent with the 5-node setting (Figure~\ref{fig:text_probing_qwen3_body}).
    }
    \label{fig:text_probing_qwen3_7nodes}
\end{figure*}

\paragraph{Qwen3-VL 8B (7-node graphs).}
We report probing results on diagrams containing 7 nodes instead of 5 (see Figure~\ref{fig:sample_diagrams_7nodes} for example diagrams).
Figures~\ref{fig:layerwise_image_probing_qwen3_vision_encoder_7nodes} and~\ref{fig:layerwise_image_probing_qwen3_language_model_7nodes} show the layer-wise results, Figures~\ref{fig:patchwise_image_probing_qwen3_vision_encoder_7nodes} and~\ref{fig:patchwise_image_probing_qwen3_language_model_7nodes} show the patch-wise results, and Figure~\ref{fig:text_probing_qwen3_7nodes} shows the text-token results.
The overall accuracy is slightly lower and fewer patches exceed the threshold than in the 5-node setting, possibly reflecting the increased task difficulty from the larger number of nodes.
Nonetheless, the results exhibit the same trends as the default 5-node setting reported in \cref{subsec:probing_results}, indicating that the observed encoding patterns are robust to moderate increases in graph size.

\begin{figure}[t]
    \centering
    \includegraphics[width=\linewidth]{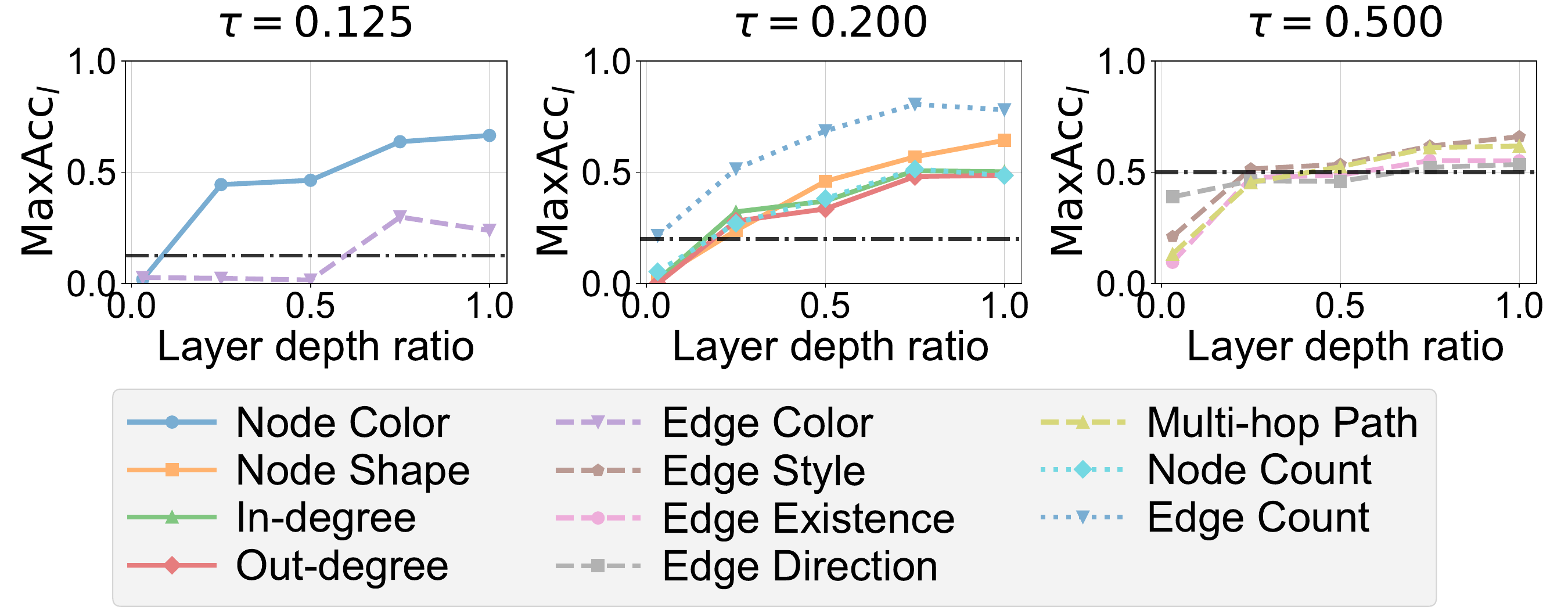}
    \caption{
        $\mathrm{MaxAcc}_{l}$ per layer in the vision encoder of Qwen2.5-VL 7B.
        The x-axis shows relative layer position (0 is the input layer, 1 is the final layer), and the y-axis shows accuracy.
        The black dash-dot line in each panel indicates the threshold $\tau$.
        Trends were similar to Qwen3-VL (Figure~\ref{fig:layerwise_image_probing_qwen3_vision_encoder}): Single- and Global-category aspects rose with depth, while Multiple-category aspects stayed near the threshold.
    }
    \label{fig:layerwise_image_probing_qwen2.5_vision_encoder}
\end{figure}

\begin{figure}[t]
    \centering
    \includegraphics[width=\linewidth]{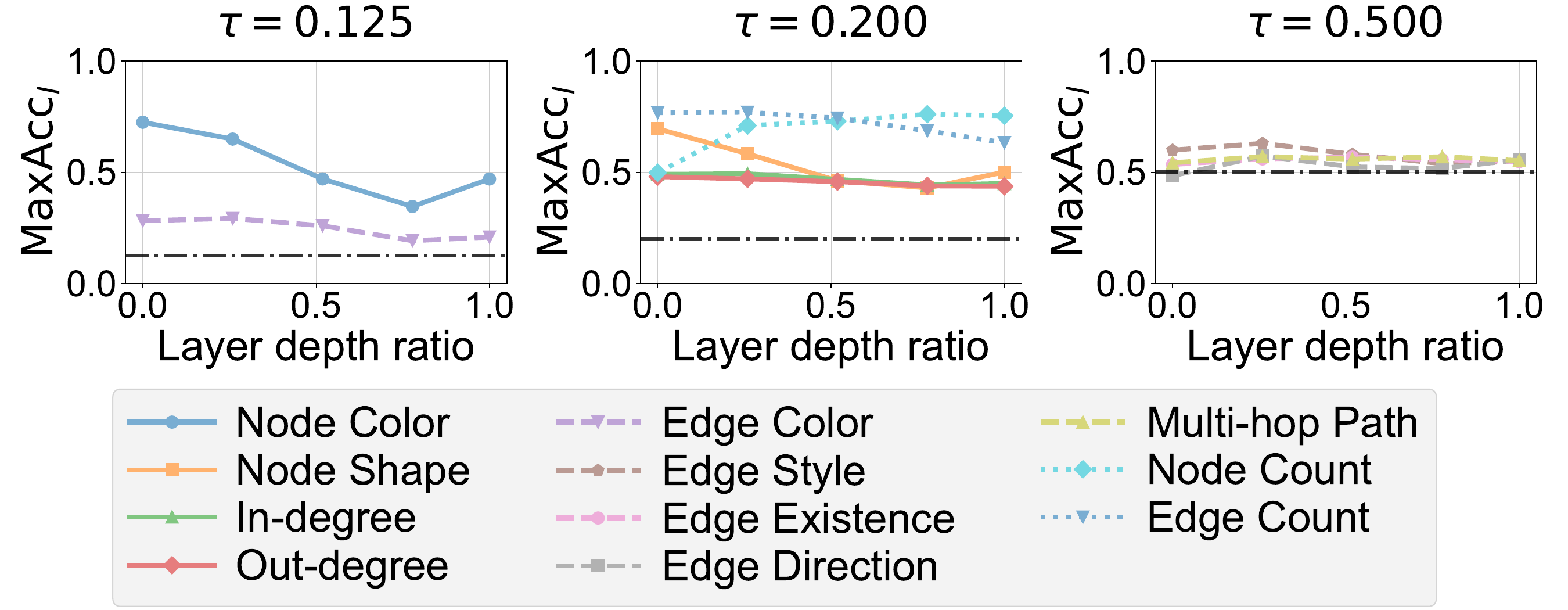}
    \caption{
        $\mathrm{MaxAcc}_{l}$ per layer in the language model of Qwen2.5-VL 7B.
        The x-axis shows relative layer position (0 is the input layer, 1 is the final layer), and the y-axis shows accuracy.
        The black dash-dot line in each panel indicates the threshold $\tau$.
        Trends were similar to Qwen3-VL (Figure~\ref{fig:layerwise_image_probing_qwen3_language_model}).
    }
    \label{fig:layerwise_image_probing_qwen2.5_language_model}
\end{figure}

\begin{figure*}[t]
    \centering
    \includegraphics[width=\linewidth]{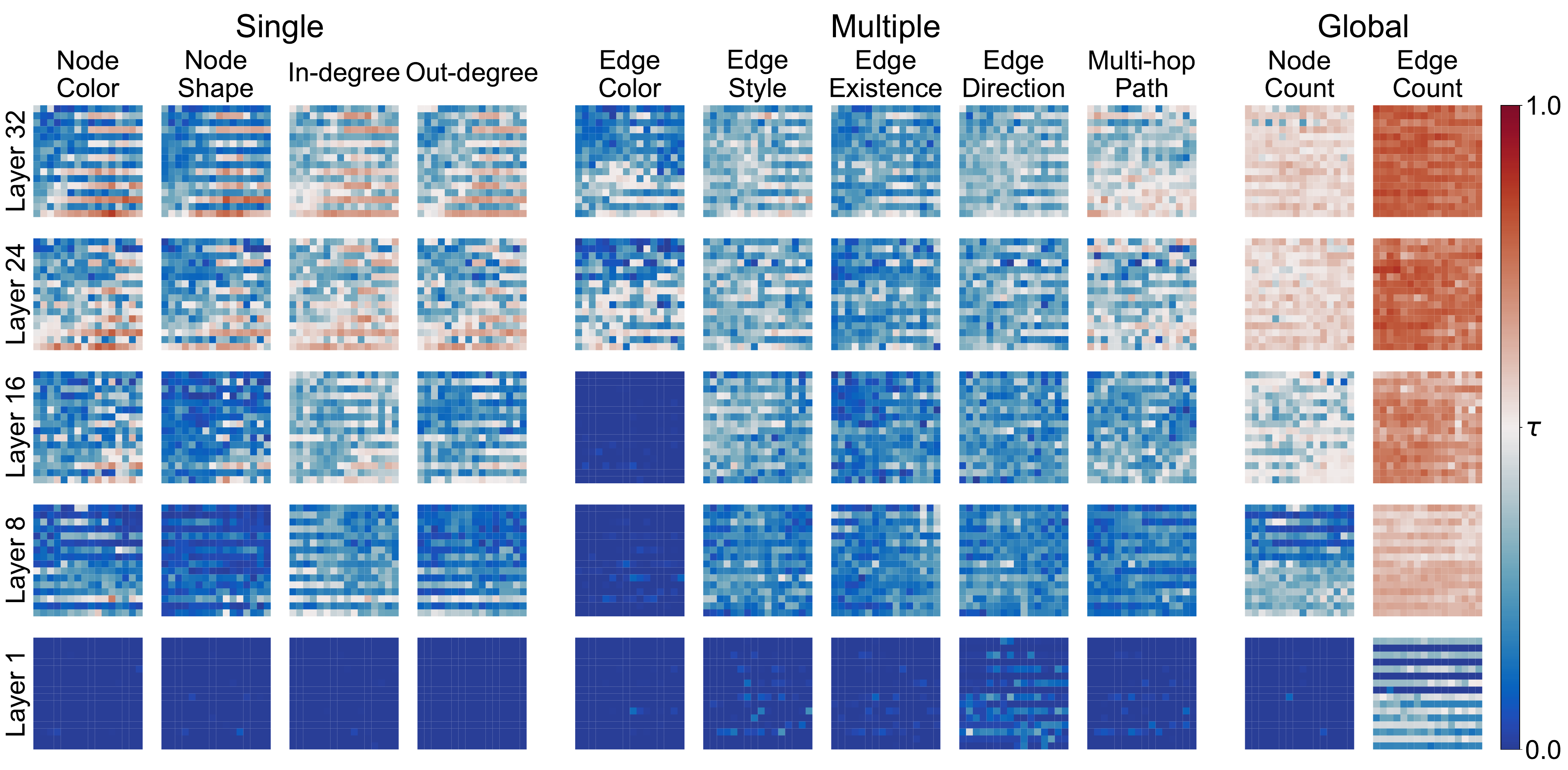}
    \caption{
        Accuracy for each hidden state in the vision encoder of Qwen2.5-VL 7B.
        Each heatmap shows the accuracy of individual hidden states for a specific layer and aspect.
        Spatial patterns were similar to Qwen3-VL (Figure~\ref{fig:patchwise_image_probing_qwen3_vision_encoder}).
    }
    \label{fig:patchwise_image_probing_qwen2.5_vision_encoder}
\end{figure*}

\begin{figure*}[t]
    \centering
    \includegraphics[width=\linewidth]{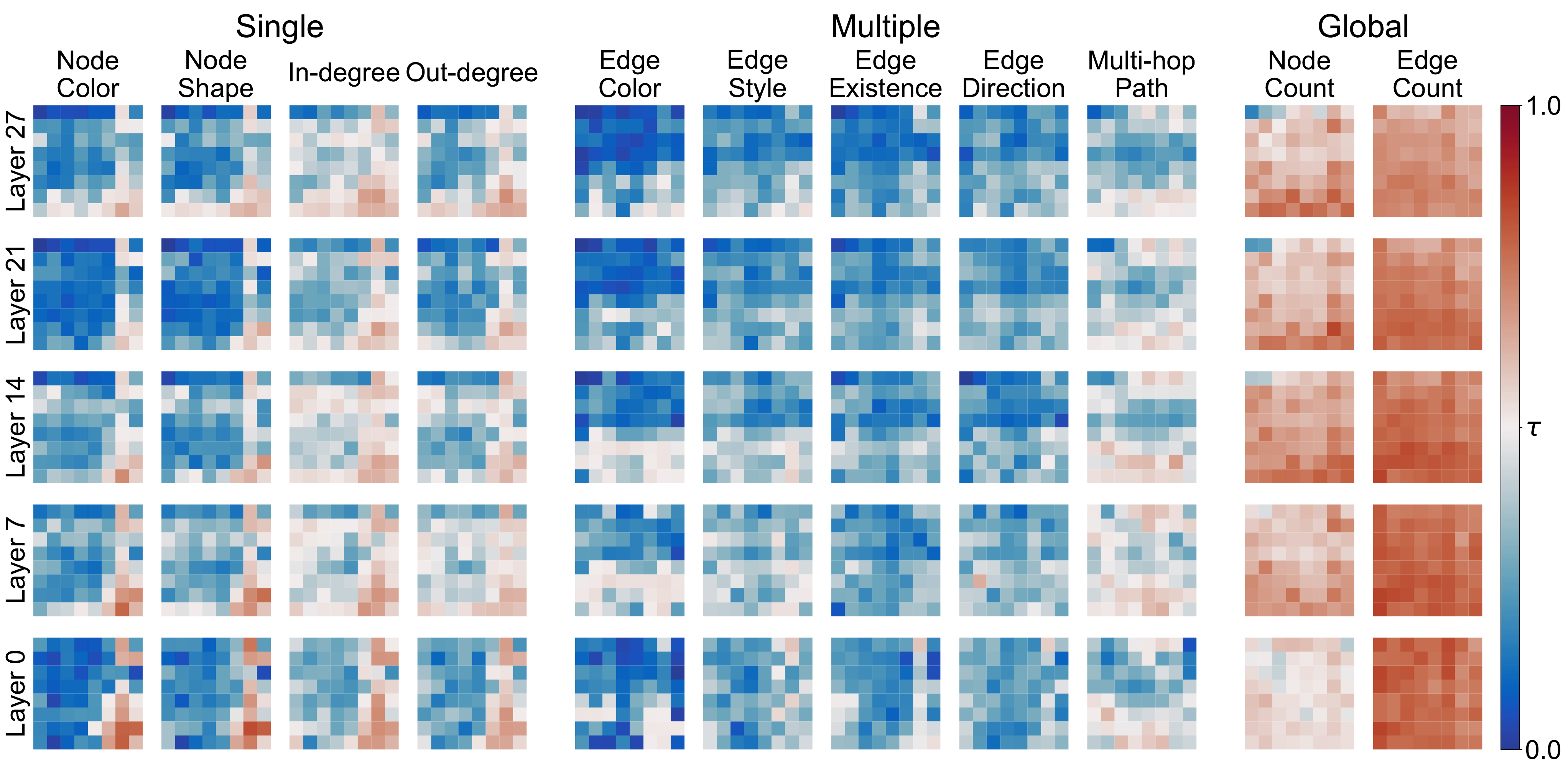}
    \caption{
        Accuracy for each hidden state in the image-input part of the Qwen2.5-VL 7B language model.
        Each heatmap shows the accuracy of individual hidden states for a specific layer and aspect.
        Spatial patterns were similar to Qwen3-VL (Figure~\ref{fig:patchwise_image_probing_qwen3_language_model}).
    }
    \label{fig:patchwise_image_probing_qwen2.5_language_model}
\end{figure*}

\paragraph{Qwen2.5-VL 7B.}
Figures~\ref{fig:layerwise_image_probing_qwen2.5_vision_encoder} and~\ref{fig:layerwise_image_probing_qwen2.5_language_model} show the layer-wise probing results, and Figures~\ref{fig:patchwise_image_probing_qwen2.5_vision_encoder} and~\ref{fig:patchwise_image_probing_qwen2.5_language_model} show the patch-wise results for Qwen2.5-VL 7B.
The overall trends are similar to Qwen3-VL, with node information showing high accuracy while edge relationships remained near chance level.
In the patch-wise analysis, patches corresponding to node positions showed higher accuracy for aspects in the Single category, while aspects in the Global category showed high accuracy across a wide range of positions, including background regions.

\begin{figure*}[t]
    \centering
    \includegraphics[width=\linewidth]{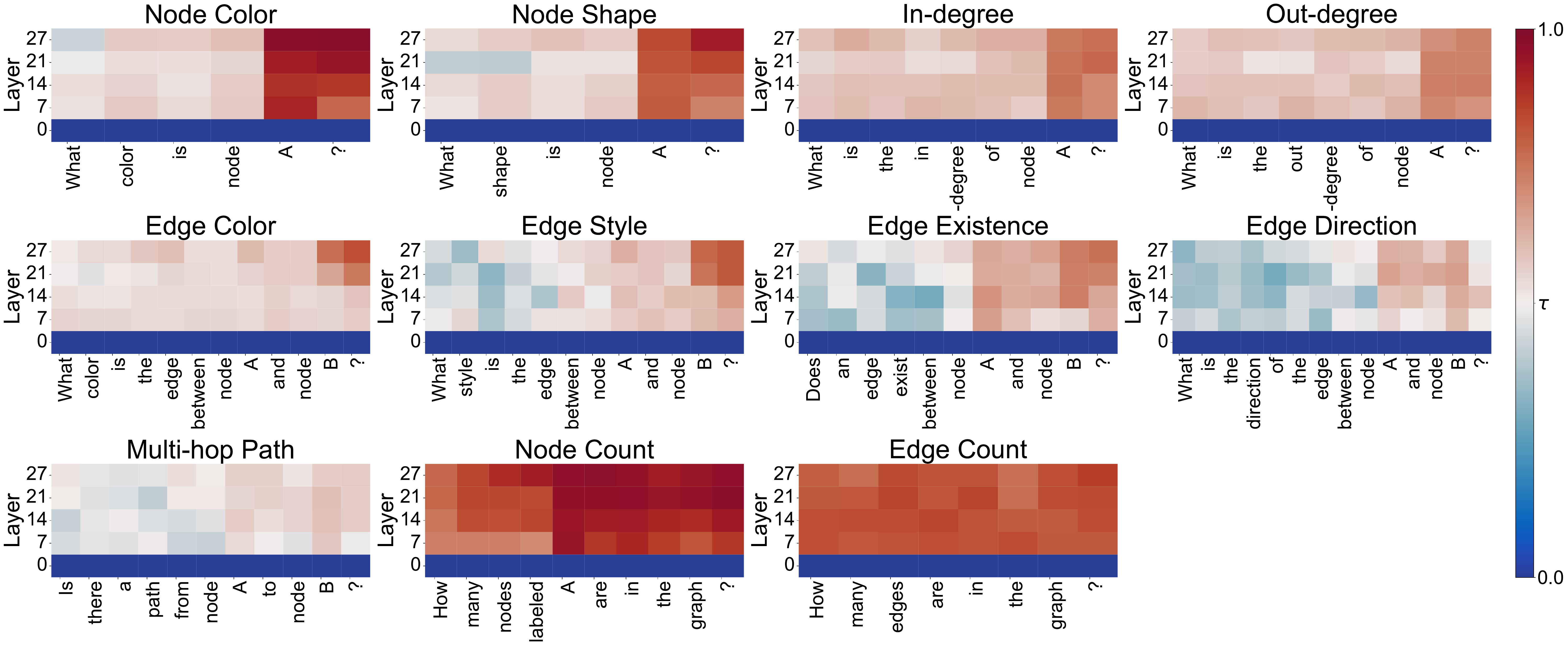}
    \caption{
        Probing results for text tokens in Qwen2.5-VL 7B.
        The x-axis shows token position in the question, and the y-axis shows layer position.
        Trends were similar to Qwen3-VL (Figure~\ref{fig:text_probing_qwen3_body}).
    }
    \label{fig:text_probing_qwen2.5}
\end{figure*}

Figure~\ref{fig:text_probing_qwen2.5} shows the probing results for text tokens.
Similar to Qwen3-VL, we observed a trend where accuracy sharply increased at token positions that specify nodes or edges.

\begin{figure}[t]
    \centering
    \includegraphics[width=\linewidth]{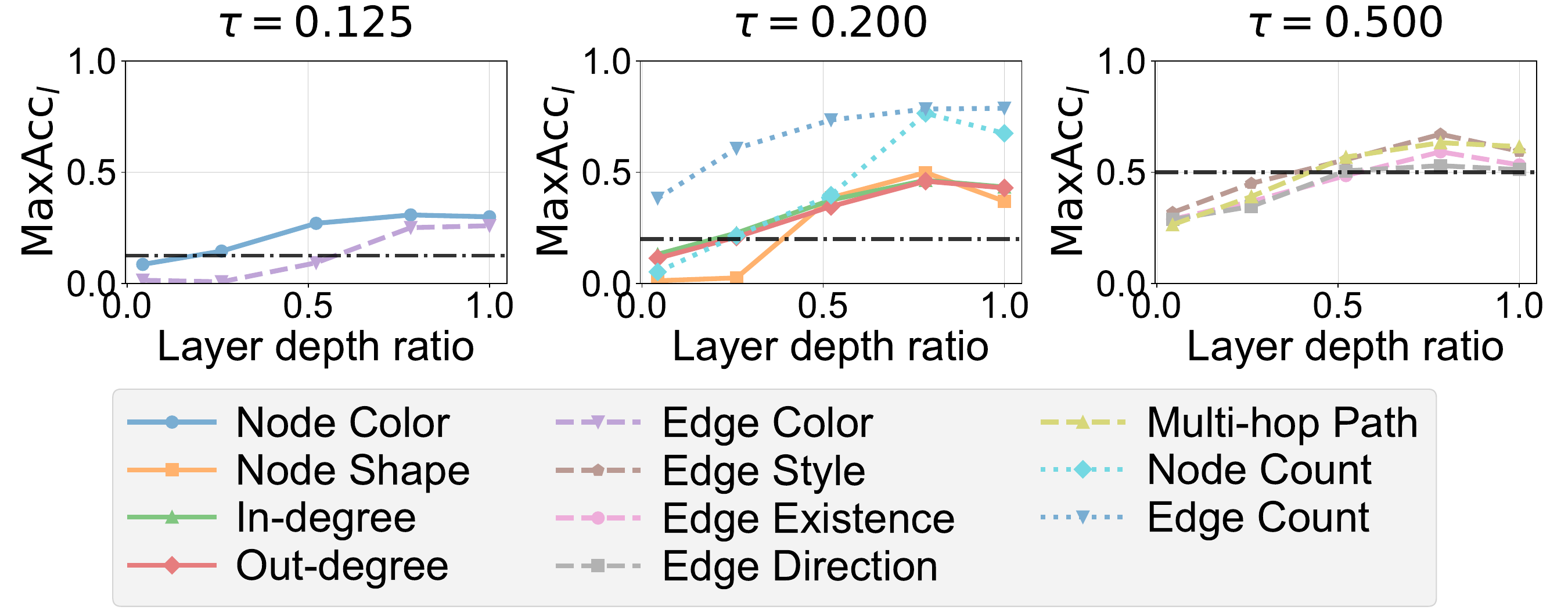}
    \caption{
        $\mathrm{MaxAcc}_{l}$ per layer in the vision encoder of LLaVA1.5 7B.
        The x-axis shows relative layer position (0 is the input layer, 1 is the final layer), and the y-axis shows accuracy.
        The black dash-dot line in each panel indicates the threshold $\tau$.
        Trends were similar to Qwen3-VL (Figure~\ref{fig:layerwise_image_probing_qwen3_vision_encoder}).
    }
    \label{fig:layerwise_image_probing_llava1.5_vision_encoder}
\end{figure}

\begin{figure}[t]
    \centering
    \includegraphics[width=\linewidth]{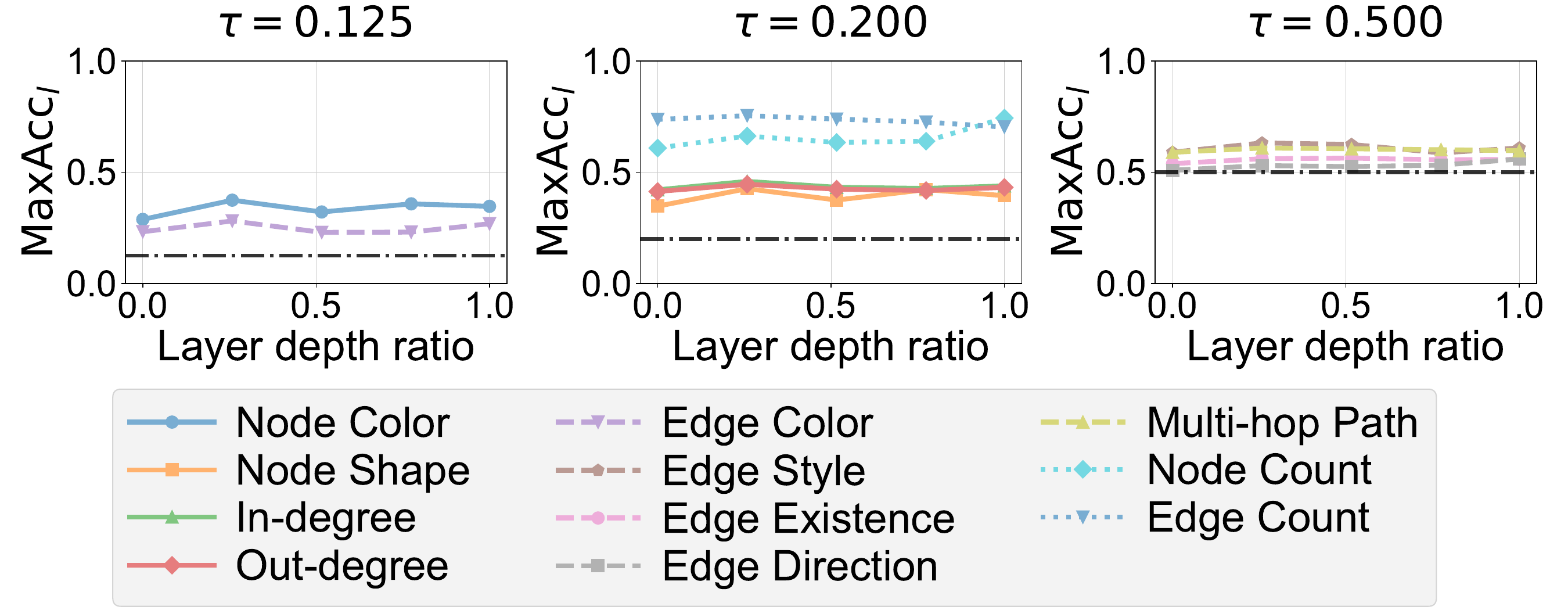}
    \caption{
        $\mathrm{MaxAcc}_{l}$ per layer in the language model of LLaVA1.5 7B.
        The x-axis shows relative layer position (0 is the input layer, 1 is the final layer), and the y-axis shows accuracy.
        The black dash-dot line in each panel indicates the threshold $\tau$.
        Trends were similar to Qwen3-VL (Figure~\ref{fig:layerwise_image_probing_qwen3_language_model}).
    }
    \label{fig:layerwise_image_probing_llava1.5_language_model}
\end{figure}

\begin{figure*}[t]
    \centering
    \includegraphics[width=\linewidth]{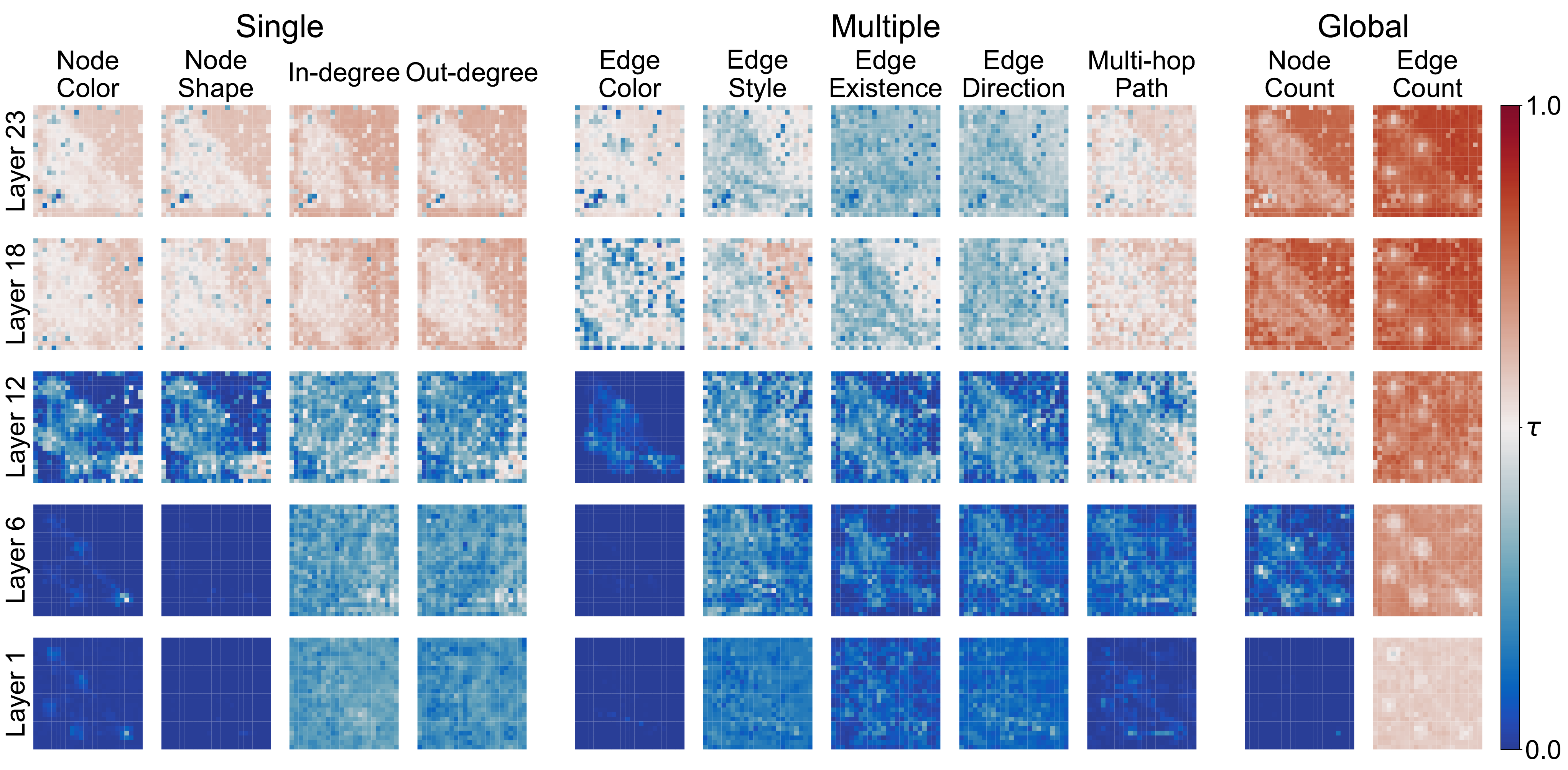}
    \caption{
        Accuracy for each hidden state in the vision encoder of LLaVA1.5 7B.
        Each heatmap shows the accuracy of individual hidden states for a specific layer and aspect.
        Unlike Qwen3-VL (Figure~\ref{fig:patchwise_image_probing_qwen3_vision_encoder}), background patches showed higher accuracy than patches corresponding to node positions.
    }
    \label{fig:patchwise_image_probing_llava1.5_vision_encoder}
\end{figure*}

\begin{figure*}[t]
    \centering
    \includegraphics[width=\linewidth]{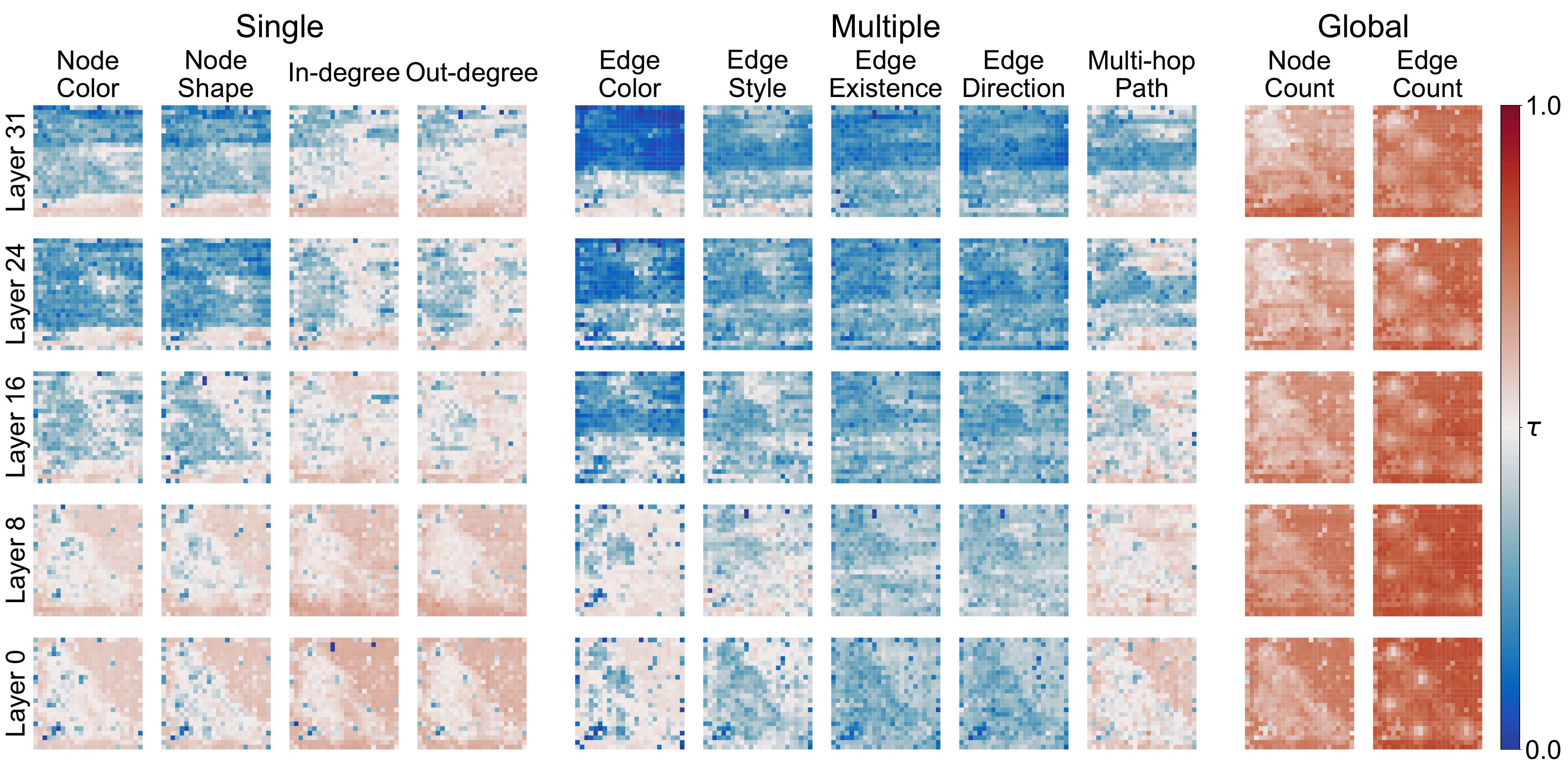}
    \caption{
        Accuracy for each hidden state in the image-input part of the LLaVA1.5 7B language model.
        Each heatmap shows the accuracy of individual hidden states for a specific layer and aspect.
        Unlike Qwen3-VL (Figure~\ref{fig:patchwise_image_probing_qwen3_language_model}), background patches showed higher accuracy than patches corresponding to node positions.
    }
    \label{fig:patchwise_image_probing_llava1.5_language_model}
\end{figure*}

\paragraph{LLaVA1.5 7B.}
Figures~\ref{fig:layerwise_image_probing_llava1.5_vision_encoder} and~\ref{fig:layerwise_image_probing_llava1.5_language_model} show the layer-wise probing results, and Figures~\ref{fig:patchwise_image_probing_llava1.5_vision_encoder} and~\ref{fig:patchwise_image_probing_llava1.5_language_model} show the patch-wise results for LLaVA1.5 7B.
The overall trends are similar to Qwen3-VL and Qwen2.5-VL, but a distinctive difference was observed in the patch-wise analysis.
In LLaVA1.5, background patches showed higher accuracy than patches corresponding to node positions.
The background of our diagrams is uniformly white and carries almost no information of its own.
The corresponding hidden states may therefore be available to store information from other positions, as discussed for aspects in the Global category in \cref{subsec:probing_results}.
In LLaVA1.5 this tendency extends to aspects in the Single category, whose signals are localized near node positions in the other models.

\begin{figure*}[t]
    \centering
    \includegraphics[width=\linewidth]{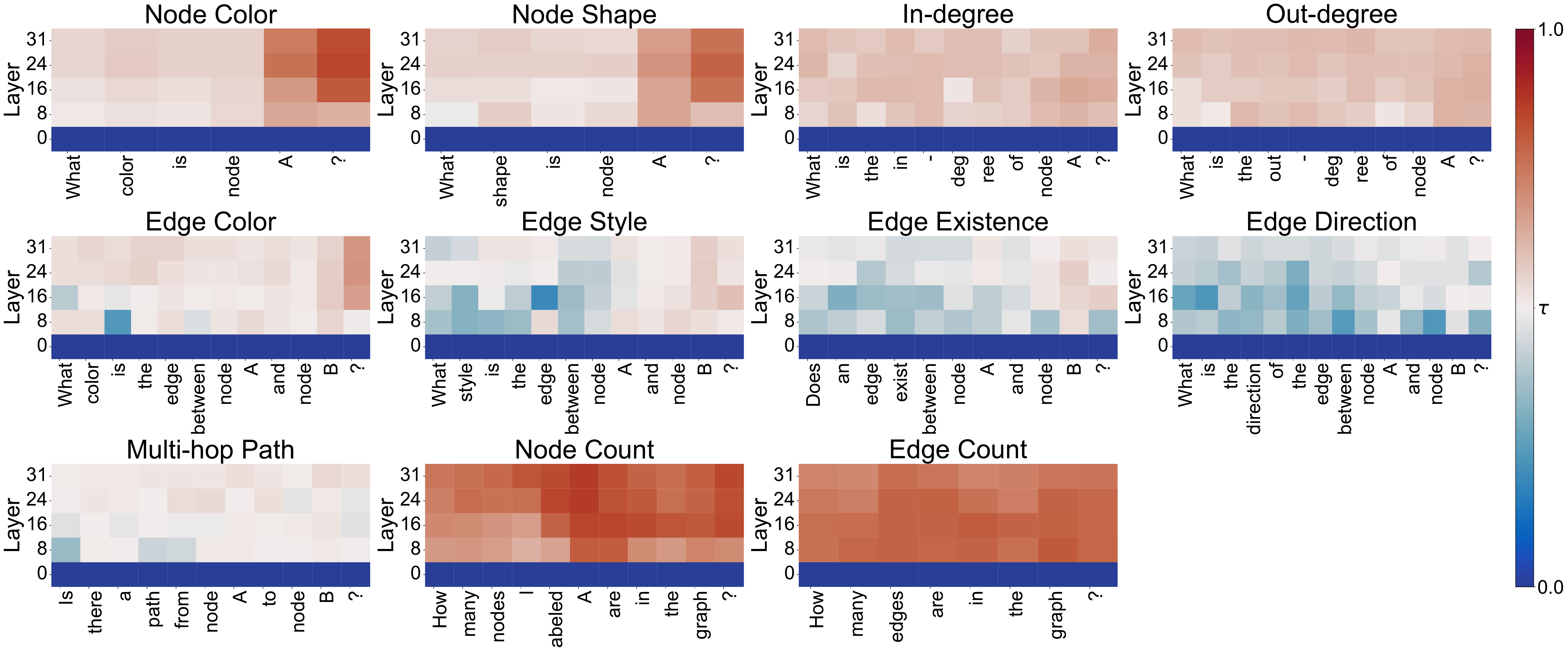}
    \caption{
        Probing results for text tokens in LLaVA1.5 7B.
        The x-axis shows token position in the question, and the y-axis shows layer position.
        Trends were broadly similar to Qwen3-VL (Figure~\ref{fig:text_probing_qwen3_body}), though less pronounced.
    }
    \label{fig:text_probing_llava1.5}
\end{figure*}

Figure~\ref{fig:text_probing_llava1.5} shows the probing results for text tokens.
Similar to other models, we observed information aggregation at token positions that specify target nodes or edges.

\begin{figure}[t]
    \centering
    \includegraphics[width=\linewidth]{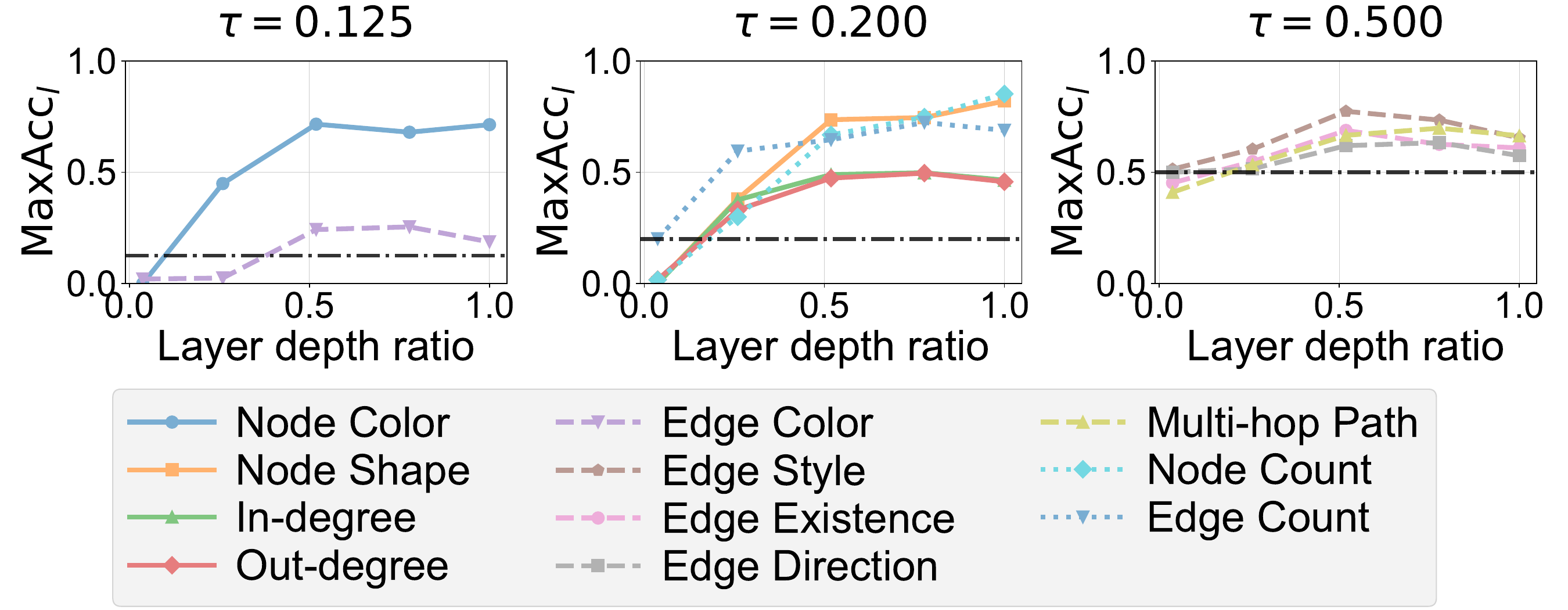}
    \caption{
        $\mathrm{MaxAcc}_{l}$ per layer in the vision encoder of Gemma3-4B-IT.
        The x-axis shows relative layer position (0 is the input layer, 1 is the final layer), and the y-axis shows accuracy.
        The black dash-dot line in each panel indicates the threshold $\tau$.
        Trends were similar to Qwen3-VL (Figure~\ref{fig:layerwise_image_probing_qwen3_vision_encoder}), though accuracy was lower for some aspects.
    }
    \label{fig:layerwise_image_probing_gemma3_vision_encoder}
\end{figure}

\begin{figure}[t]
    \centering
    \includegraphics[width=\linewidth]{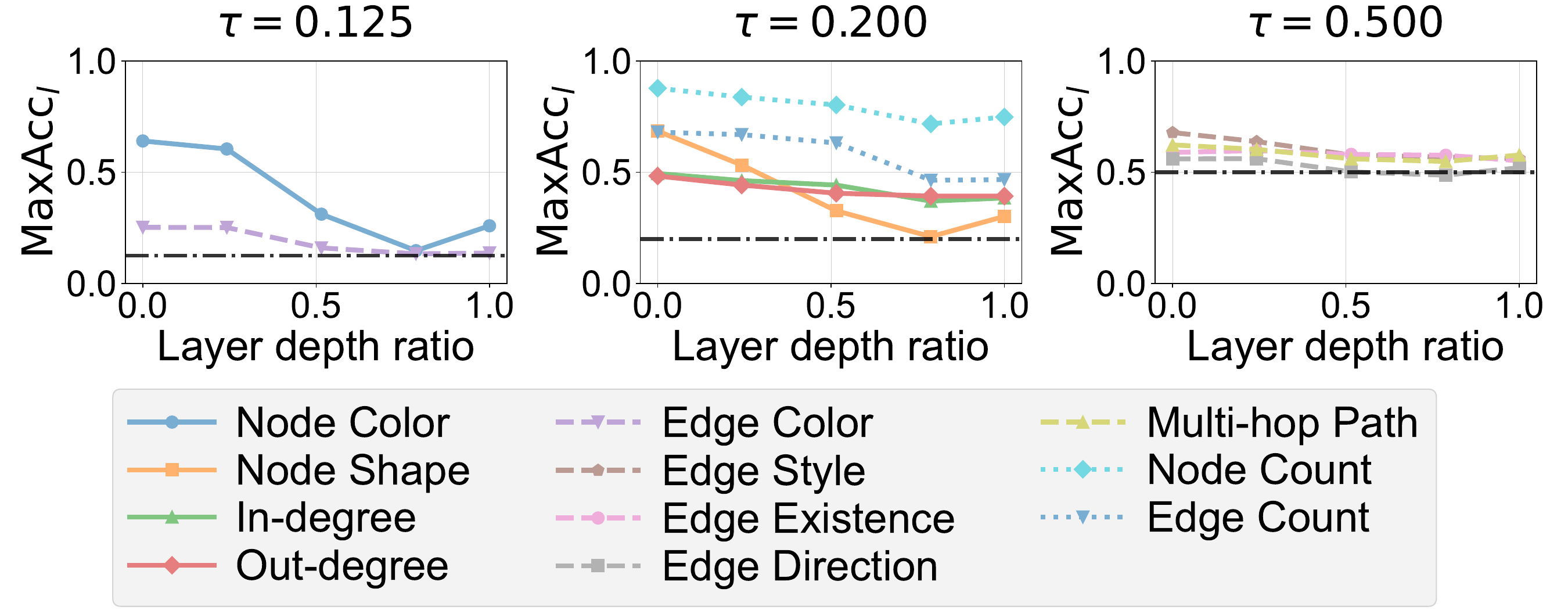}
    \caption{
        $\mathrm{MaxAcc}_{l}$ per layer in the language model of Gemma3-4B-IT.
        The x-axis shows relative layer position (0 is the input layer, 1 is the final layer), and the y-axis shows accuracy.
        The black dash-dot line in each panel indicates the threshold $\tau$.
        Trends were similar to Qwen3-VL (Figure~\ref{fig:layerwise_image_probing_qwen3_language_model}), though accuracy was lower for some aspects.
    }
    \label{fig:layerwise_image_probing_gemma3_language_model}
\end{figure}

\begin{figure*}[t]
    \centering
    \includegraphics[width=\linewidth]{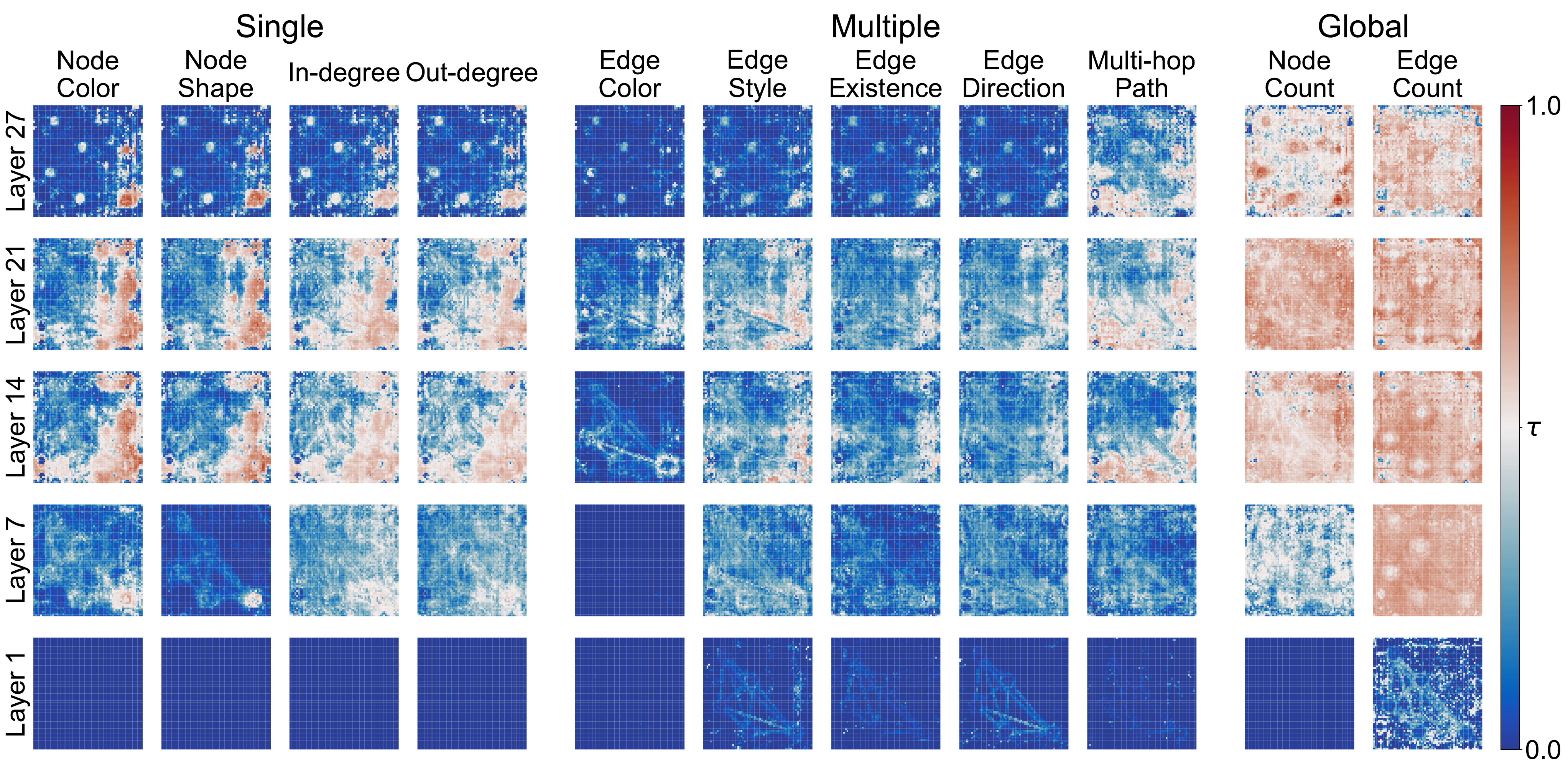}
    \caption{
        Accuracy for each hidden state in the vision encoder of Gemma3-4B-IT.
        Each heatmap shows the accuracy of individual hidden states for a specific layer and aspect.
        Spatial patterns were similar to Qwen3-VL (Figure~\ref{fig:patchwise_image_probing_qwen3_vision_encoder}), though accuracy was lower for some aspects.
    }
    \label{fig:patchwise_image_probing_gemma3_vision_encoder}
\end{figure*}

\begin{figure*}[t]
    \centering
    \includegraphics[width=\linewidth]{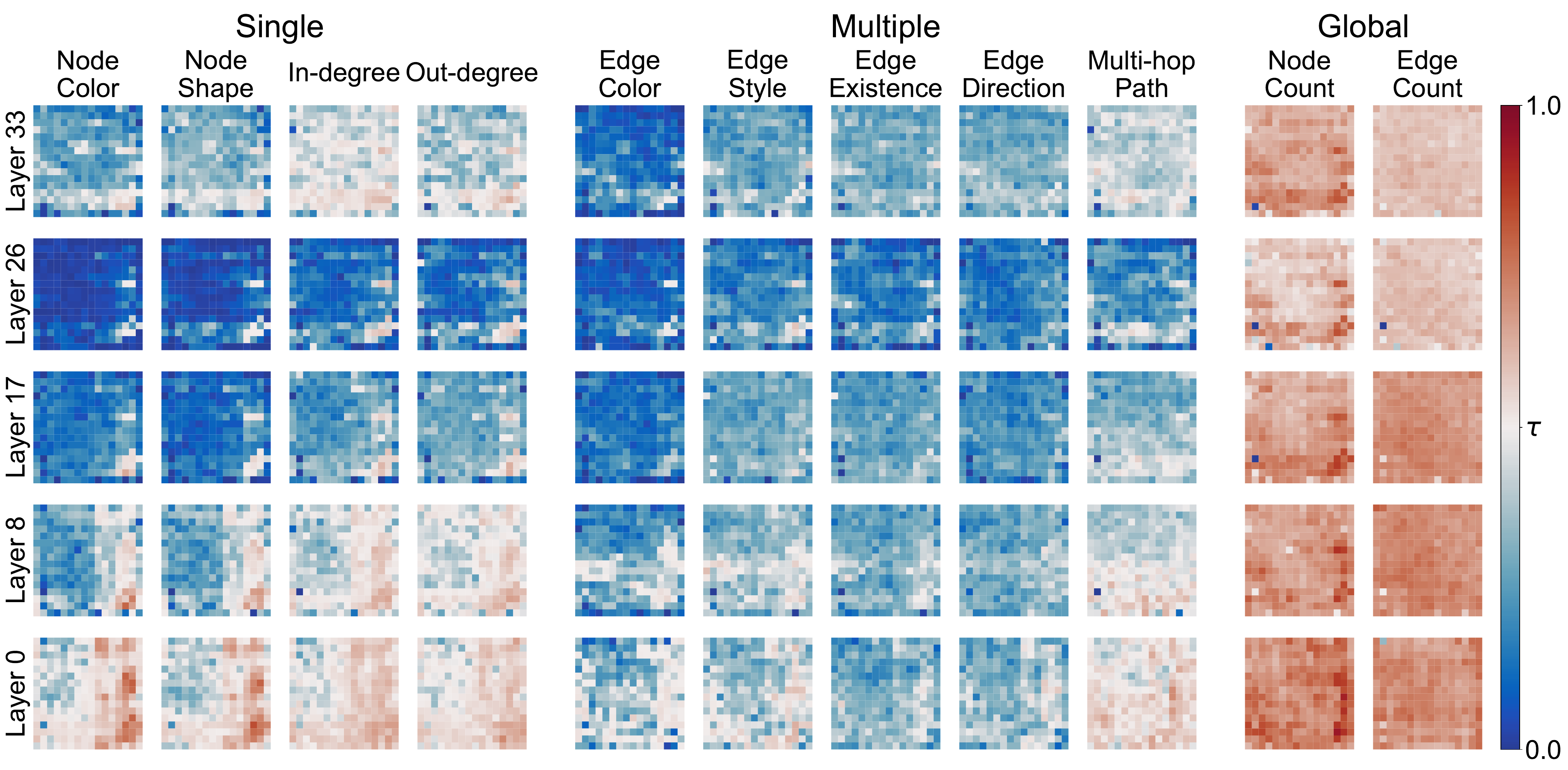}
    \caption{
        Accuracy for each hidden state in the image-input part of the Gemma3-4B-IT language model.
        Each heatmap shows the accuracy of individual hidden states for a specific layer and aspect.
        Spatial patterns were similar to Qwen3-VL (Figure~\ref{fig:patchwise_image_probing_qwen3_language_model}), though accuracy was lower for some aspects.
    }
    \label{fig:patchwise_image_probing_gemma3_language_model}
\end{figure*}

\paragraph{Gemma3-4B-IT.}
Figures~\ref{fig:layerwise_image_probing_gemma3_vision_encoder} and~\ref{fig:layerwise_image_probing_gemma3_language_model} show the layer-wise probing results, and Figures~\ref{fig:patchwise_image_probing_gemma3_vision_encoder} and~\ref{fig:patchwise_image_probing_gemma3_language_model} show the patch-wise results for Gemma3-4B-IT.
The overall trends are similar to Qwen3-VL and Qwen2.5-VL, but accuracy was lower for some node information and edge relationships compared to other models.

\begin{figure*}[t]
    \centering
    \includegraphics[width=\linewidth]{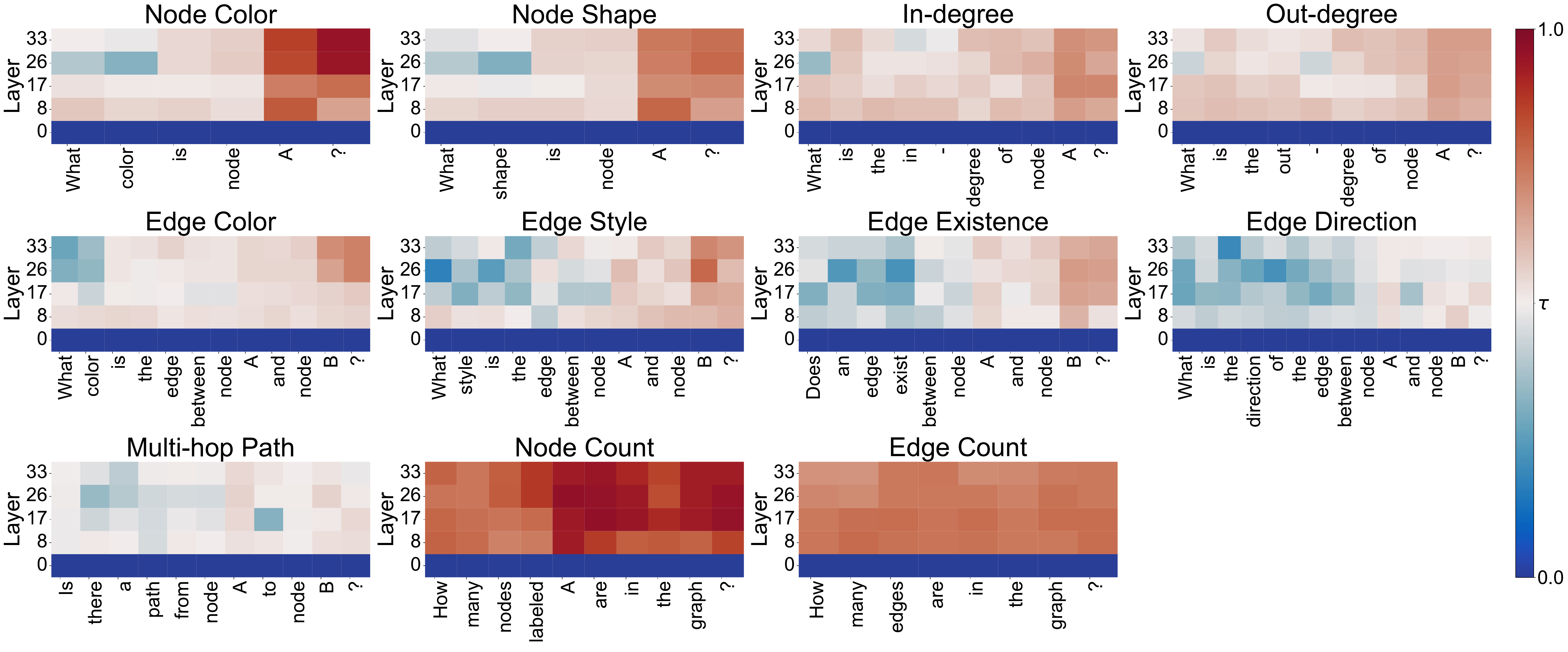}
    \caption{
        Probing results for text tokens in Gemma3-4B-IT.
        The x-axis shows token position in the question, and the y-axis shows layer position.
        Trends were broadly similar to Qwen3-VL (Figure~\ref{fig:text_probing_qwen3_body}), though less pronounced.
    }
    \label{fig:text_probing_gemma3}
\end{figure*}

Figure~\ref{fig:text_probing_gemma3} shows the probing results for text tokens.
As with the other models, we observe an aggregation of information at the token positions that specify the target nodes or edges; however, the effect is weaker than in the other models.

\section{Additional Results for Causal Intervention}
\label{sec:appendix_intervention}
In \cref{sec:causal_intervention}, we reported the causal intervention results as the accuracy difference between the \textit{accuracy-based} and \textit{controlled} settings for Qwen3-VL 8B.
Here, we additionally report the absolute accuracies for the \textit{accuracy-based}, \textit{controlled}, and \textit{clean} settings, where \textit{clean} denotes the accuracy obtained without any intervention, and extend the analysis to Qwen2.5-VL 7B, LLaVA1.5 7B, and Gemma3-4B-IT.

\paragraph{Selective intervention for \texttt{Edge Count}.}
As described in \cref{sec:causal_intervention}, for \texttt{Edge Count} the probing accuracy exceeds the threshold at all positions in most layers, so the \textit{accuracy-based} and \textit{controlled} settings coincide and the causal contribution cannot be assessed.
We therefore apply a selective intervention that targets only the top 50\% of patches by probing accuracy.
Table~\ref{tab:selective_intervention_edge_count} reports the results.
Intervening on patches with high probing accuracy lowered VQA accuracy by 14.9 to 21.7 points compared with intervening on the same number of randomly selected patches.
The only exception is the final layer of the language model; this exception is consistent with the late-layer trend observed for the other aspects in the main text.
These results confirm that, for \texttt{Edge Count} as well, the signals detected by probing causally contribute to the model's predictions.

\begin{table*}[t]
    \small
    \centering
    \begin{tabular}{llrrr}
        \toprule
        Component & Layer & Accuracy-based (\%) & Controlled (\%) & Diff. \\
        \midrule
        \multirow{3}{*}{Vision encoder} & 1 & 25.6 & 42.4 & $-$16.8 \\
        & 14 & 51.6 & 66.5 & $-$14.9 \\
        & 27 & 50.4 & 72.1 & $-$21.7 \\
        \midrule
        \multirow{3}{*}{Language model} & 0 & 30.7 & 51.8 & $-$21.2 \\
        & 18 & 54.8 & 75.8 & $-$20.9 \\
        & 35 & 83.9 & 83.8 & $+$0.1 \\
        \bottomrule
    \end{tabular}
    \caption{
        VQA accuracy for \texttt{Edge Count} under the selective (top-50\%) intervention on Qwen3-VL 8B, comparing the \textit{accuracy-based} setting (intervening on the highest-probing-accuracy patches) with the \textit{controlled} setting (intervening on the same number of randomly selected patches).
        Values are averaged over the evaluation seeds.
    }
    \label{tab:selective_intervention_edge_count}
\end{table*}

\begin{figure*}[t]
    \centering
    \includegraphics[width=\linewidth]{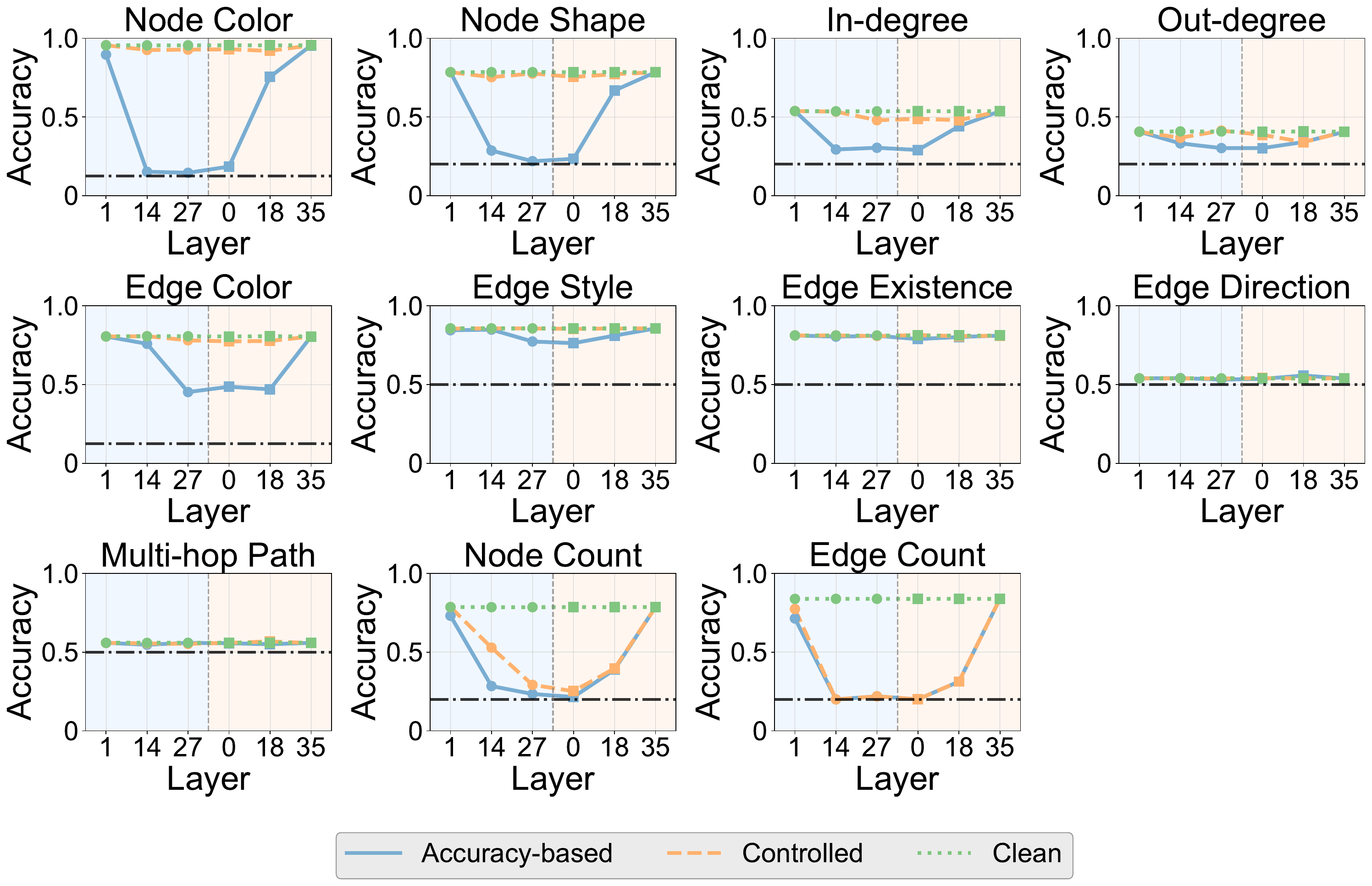}
    \caption{
        \textit{Clean}, \textit{accuracy-based}, and \textit{controlled} accuracy at each intervention layer for Qwen3-VL 8B.
        The background colors of the plots indicate the results for \colorbox[RGB]{242,247,254}{the vision encoder} and \colorbox[RGB]{254,247,241}{the language model}, respectively.
    }
    \label{fig:causal_intervention_qwen3_plot}
\end{figure*}

\paragraph{Qwen3-VL 8B.}
Figure~\ref{fig:causal_intervention_qwen3_plot} shows the \textit{clean}, \textit{accuracy-based}, and \textit{controlled} accuracy at each intervention layer for Qwen3-VL 8B.
The results are discussed in detail in \cref{sec:causal_intervention}.

\begin{figure*}[t]
    \centering
    \includegraphics[width=\linewidth]{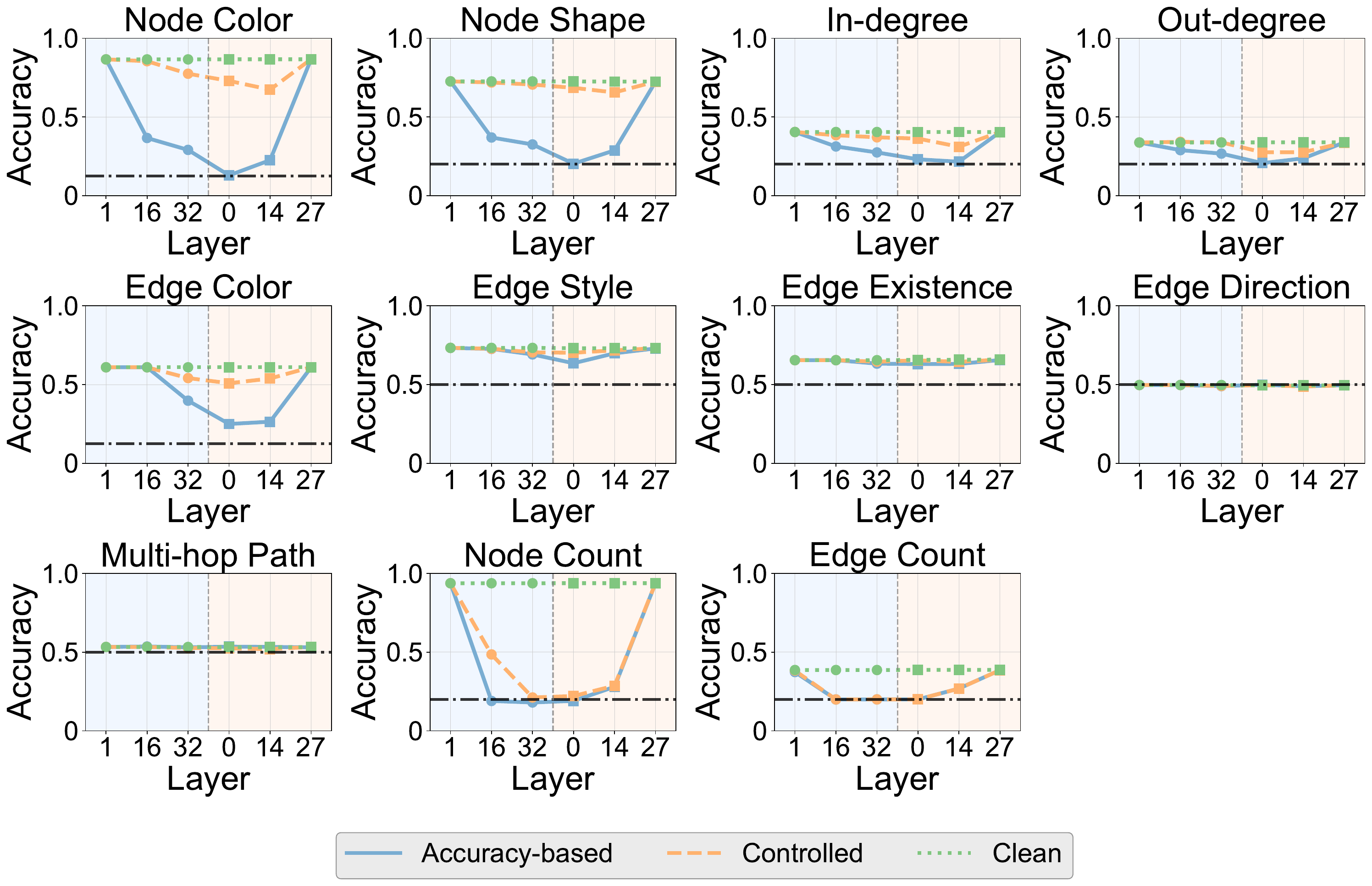}
    \caption{
        \textit{Clean}, \textit{accuracy-based}, and \textit{controlled} accuracy at each intervention layer for Qwen2.5-VL 7B.
        The background colors of the plots indicate the results for \colorbox[RGB]{242,247,254}{the vision encoder} and \colorbox[RGB]{254,247,241}{the language model}, respectively.
    }
    \label{fig:causal_intervention_qwen2.5_plot}
\end{figure*}

\paragraph{Qwen2.5-VL 7B.}
Table~\ref{tab:causal_intervention_diff_qwen2.5} and Figure~\ref{fig:causal_intervention_qwen2.5_plot} report the results for Qwen2.5-VL 7B.
Similar to Qwen3-VL, the intervention caused a large drop in accuracy for aspects such as \texttt{Node Color}, \texttt{Node Shape}, \texttt{Edge Color}, and \texttt{Node Count}, indicating that the signals detected by probing in both the vision encoder and the language model causally contribute to the model's predictions.

\begin{table*}[t]
    \tabcolsep 1mm
    \small
    \centering
    \begin{tabular}{lcrrrrrrrrrrr}
        \toprule
        & Layer
        & \makecell{\texttt{Node}\\\texttt{Color}}
        & \makecell{\texttt{Node}\\\texttt{Shape}}
        & \makecell{\texttt{In-deg.}}
        & \makecell{\texttt{Out-deg.}}
        & \makecell{\texttt{Edge}\\\texttt{Color}}
        & \makecell{\texttt{Edge}\\\texttt{Style}}
        & \makecell{\texttt{Edge}\\\texttt{Exist.}}
        & \makecell{\texttt{Edge}\\\texttt{Direct.}}
        & \makecell{\texttt{Multi-hop}\\\texttt{Path}}
        & \makecell{\texttt{Node}\\\texttt{Count}}
        & \makecell{\texttt{Edge}\\\texttt{Count}} \\
        \midrule
        \multirow{3}{*}{\makecell[l]{Vision\\encoder}}
        & 1
        & $\pm$0.0
        & $\pm$0.0
        & $\pm$0.0
        & $\pm$0.0
        & $\pm$0.0
        & $\pm$0.0
        & $\pm$0.0
        & $\pm$0.0
        & $\pm$0.0
        & $\pm$0.0
        & \textbf{$-$0.9} \\
        & 16
        & \textbf{$-$48.8}
        & \textbf{$-$34.9}
        & \textbf{$-$7.1}
        & \textbf{$-$5.3}
        & $\pm$0.0
        & \textbf{$-$0.1}
        & $\pm$0.0
        & $\pm$0.0
        & $+$0.3
        & \textbf{$-$29.6}
        & $\pm$0.0 \\
        & 32
        & \textbf{$-$48.4}
        & \textbf{$-$38.0}
        & \textbf{$-$9.6}
        & \textbf{$-$7.2}
        & \textbf{$-$14.3}
        & \textbf{$-$1.3}
        & \textbf{$-$1.2}
        & \textbf{$-$0.1}
        & $+$0.4
        & \textbf{$-$3.2}
        & $\pm$0.0 \\
        \midrule
        \multirow{3}{*}{\makecell[l]{Language\\model}}
        & 0
        & \textbf{$-$60.0}
        & \textbf{$-$48.2}
        & \textbf{$-$13.2}
        & \textbf{$-$6.9}
        & \textbf{$-$25.9}
        & \textbf{$-$6.7}
        & \textbf{$-$2.3}
        & $+$0.3
        & $+$1.1
        & \textbf{$-$3.1}
        & $\pm$0.0 \\
        & 14
        & \textbf{$-$44.8}
        & \textbf{$-$36.8}
        & \textbf{$-$9.5}
        & \textbf{$-$3.9}
        & \textbf{$-$27.4}
        & \textbf{$-$1.9}
        & \textbf{$-$1.3}
        & $+$0.2
        & $+$1.4
        & \textbf{$-$0.7}
        & $\pm$0.0 \\
        & 27
        & $\pm$0.0
        & \textbf{$-$0.1}
        & $\pm$0.0
        & $+$0.1
        & \textbf{$-$0.2}
        & \textbf{$-$0.1}
        & \textbf{$-$0.2}
        & $+$0.1
        & \textbf{$-$0.3}
        & \textbf{$-$0.1}
        & $\pm$0.0 \\
        \bottomrule
    \end{tabular}
    \caption{
        Accuracy difference between \textit{accuracy-based} and \textit{controlled} settings (\textit{accuracy-based} $-$ \textit{controlled}) for Qwen2.5-VL 7B.
        More negative values (\textbf{bold}) indicate a larger accuracy decrease under the \textit{accuracy-based} intervention than under the \textit{controlled} intervention.
    }
    \label{tab:causal_intervention_diff_qwen2.5}
\end{table*}

\begin{figure*}[t]
    \centering
    \includegraphics[width=\linewidth]{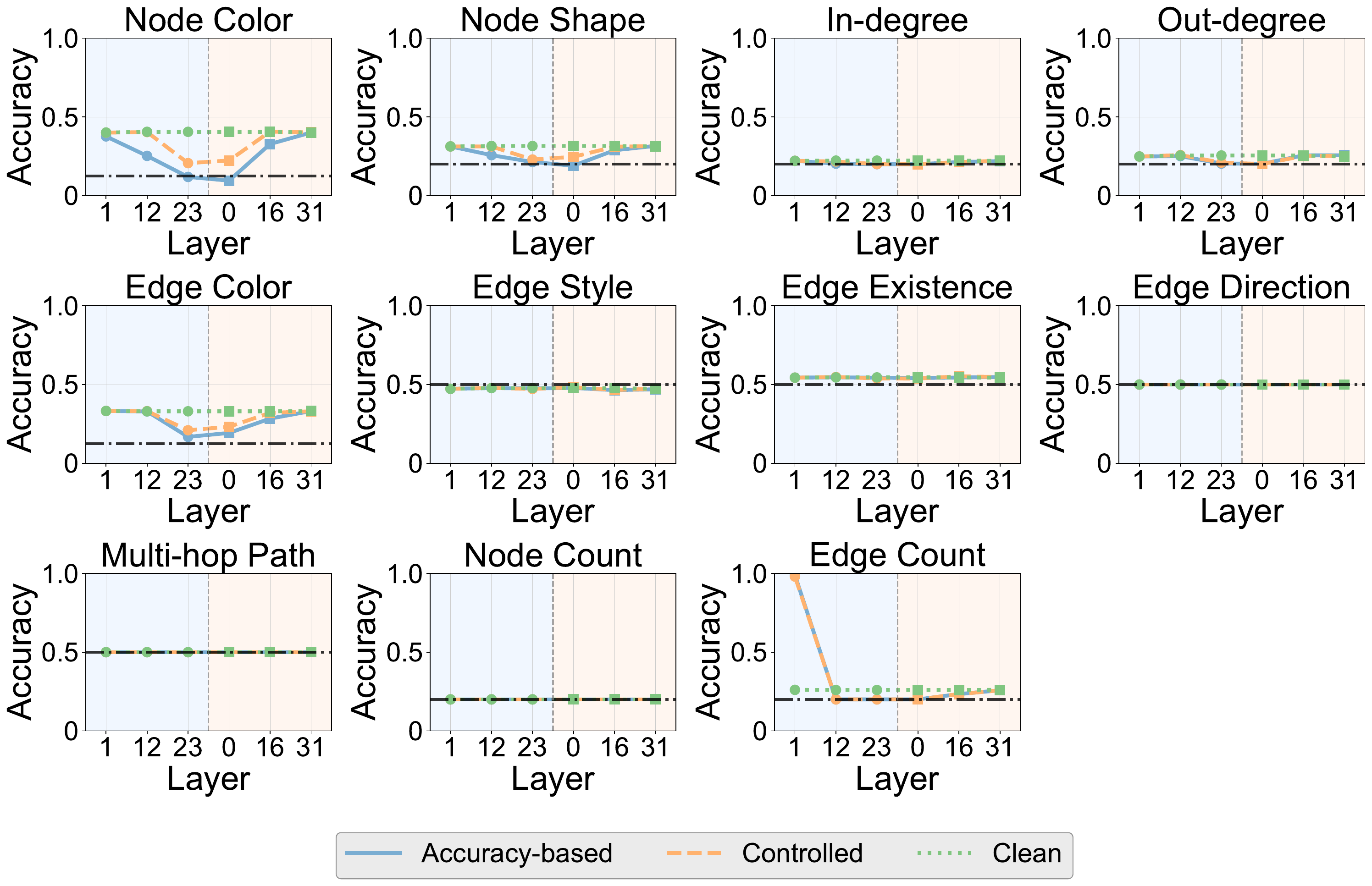}
    \caption{
        \textit{Clean}, \textit{accuracy-based}, and \textit{controlled} accuracy at each intervention layer for LLaVA1.5 7B.
        The background colors of the plots indicate the results for \colorbox[RGB]{242,247,254}{the vision encoder} and \colorbox[RGB]{254,247,241}{the language model}, respectively.
    }
    \label{fig:causal_intervention_llava1.5_plot}
\end{figure*}

\paragraph{LLaVA1.5 7B.}
Table~\ref{tab:causal_intervention_diff_llava1.5} and Figure~\ref{fig:causal_intervention_llava1.5_plot} report the results for LLaVA1.5 7B.
Although LLaVA1.5 has lower baseline VQA performance than the other models, we still observed an accuracy drop after the intervention for aspects such as \texttt{Node Color}, \texttt{Node Shape}, and \texttt{Edge Color}.

\begin{table*}[t]
    \tabcolsep 1mm
    \small
    \centering
    \begin{tabular}{lcrrrrrrrrrrr}
        \toprule
        & Layer
        & \makecell{\texttt{Node}\\\texttt{Color}}
        & \makecell{\texttt{Node}\\\texttt{Shape}}
        & \makecell{\texttt{In-deg.}}
        & \makecell{\texttt{Out-deg.}}
        & \makecell{\texttt{Edge}\\\texttt{Color}}
        & \makecell{\texttt{Edge}\\\texttt{Style}}
        & \makecell{\texttt{Edge}\\\texttt{Exist.}}
        & \makecell{\texttt{Edge}\\\texttt{Direct.}}
        & \makecell{\texttt{Multi-hop}\\\texttt{Path}}
        & \makecell{\texttt{Node}\\\texttt{Count}}
        & \makecell{\texttt{Edge}\\\texttt{Count}} \\
        \midrule
        \multirow{3}{*}{\makecell[l]{Vision\\encoder}}
        & 1
        & \textbf{$-$2.4}
        & $\pm$0.0
        & $\pm$0.0
        & $\pm$0.0
        & $\pm$0.0
        & $\pm$0.0
        & $\pm$0.0
        & $\pm$0.0
        & $\pm$0.0
        & $\pm$0.0
        & $+$0.9 \\
        & 12
        & \textbf{$-$15.1}
        & \textbf{$-$5.5}
        & \textbf{$-$1.4}
        & \textbf{$-$0.7}
        & \textbf{$-$0.2}
        & \textbf{$-$0.1}
        & \textbf{$-$0.2}
        & $\pm$0.0
        & $\pm$0.0
        & $\pm$0.0
        & $\pm$0.0 \\
        & 23
        & \textbf{$-$8.8}
        & \textbf{$-$1.6}
        & $\pm$0.0
        & \textbf{$-$0.6}
        & \textbf{$-$4.2}
        & $+$0.4
        & $+$0.4
        & $\pm$0.0
        & $\pm$0.0
        & $\pm$0.0
        & $\pm$0.0 \\
        \midrule
        \multirow{3}{*}{\makecell[l]{Language\\model}}
        & 0
        & \textbf{$-$12.8}
        & \textbf{$-$5.7}
        & $\pm$0.0
        & $+$0.1
        & \textbf{$-$3.9}
        & \textbf{$-$0.5}
        & $+$0.3
        & $\pm$0.0
        & $\pm$0.0
        & $\pm$0.0
        & $\pm$0.0 \\
        & 16
        & \textbf{$-$7.9}
        & \textbf{$-$2.3}
        & $+$0.0
        & $+$0.4
        & \textbf{$-$3.6}
        & \textbf{$-$0.1}
        & \textbf{$-$0.8}
        & $\pm$0.0
        & $\pm$0.0
        & $\pm$0.0
        & $\pm$0.0 \\
        & 31
        & \textbf{$-$0.1}
        & $+$0.1
        & \textbf{$-$0.2}
        & $+$0.4
        & $+$0.1
        & \textbf{$-$0.3}
        & $\pm$0.0
        & $\pm$0.0
        & $\pm$0.0
        & $\pm$0.0
        & $+$0.0 \\
        \bottomrule
    \end{tabular}
    \caption{
        Accuracy difference between \textit{accuracy-based} and \textit{controlled} settings (\textit{accuracy-based} $-$ \textit{controlled}) for LLaVA1.5 7B.
        More negative values (\textbf{bold}) indicate a larger accuracy decrease under the \textit{accuracy-based} intervention than under the \textit{controlled} intervention.
    }
    \label{tab:causal_intervention_diff_llava1.5}
\end{table*}

\begin{figure*}[t]
    \centering
    \includegraphics[width=\linewidth]{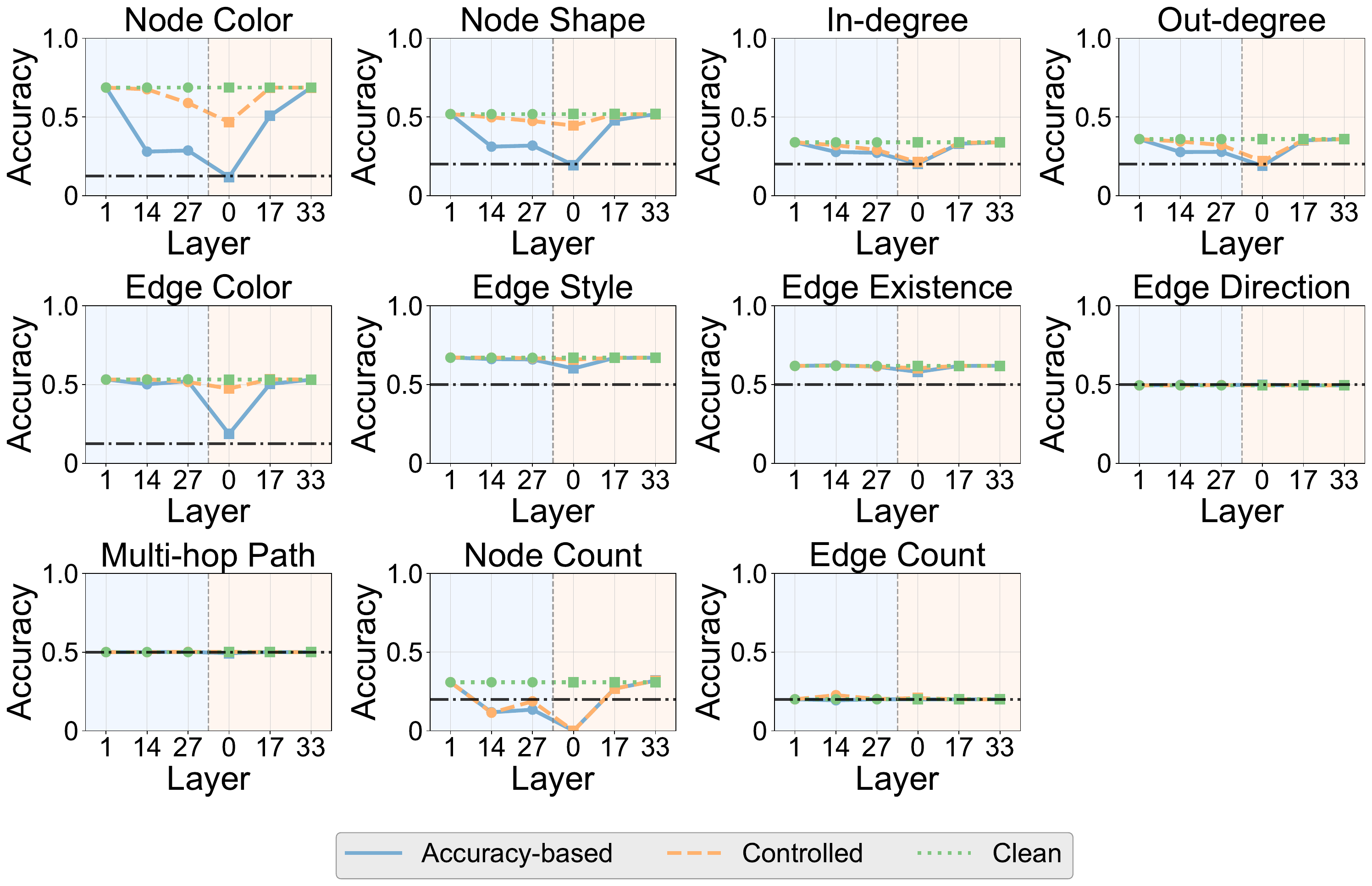}
    \caption{
        \textit{Clean}, \textit{accuracy-based}, and \textit{controlled} accuracy at each intervention layer for Gemma3-4B-IT.
        The background colors of the plots indicate the results for \colorbox[RGB]{242,247,254}{the vision encoder} and \colorbox[RGB]{254,247,241}{the language model}, respectively.
    }
    \label{fig:causal_intervention_gemma3_plot}
\end{figure*}

\paragraph{Gemma3-4B-IT.}
Table~\ref{tab:causal_intervention_diff_gemma3} and Figure~\ref{fig:causal_intervention_gemma3_plot} report the results for Gemma3-4B-IT.
We observe a drop in performance after the intervention for multiple aspects, including \texttt{Node Color}, \texttt{Node Shape}, and \texttt{Edge Color}, indicating that linearly decodable information in both the vision encoder and the language model contributes to VQA inference.

\begin{table*}[t]
    \tabcolsep 1mm
    \small
    \centering
    \begin{tabular}{lcrrrrrrrrrrr}
        \toprule
        & Layer
        & \makecell{\texttt{Node}\\\texttt{Color}}
        & \makecell{\texttt{Node}\\\texttt{Shape}}
        & \makecell{\texttt{In-deg.}}
        & \makecell{\texttt{Out-deg.}}
        & \makecell{\texttt{Edge}\\\texttt{Color}}
        & \makecell{\texttt{Edge}\\\texttt{Style}}
        & \makecell{\texttt{Edge}\\\texttt{Exist.}}
        & \makecell{\texttt{Edge}\\\texttt{Direct.}}
        & \makecell{\texttt{Multi-hop}\\\texttt{Path}}
        & \makecell{\texttt{Node}\\\texttt{Count}}
        & \makecell{\texttt{Edge}\\\texttt{Count}} \\
        \midrule
        \multirow{3}{*}{\makecell[l]{Vision\\encoder}}
        & 1
        & $\pm$0.0
        & $\pm$0.0
        & $\pm$0.0
        & $\pm$0.0
        & $+$0.1
        & $\pm$0.0
        & $\pm$0.0
        & $\pm$0.0
        & $\pm$0.0
        & $\pm$0.0
        & $\pm$0.0 \\
        & 14
        & \textbf{$-$39.6}
        & \textbf{$-$18.6}
        & \textbf{$-$4.3}
        & \textbf{$-$6.7}
        & \textbf{$-$3.1}
        & \textbf{$-$0.9}
        & $+$0.2
        & $+$0.3
        & $\pm$0.0
        & $+$0.3
        & \textbf{$-$3.3} \\
        & 27
        & \textbf{$-$30.2}
        & \textbf{$-$15.6}
        & \textbf{$-$1.8}
        & \textbf{$-$4.3}
        & $+$0.5
        & $\pm$0.8
        & \textbf{$-$0.1}
        & $+$0.1
        & $\pm$0.0
        & \textbf{$-$5.3}
        & \textbf{$-$0.1} \\
        \midrule
        \multirow{3}{*}{\makecell[l]{Language\\model}}
        & 0
        & \textbf{$-$34.9}
        & \textbf{$-$25.1}
        & \textbf{$-$1.0}
        & \textbf{$-$3.0}
        & \textbf{$-$28.8}
        & \textbf{$-$5.7}
        & \textbf{$-$2.3}
        & $+$0.3
        & \textbf{$-$0.8}
        & $+$0.1
        & \textbf{$-$0.9} \\
        & 17
        & \textbf{$-$17.9}
        & \textbf{$-$3.9}
        & \textbf{$-$0.6}
        & \textbf{$-$0.3}
        & \textbf{$-$3.0}
        & $\pm$0.0
        & $\pm$0.0
        & \textbf{$-$0.1}
        & $\pm$0.0
        & $\pm$0.0
        & $\pm$0.0 \\
        & 33
        & $\pm$0.0
        & $\pm$0.0
        & $\pm$0.0
        & $\pm$0.0
        & $\pm$0.0
        & $\pm$0.0
        & $+$0.1
        & $\pm$0.0
        & $\pm$0.0
        & $\pm$0.0
        & $\pm$0.0 \\
        \bottomrule
    \end{tabular}
    \caption{
        Accuracy difference between \textit{accuracy-based} and \textit{controlled} settings (\textit{accuracy-based} $-$ \textit{controlled}) for Gemma3-4B-IT.
        More negative values (\textbf{bold}) indicate a larger accuracy decrease under the \textit{accuracy-based} intervention than under the \textit{controlled} intervention.
    }
    \label{tab:causal_intervention_diff_gemma3}
\end{table*}

Across all four models, replacing hidden states at positions with high probing accuracy with the dataset-wide mean reduced accuracy for \texttt{Node Color}, \texttt{Node Shape}, and \texttt{Edge Color} in both the vision encoder and the language model.
Qwen3-VL and Qwen2.5-VL additionally showed drops for \texttt{Node Count}.
These results indicate that the signals identified by probing are not merely decodable but contribute causally to VQA predictions.
In contrast, aspects that are poorly captured by linear probes, such as \texttt{Edge Direction}, \texttt{Edge Existence}, and \texttt{Multi-hop Path}, showed little to no change after intervention.
For these aspects, VQA accuracy is already close to chance level, leaving little room for further decreases, so the intervention is not informative about their causal contribution.

\section{Experiments on Data Closer to Real-World Diagrams}
\label{sec:appendix_real_world}
To examine whether the trends generalize beyond synthetic graphs, we conduct VQA and probing experiments on flowchart images derived from FlowVQA~\cite{Singh2024-qm}, using Qwen3-VL 8B as the target model.
Although the dataset remains controlled because node and edge attributes are edited and augmented, the flowcharts contain features characteristic of real-world flowcharts, such as text labels and canonical layouts.

\subsection{Evaluation Aspects and Class Configuration}
\label{subsec:appendix_real_world_aspects}
We analyze 10 of the 11 aspects defined in \cref{subsec:aspect_definition} of the main text, excluding \texttt{Node Shape} (\cref{sec:flowchart_main}).

The class configuration for each aspect is shown in Table~\ref{tab:appendix_real_world_classes}.
While the configuration largely matches that of the synthetic data, it differs in the following respects:
\begin{itemize}
    \item \texttt{In-degree} / \texttt{Out-degree}: 5 classes (0--4) in the synthetic setting $\to$ 4 classes (1--4) in the flowchart setting.
        Except for the start and end nodes, every node in a flowchart is connected by at least one incoming and one outgoing edge, so a degree of zero does not arise.
    \item \texttt{Node Count}: in the synthetic setting, every diagram contains exactly five nodes, so this aspect asks how many nodes share the identifier A (5 classes, 1--5).
        In the flowchart setting, the number of nodes varies across flowcharts, so it asks for the total number of nodes (4 classes, 7--10).
    \item \texttt{Edge Count}: 5 classes (1--5) in the synthetic setting $\to$ 4 classes (7--10) in the flowchart setting.
        We adjusted the range to preserve the distribution of the original FlowVQA data while matching the synthetic data in complexity.
\end{itemize}

\begin{table}[t]
    \tabcolsep 1mm
    \small
    \centering
    \begin{tabular}{lll}
        \toprule
        Category & Aspect $a$ & Label set $\mathcal{Y}$ \\
        \midrule
        \multirow[c]{3}{*}{Single}
        & \texttt{Node Color} & \makecell[l]{\{red, green, yellow, blue,\\brown, orange, pink, purple\}} \\
        & \texttt{In-deg.} & \{1, 2, 3, 4\} \\
        & \texttt{Out-deg.} & \{1, 2, 3, 4\} \\
        \midrule
        \multirow[c]{5}{*}{Multiple}
        & \texttt{Edge Color} & \makecell[l]{\{red, green, yellow, blue,\\brown, orange, pink, purple\}} \\
        & \texttt{Edge Style} & \{solid, dotted\} \\
        & \texttt{Edge Direct.} & \{forward, backward\} \\
        & \texttt{Edge Exist.} & \{exist, not exist\} \\
        & \texttt{Multi-hop Path} & \{exist, not exist\} \\
        \midrule
        \multirow[c]{2}{*}{Global}
        & \texttt{Node Count} & \{7, 8, 9, 10\} \\
        & \texttt{Edge Count} & \{7, 8, 9, 10\} \\
        \bottomrule
    \end{tabular}
    \caption{
        Class configuration of the 10 aspects analyzed in the flowchart setting.
    }
    \label{tab:appendix_real_world_classes}
\end{table}

\subsection{Dataset Construction}
\label{subsec:appendix_real_world_dataset}

We construct the dataset in three steps: (1) sampling base flowcharts from FlowVQA, (2) editing and data augmentation, and (3) Mermaid rendering.

\paragraph{Step 1: Sampling base flowcharts.}
To keep the graph complexity comparable to the synthetic setting (\cref{sec:dataset_construction}), we restrict the source to FlowVQA samples with at most 10 nodes.
From this pool, we independently sample base flowcharts for each aspect.

For aspects that analyze a specific node or node pair (the Single and Multiple categories), we sample 10 flowcharts for training and one for evaluation, all sharing a common target node (or node pair).
Aspects in the Single category target the node with the label ``Return True'', while aspects in the Multiple category target the node pair (``Return True'', ``End'').
We choose these as targets because they are the labels that most frequently appear across FlowVQA flowcharts, allowing us to fix a consistent target across aspects while retaining a sufficient number of candidate flowcharts.

In contrast, for aspects in the Global category (\texttt{Node Count}, \texttt{Edge Count}), which analyze the diagram as a whole, the classification target is the total number of nodes (or edges) in the flowchart.
We therefore sample 10 training and 1 evaluation flowcharts for each class, where each class corresponds to a specific node count (or edge count).
For example, for \texttt{Node Count}, we sample flowcharts with 7, 8, 9, and 10 nodes for each respective class.

We restrict evaluation to a single base flowchart (one per class for aspects in the Global category) because computing patch-wise accuracy requires a fixed node layout within each evaluation set.
Using flowcharts with random node layouts for training and a fixed layout for evaluation also aligns with the synthetic setting.

\paragraph{Step 2: Editing and data augmentation.}
For each base flowchart, we generate samples for each class by editing the attributes of the target node (or pair), following the edit operations in Table~\ref{tab:appendix_real_world_edits}.
Aspects in the Global category require no such edit because their classes are determined at sampling time.
We then apply the following augmentations to obtain 100 samples per class:
\begin{itemize}
    \item Common: randomize the colors of non-target nodes, edge colors, and edge styles.
    \item Training only: shuffle the labels of non-target nodes and randomize their shapes.
\end{itemize}
We apply label shuffling and shape randomization only to the training data to prevent the probe from exploiting non-target node text labels or shapes as shortcuts.

\begin{table*}[t]
    \small
    \centering
    \begin{tabular}{l l}
        \toprule
        Aspect & Edit operation \\
        \midrule
        \texttt{Node Color} & Randomly select the color of the target node from 8 colors. \\
        \texttt{In-deg.} & Add or delete in-edges to set the in-degree of the target node to one of 1--4. \\
        \texttt{Out-deg.} & Add or delete out-edges to set the out-degree of the target node to one of 1--4. \\
        \texttt{Edge Color} & Randomly select the color of the edge between the target pair from 8 colors. \\
        \texttt{Edge Style} & Change the edge between the target pair to solid or dotted. \\
        \texttt{Edge Direct.} & Reverse the direction of the edge between the target pair. \\
        \texttt{Edge Exist.} & Add or delete the edge between the target pair. \\
        \texttt{Multi-hop Path} & Edit intermediate edges to change the reachability between the target pair. \\
        \texttt{Node Count} & No edits (the class is determined at sampling time). \\
        \texttt{Edge Count} & No edits (the class is determined at sampling time). \\
        \bottomrule
    \end{tabular}
    \caption{
        Edit operations for each aspect.
    }
    \label{tab:appendix_real_world_edits}
\end{table*}

\begin{figure}[t]
    \centering
    \includegraphics[width=\linewidth]{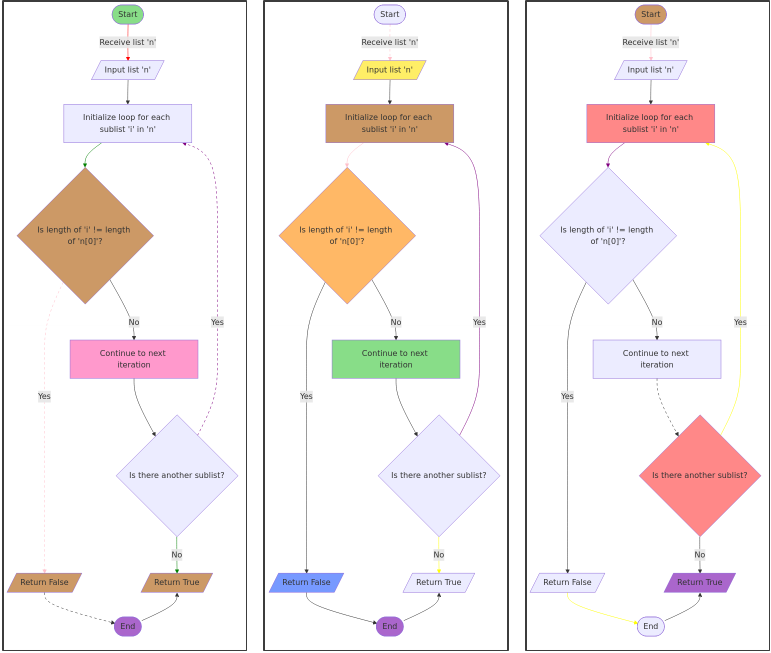}
    \caption{
        Examples of flowcharts used for evaluation.
        The three samples share the same node layout, while node colors, edge colors, and edge styles are randomized.
    }
    \label{fig:sample_flowcharts}
\end{figure}

\paragraph{Step 3: Mermaid rendering.}
We render the edited Mermaid code as PNG images using \texttt{mermaid-cli} and provide them directly to Qwen3-VL without resizing or other preprocessing.
Figure~\ref{fig:sample_flowcharts} shows examples of the rendered flowcharts.

\subsection{Preliminary Experiment: VQA}
\label{subsec:appendix_real_world_vqa}

We design our own multiple-choice questions for this analysis rather than using those provided in FlowVQA.
The questions follow the same format as in \cref{sec:experiment_vqa} of the main text, explicitly specifying the target node (or pair) for each aspect.
The full list of questions for each aspect is provided in Table~\ref{tab:appendix_real_world_questions}.
As in the synthetic setting, we evaluate VQA using accuracy.

\begin{table*}[t]
    \small
    \centering
    \begin{tabular}{l l}
        \toprule
        Aspect & Question \\
        \midrule
        \texttt{Node Color} & What color is the `Return True' node? \\
        \texttt{In-deg.} & What is the in-degree of the `Return True' node? \\
        \texttt{Out-deg.} & What is the out-degree of the `Return True' node? \\
        \texttt{Edge Color} & What color is the edge between the `Return True' node and the `End' node? \\
        \texttt{Edge Style} & What style is the edge between the `Return True' node and the `End' node? \\
        \texttt{Edge Direct.} & What is the direction of the edge between the `Return True' node and the `End' node? \\
        \texttt{Edge Exist.} & Does an edge exist between the `Return True' node and the `End' node? \\
        \texttt{Multi-hop Path} & Is there a path from the `Return True' node to the `End' node? \\
        \texttt{Node Count} & How many nodes are in the graph? \\
        \texttt{Edge Count} & How many edges are in the graph? \\
        \bottomrule
    \end{tabular}
    \caption{
        Questions used to evaluate each aspect in the flowchart setting.
        ``Answer with only one word from the choices.'' and `` (Choices: <comma-separated-choices>)'' are appended to every question.
    }
    \label{tab:appendix_real_world_questions}
\end{table*}

\subsection{Probing}
\label{subsec:appendix_real_world_probing}

\paragraph{Setup.}
We use the same single-layer linear classifier as in \cref{sec:probing} of the main text, training separate probes on the hidden states of the vision encoder, the image-input part of the language model, and the text-input part of the language model.
The probe architecture and training procedure are identical to those used on the synthetic data.
We modify two hyperparameters of the vision encoder probe, as listed in Table~\ref{tab:appendix_real_world_probe_hyperparams}: we reduce the maximum number of epochs to 200, which we found sufficient for convergence, and shift the batch-size search range to larger values to accommodate the increased number of patches caused by the higher input resolution.
As in the synthetic setting, we perform a layer-wise analysis that tracks accuracy across layers and a position-wise analysis that examines how accuracy is distributed across patch positions (for the vision encoder and the image-input part) or token positions (for the text-input part).

\begin{table}[t]
    \small
    \centering
    \begin{tabular}{lc}
        \toprule
        Parameter & Vision encoder probe \\
        \midrule
        Epochs & 200 \\
        Batch size\textsuperscript{*} & \{4000, 8000, 16000, 32000, 64000\} \\
        \bottomrule
    \end{tabular}
    \caption{
        Hyperparameters of probe training in the flowchart setting that differ from the synthetic setting (Table~\ref{tab:probe_hyperparameters}).
        All other hyperparameters are identical to those of the synthetic setting.
        \textsuperscript{*}Subject to hyperparameter search.
    }
    \label{tab:appendix_real_world_probe_hyperparams}
\end{table}

\begin{figure}[t]
    \centering
    \includegraphics[width=\linewidth]{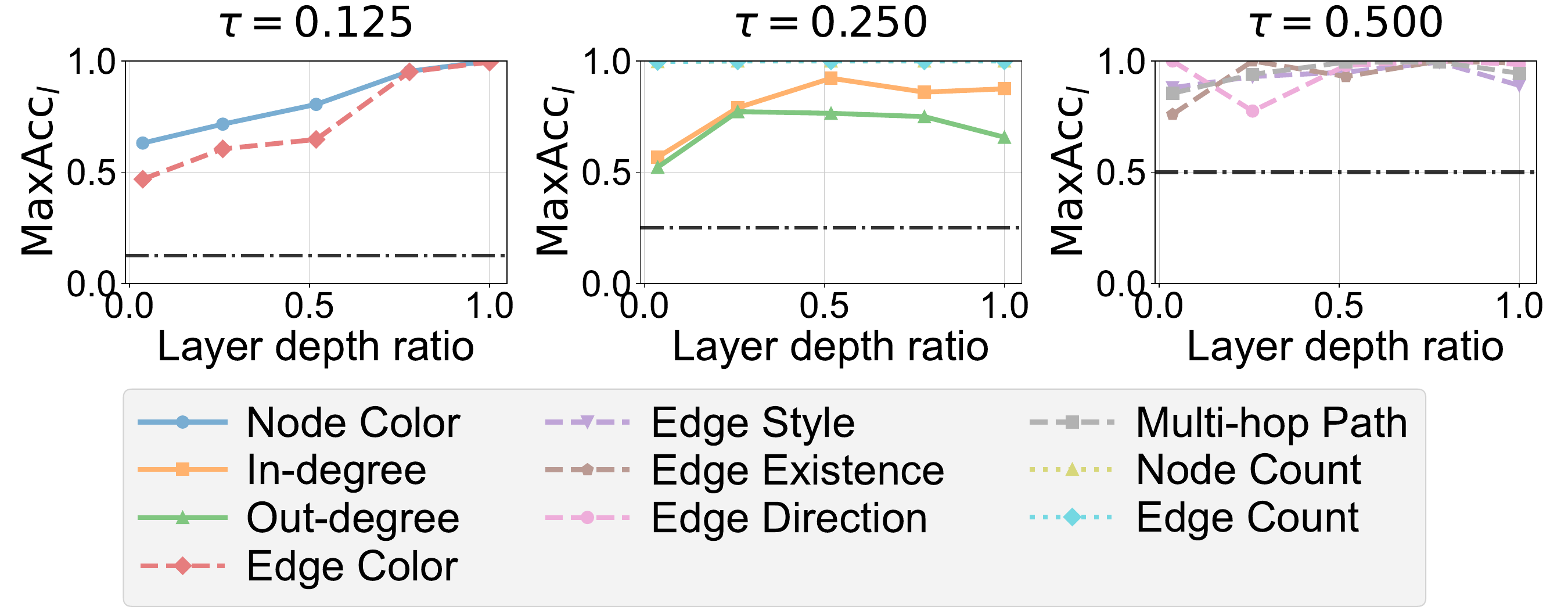}
    \caption{
        $\mathrm{MaxAcc}_{l}$ per layer in the vision encoder of Qwen3-VL 8B on flowcharts.
        The x-axis shows relative layer position (0 is the input layer, 1 is the final layer), and the y-axis shows accuracy.
        The black dash-dot line in each panel indicates the threshold $\tau$.
        All aspects exceeded the threshold within the vision encoder, including Multiple-category aspects whose accuracy did not reach the threshold in the synthetic setting.
    }
    \label{fig:appendix_real_world_layerwise_vision}
\end{figure}

\begin{figure*}[t]
    \centering
    \includegraphics[width=\linewidth]{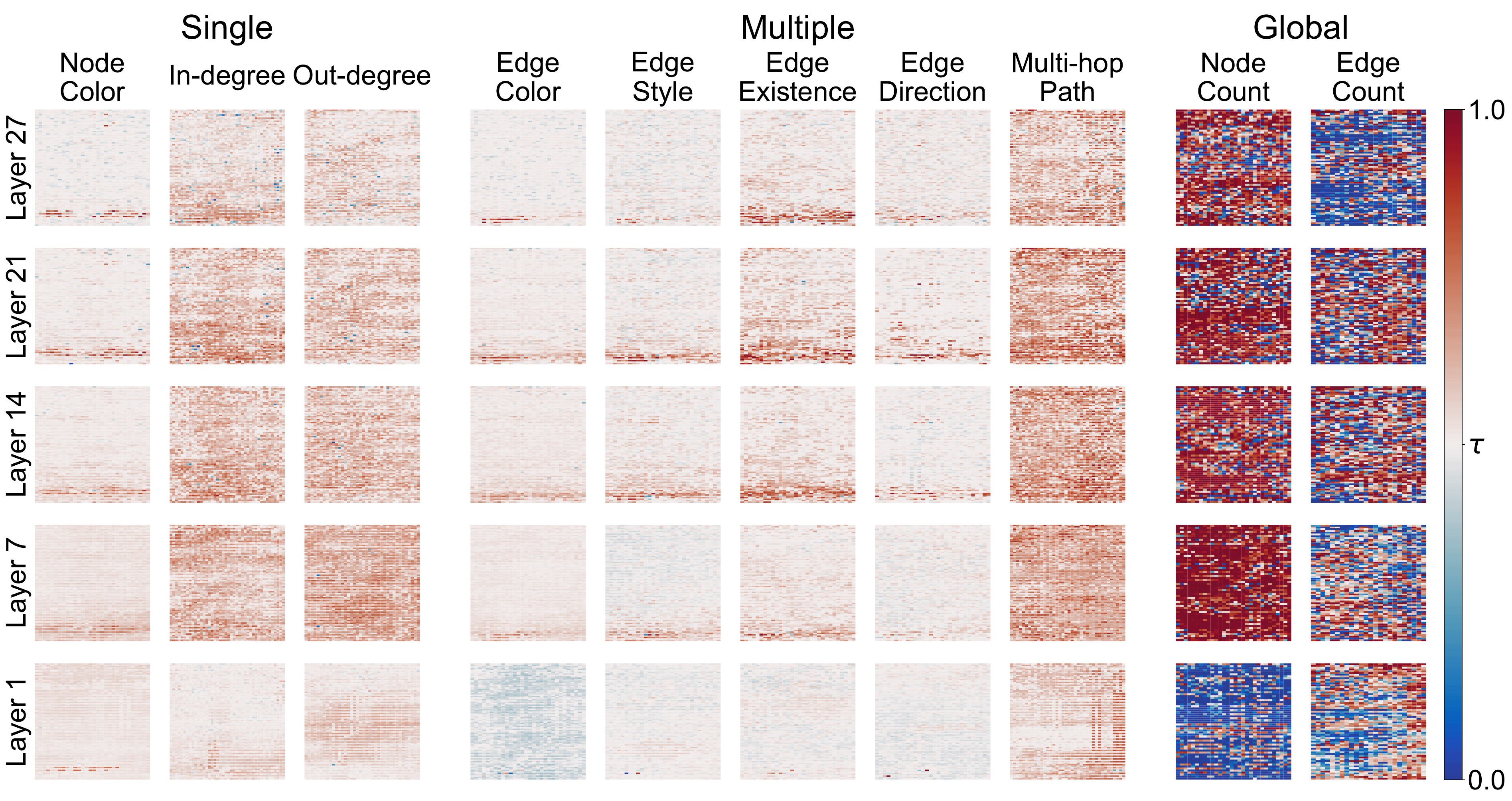}
    \caption{
        $\mathrm{acc}_{l, t}(d_{\mathrm{fix}})$ in the vision encoder of Qwen3-VL 8B on flowcharts, for a representative subset $d_{\mathrm{fix}}$.
        Each heatmap shows accuracy by patch position for a specific layer and aspect.
        Single-category aspects localized near the target node, Global-category aspects spread across background patches, and Multiple-category aspects also showed high-accuracy patches, unlike in the synthetic setting.
    }
    \label{fig:appendix_real_world_patchwise_vision}
\end{figure*}

\paragraph{All aspects, including those in the Multiple category, are linearly decodable in the vision encoder.}
Figure~\ref{fig:appendix_real_world_layerwise_vision} shows $\mathrm{MaxAcc}_{l}$ at each layer of the vision encoder, and Figure~\ref{fig:appendix_real_world_patchwise_vision} shows the patch-wise accuracy distribution.
For all aspects, accuracy exceeded the threshold within the vision encoder, indicating that the information for every aspect is linearly decodable from a single hidden state already within the vision encoder.
In the patch-wise distribution, aspects in the Single category showed high-accuracy patches concentrated around the target node, while aspects in the Global category showed high accuracy across a broad range of patches, including background patches.
These spatial patterns for the two categories are consistent with the synthetic results (\cref{subsec:probing_results}).
In contrast, aspects in the Multiple category showed a different trend from the synthetic data:
while most aspects in the Multiple category did not exceed the threshold within the vision encoder in the synthetic setting, in this setting all aspects in the Multiple category exceeded the threshold.

\begin{figure}[t]
    \centering
    \includegraphics[width=\linewidth]{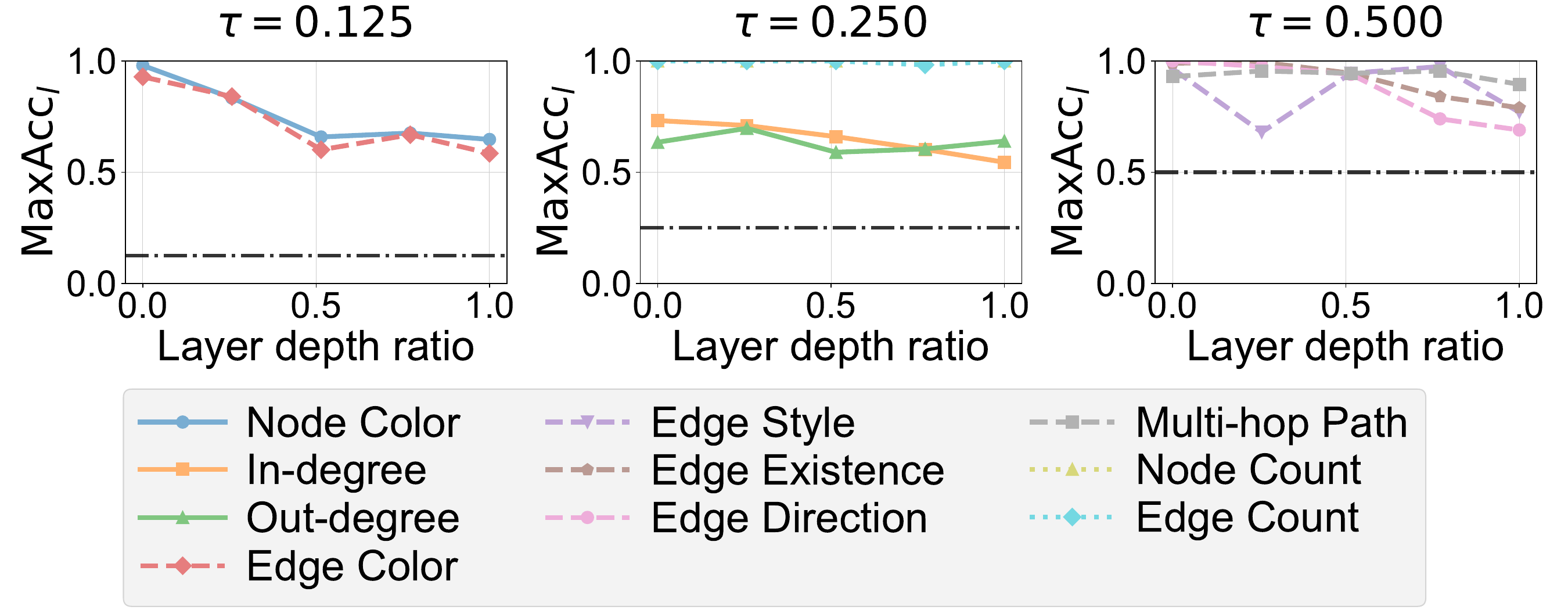}
    \caption{
        $\mathrm{MaxAcc}_{l}$ per layer in the image-input part of the language model in Qwen3-VL 8B on flowcharts.
        The x-axis shows relative layer position (0 is the input layer, 1 is the final layer), and the y-axis shows accuracy.
        The black dash-dot line in each panel indicates the threshold $\tau$.
        All aspects maintained accuracy above the threshold.
    }
    \label{fig:appendix_real_world_layerwise_lm_image}
\end{figure}

\begin{figure*}[t]
    \centering
    \includegraphics[width=\linewidth]{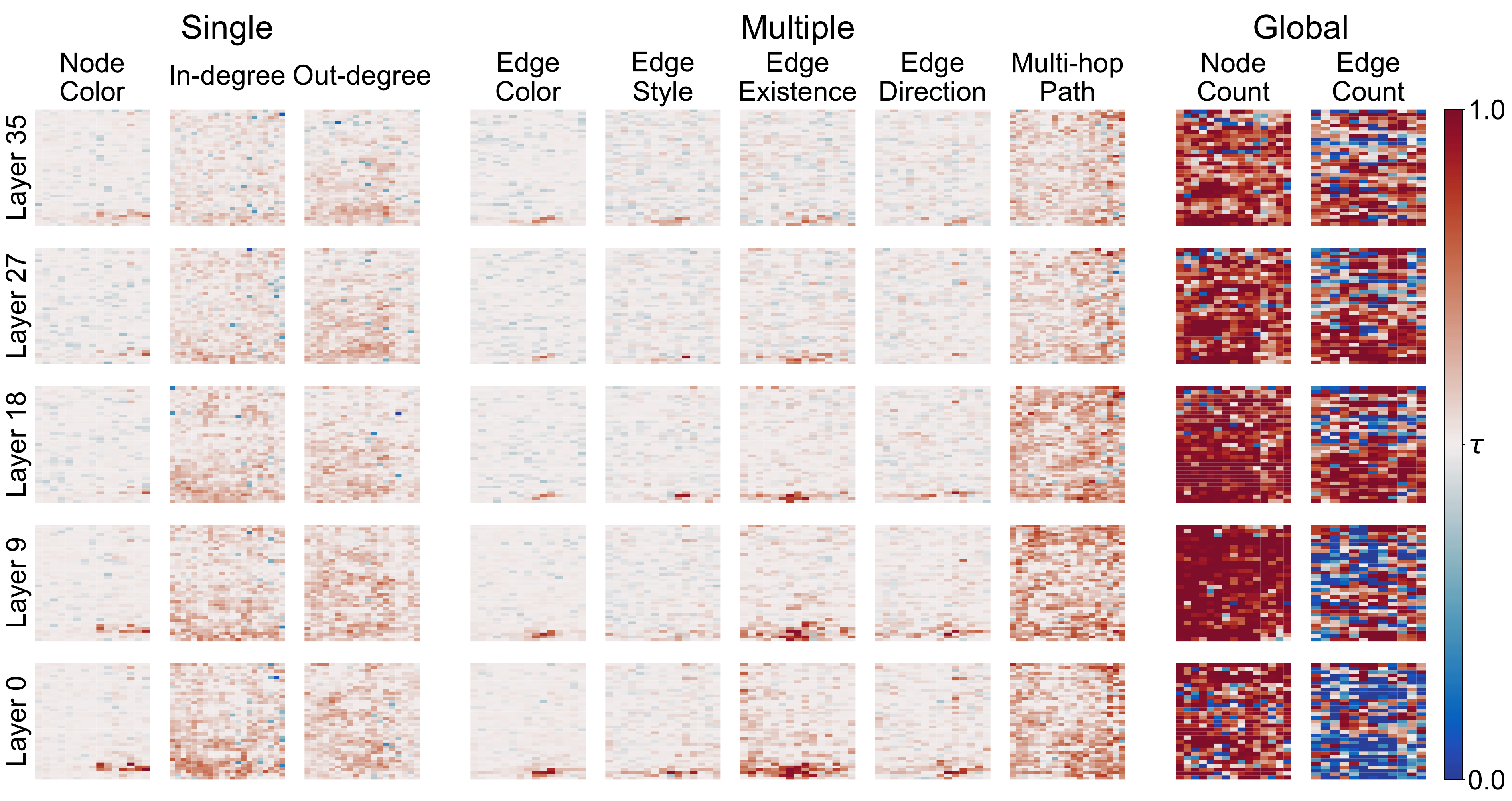}
    \caption{
        $\mathrm{acc}_{l, t}(d_{\mathrm{fix}})$ in the image-input part of the language model in Qwen3-VL 8B on flowcharts, for a representative subset $d_{\mathrm{fix}}$.
        The spatial distribution of high-accuracy patches was largely unchanged from the vision encoder result.
    }
    \label{fig:appendix_real_world_patchwise_lm_image}
\end{figure*}

\paragraph{Linearly decodable signals are preserved in the image-input part of the language model.}
Figure~\ref{fig:appendix_real_world_layerwise_lm_image} shows $\mathrm{MaxAcc}_{l}$ across language model layers, and Figure~\ref{fig:appendix_real_world_patchwise_lm_image} shows the patch-wise accuracy distribution.
All aspects maintained accuracy above the threshold throughout the image-input part, and the patch-wise distribution showed little change from the vision encoder result.
These results indicate that the linearly decodable signals observed in the vision encoder were preserved through the image-input part, consistent with the synthetic results (\cref{subsec:probing_results}).

\begin{figure*}[t]
    \centering
    \includegraphics[width=\linewidth]{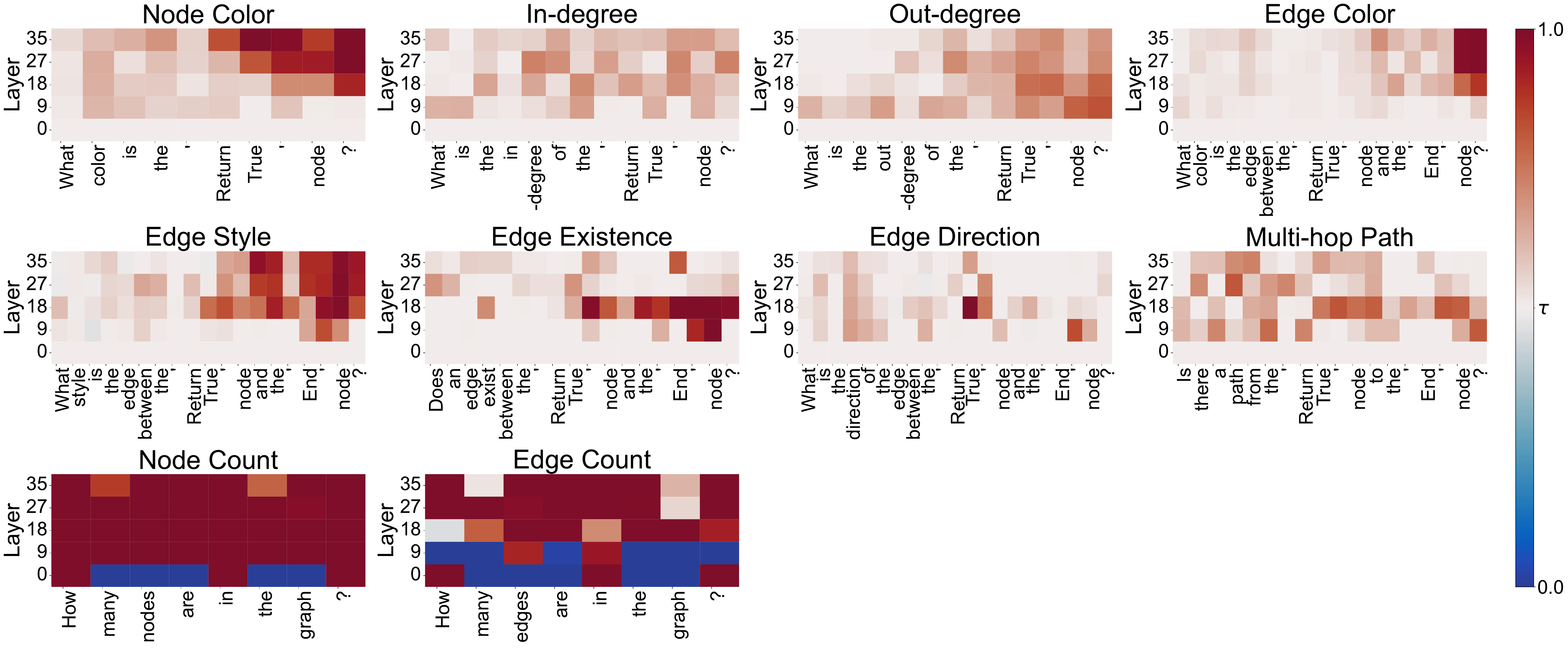}
    \caption{
        Probing results for text tokens in Qwen3-VL 8B on flowcharts.
        The x-axis shows token position in the question, and the y-axis shows layer position.
        For \texttt{Edge Direction} and \texttt{Edge Existence}, high accuracy appeared only in intermediate layers and did not propagate to the final layer.
    }
    \label{fig:appendix_real_world_text_probing}
\end{figure*}

\paragraph{Signals for \texttt{Edge Direction} and \texttt{Edge Existence} are confined to intermediate text-input layers.}
Figure~\ref{fig:appendix_real_world_text_probing} shows $\mathrm{acc}_{l, t}$ across layers and token positions in the text-input part of the language model.
Similar to the synthetic results (\cref{subsec:probing_results}), accuracy was particularly high at token positions that specify the target node or pair.
However, for \texttt{Edge Direction} and \texttt{Edge Existence}, high accuracy appeared only in intermediate layers and did not propagate to the final layer.
This suggests that, although signals corresponding to these aspects exist within the model, they are not propagated to the final layer; this differs from the synthetic results, where these aspects maintained high accuracy through the final layer.

\paragraph{Possible causes of the differences from the synthetic results.}
The results in this setting differ from those on the synthetic data in two respects:
(1) aspects in the Multiple category became linearly decodable already within the vision encoder, and (2) in the text-input part of the language model, the signals for \texttt{Edge Direction} and \texttt{Edge Existence} remained in intermediate layers and did not propagate to the final layer.
Several factors may explain these differences, including the complexity of text labels in FlowVQA flowcharts, the higher input image resolution, and the diversity of node and edge layouts.

\section{Analysis of LLaVA Training Data}
\label{sec:appendix_llava_training_data}

Prior studies report that LVLMs struggle to understand relational information indicated by arrows and lines~\cite{Hou2025-mf, Zhu2025-mw, Pan2024-ls}.
Consistent with these findings, our VQA experiments show that all models achieved near-chance accuracy on \texttt{Edge Direction} (\cref{subsec:vqa_experimental_results}, Appendix~\ref{sec:appendix_vqa}).
One possible explanation is that text describing arrow directions is underrepresented in the training data.
To investigate this hypothesis, we analyze the publicly available pre-training and instruction tuning datasets of LLaVA1.5.

\subsection{Setup}

We quantify how many samples mention arrow directions in the pre-training and instruction tuning datasets of LLaVA1.5.
First, we filter samples whose text contains the substring ``arrow.''
We lowercase the text before matching.
We then classify whether each filtered sample describes an arrow direction (e.g., pointing left or upward) using gpt-oss-120b~\cite{OpenAI2025-gk} as an LLM judge.
Note that we analyze only the text portion of each sample, without referring to the associated images.

\subsection{Results}

The substring ``arrow'' appears in 1,108 of 558,128 pre-training samples (0.20\%) and in 4,408 of 665,298 instruction tuning samples (0.66\%).
Among these, only 161 pre-training samples (0.03\% of all pre-training samples) and 646 instruction tuning samples (0.10\%) were judged as describing an arrow direction.
These results suggest that the scarcity of arrow direction descriptions in the training data may be one factor contributing to the near-chance \texttt{Edge Direction} accuracy.
However, since our analysis targets only the text portion and does not account for visual information in the images, further investigation is needed to draw stronger conclusions.
Additionally, this analysis covers only LLaVA1.5; whether similar trends hold for the training datasets of other models remains an open question.

\section{Fine-Tuned Model Analysis}
\label{sec:appendix_finetuning}
To investigate whether training on diagram data changes how LVLMs represent visual information, we fine-tune Qwen3-VL 8B on our synthetic diagram dataset and analyze the resulting changes in VQA accuracy and internal representations.

\subsection{Training Setup}
\label{subsec:appendix_finetuning_setup}
We fine-tune Qwen3-VL 8B using LoRA~\cite{Hu2022-zw} in two configurations: (1) updating only the language model (FT w/L), and (2) updating both the vision encoder and the language model (FT w/VL).
The input format follows the VQA setting described in \cref{sec:general_setting}: each training sample consists of a diagram image and a question, and the model is trained to output the corresponding label as text.

We combine $\mathcal{D}_{\mathrm{rand}}$ from all 11 aspects into a single training set of 4,900 samples.
These diagrams are generated with a different random seed from those used for probe training, so the fine-tuning data and the probe training data do not overlap.
The LoRA and training hyperparameters are summarized in Table~\ref{tab:finetuning_hyperparameters}.

\begin{table}[t]
    \small
    \centering
    \begin{tabular}{lcc}
        \toprule
        Parameter & FT w/L & FT w/VL \\
        \midrule
        Epochs & 10 & 10 \\
        Batch size & 8 & 8 \\
        Learning rate & 1e-4 & 1e-4 \\
        Optimizer & AdamW & AdamW \\
        \quad betas & [0.9, 0.999] & [0.9, 0.999] \\
        \quad eps & 1e-8 & 1e-8 \\
        \quad weight decay & 0.01 & 0.01 \\
        LoRA rank & 4 & 4 \\
        LoRA alpha & 16 & 16 \\
        Target modules & q\_proj, v\_proj & qkv, q\_proj, v\_proj \\
        \bottomrule
    \end{tabular}
    \caption{
        Hyperparameters for LoRA fine-tuning.
    }
    \label{tab:finetuning_hyperparameters}
\end{table}

\subsection{VQA Accuracy after Fine-Tuning}
\label{subsec:appendix_finetuning_vqa}
We evaluate the VQA accuracy of the fine-tuned models using the same protocol as in \cref{sec:experiment_vqa}.
Table~\ref{tab:finetuning_vqa} reports the results.
Fine-tuning improved accuracy on almost all aspects, with the exception of \texttt{Node Color} and \texttt{Node Count} under FT w/VL.

In particular, accuracy improved on all aspects even when only the language model was fine-tuned without updating the vision encoder (FT w/L).
Since the vision encoder is frozen in this configuration, this suggests that the improvement does not come from newly acquired signals in the vision encoder, but rather that the language model became able to use signals that were already present.

Comparing the two configurations, aspects in the Single and Global categories achieved higher accuracy when only the language model was fine-tuned (FT w/L), with the exception of \texttt{Out-degree}.
In contrast, aspects in the Multiple category consistently achieved higher accuracy when both the vision encoder and language model were fine-tuned (FT w/VL).
This suggests that updating the vision encoder is beneficial mainly for aspects whose signals are not yet strongly present in the pre-trained vision encoder, such as those in the Multiple category.

\begin{table*}[t]
    \tabcolsep 1mm
    \footnotesize
    \centering
    \begin{tabular}{lrrrrrrrrrrr}
        \toprule
        & \makecell{\texttt{Node}\\\texttt{Color}}
        & \makecell{\texttt{Node}\\\texttt{Shape}}
        & \makecell{\texttt{In-deg.}}
        & \makecell{\texttt{Out-deg.}}
        & \makecell{\texttt{Edge}\\\texttt{Color}}
        & \makecell{\texttt{Edge}\\\texttt{Style}}
        & \makecell{\texttt{Edge}\\\texttt{Exist.}}
        & \makecell{\texttt{Edge}\\\texttt{Direct.}}
        & \makecell{\texttt{Multi-hop}\\\texttt{Path}}
        & \makecell{\texttt{Node}\\\texttt{Count}}
        & \makecell{\texttt{Edge}\\\texttt{Count}} \\
        \midrule
        Baseline
        & 95.5
        & 78.4
        & 53.7
        & 40.6
        & 80.6
        & 85.7
        & 81.2
        & 53.9
        & 55.9
        & 78.7
        & 83.8 \\
        FT w/L
        & \textbf{99.0}
        & \textbf{86.8}
        & \textbf{71.1}
        & 66.0
        & 90.6
        & 91.1
        & 86.1
        & 70.8
        & 59.1
        & \textbf{79.9}
        & \textbf{88.3} \\
        FT w/VL
        & 93.5
        & 85.0
        & 63.8
        & \textbf{68.3}
        & \textbf{91.6}
        & \textbf{91.5}
        & \textbf{87.8}
        & \textbf{72.7}
        & \textbf{68.0}
        & 77.3
        & 87.8 \\
        \midrule
        Chance level
        & 12.5
        & 20.0
        & 20.0
        & 20.0
        & 12.5
        & 50.0
        & 50.0
        & 50.0
        & 50.0
        & 20.0
        & 20.0 \\
        \bottomrule
    \end{tabular}
    \caption{
        VQA accuracy of Qwen3-VL 8B before and after fine-tuning.
        Baseline denotes the pre-trained model without fine-tuning.
        FT w/L denotes fine-tuning only the language model, and FT w/VL denotes fine-tuning both the vision encoder and the language model.
        FT w/L improved accuracy on all aspects, while FT w/VL slightly decreased \texttt{Node Color} and \texttt{Node Count}.
        Single- and Global-category aspects benefited more from FT w/L, while Multiple-category aspects benefited more from FT w/VL.
    }
    \label{tab:finetuning_vqa}
\end{table*}

\subsection{Changes in Internal Representations}
\label{subsec:appendix_finetuning_representations}
\begin{figure}[t]
    \centering
    \includegraphics[width=\linewidth]{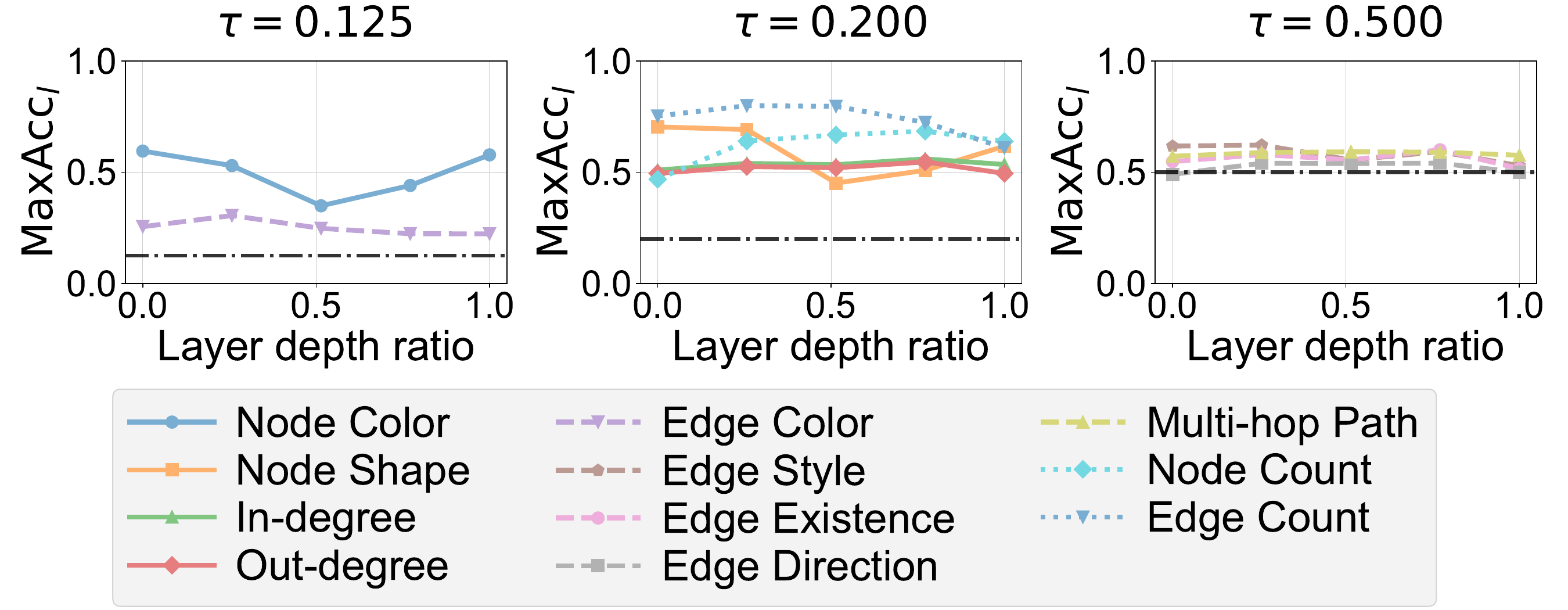}
    \caption{
        $\mathrm{MaxAcc}_{l}$ per layer in the language model of Qwen3-VL 8B (FT w/L).
        Axes and the threshold line follow Figure~\ref{fig:layerwise_image_probing_qwen3_language_model}.
        Trends were largely unchanged from the pre-trained model (Figure~\ref{fig:layerwise_image_probing_qwen3_language_model}).
    }
    \label{fig:layerwise_image_probing_qwen3_ftL_language_model}
\end{figure}

\begin{figure*}[t]
    \centering
    \includegraphics[width=\linewidth]{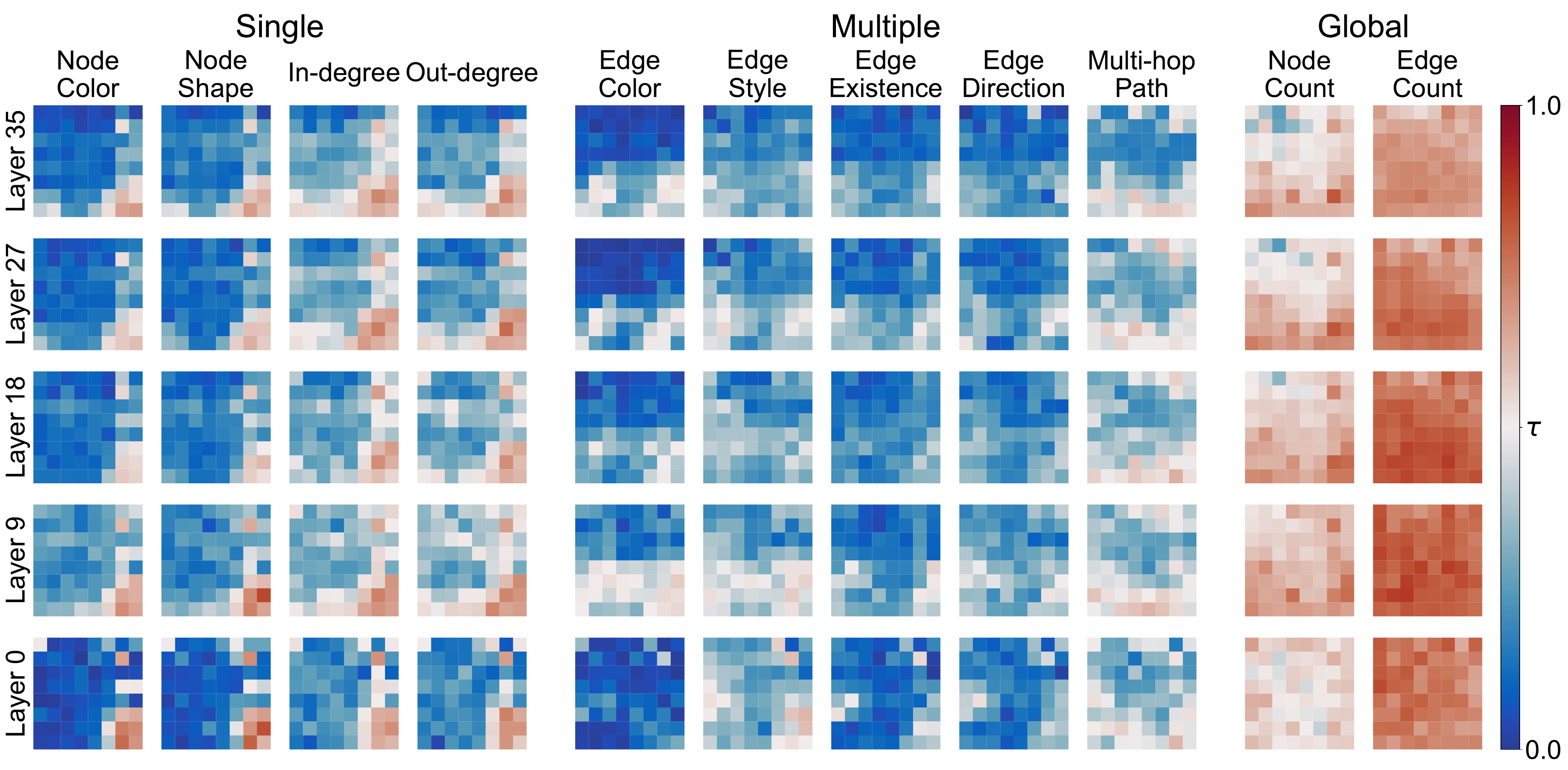}
    \caption{
        Accuracy for each hidden state in the image-input part of the Qwen3-VL 8B language model (FT w/L).
        Each heatmap shows the accuracy of individual hidden states for a specific layer and aspect.
    }
    \label{fig:patchwise_image_probing_qwen3_ftL_language_model}
\end{figure*}

\begin{figure*}[t]
    \centering
    \includegraphics[width=\linewidth]{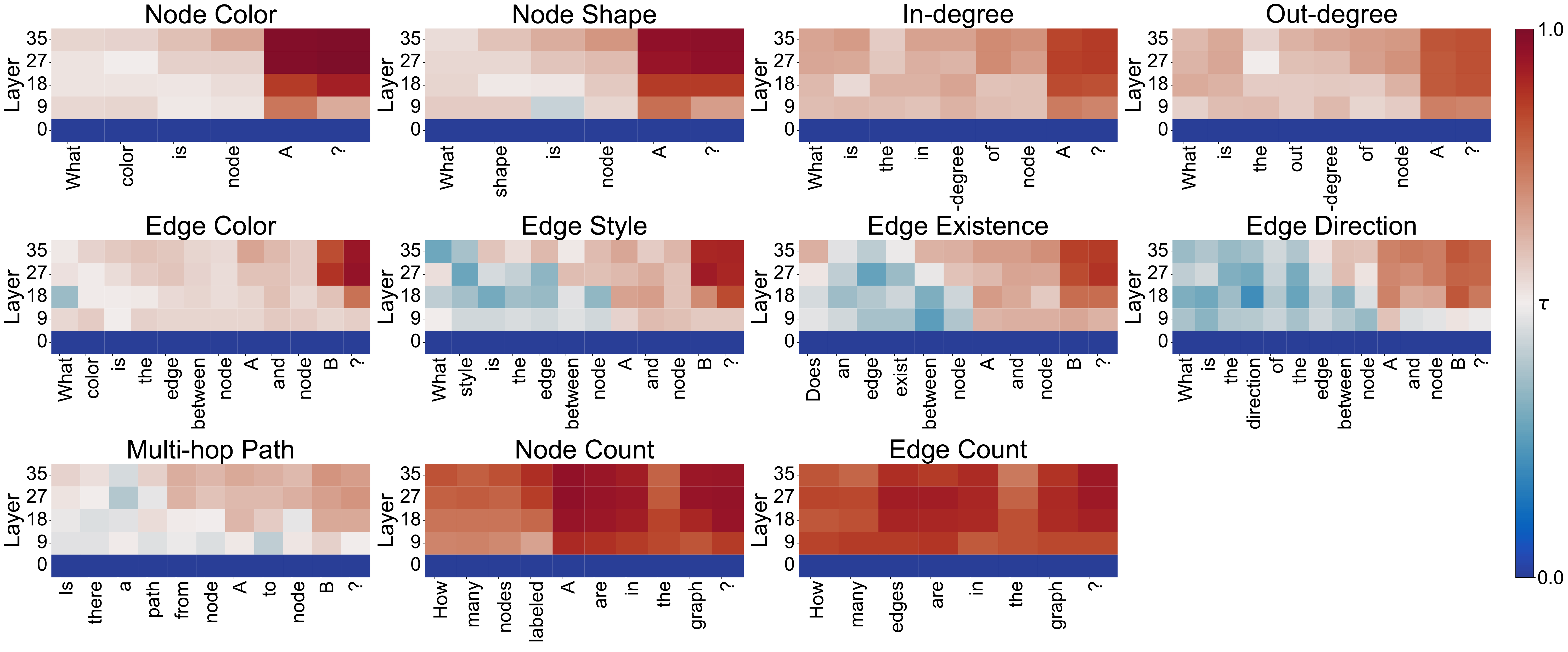}
    \caption{
        Probing results for text tokens in Qwen3-VL 8B (FT w/L).
        The x-axis shows token position in the question, and the y-axis shows layer position.
    }
    \label{fig:text_probing_qwen3_ftL}
\end{figure*}

\begin{figure}[t]
    \centering
    \includegraphics[width=\linewidth]{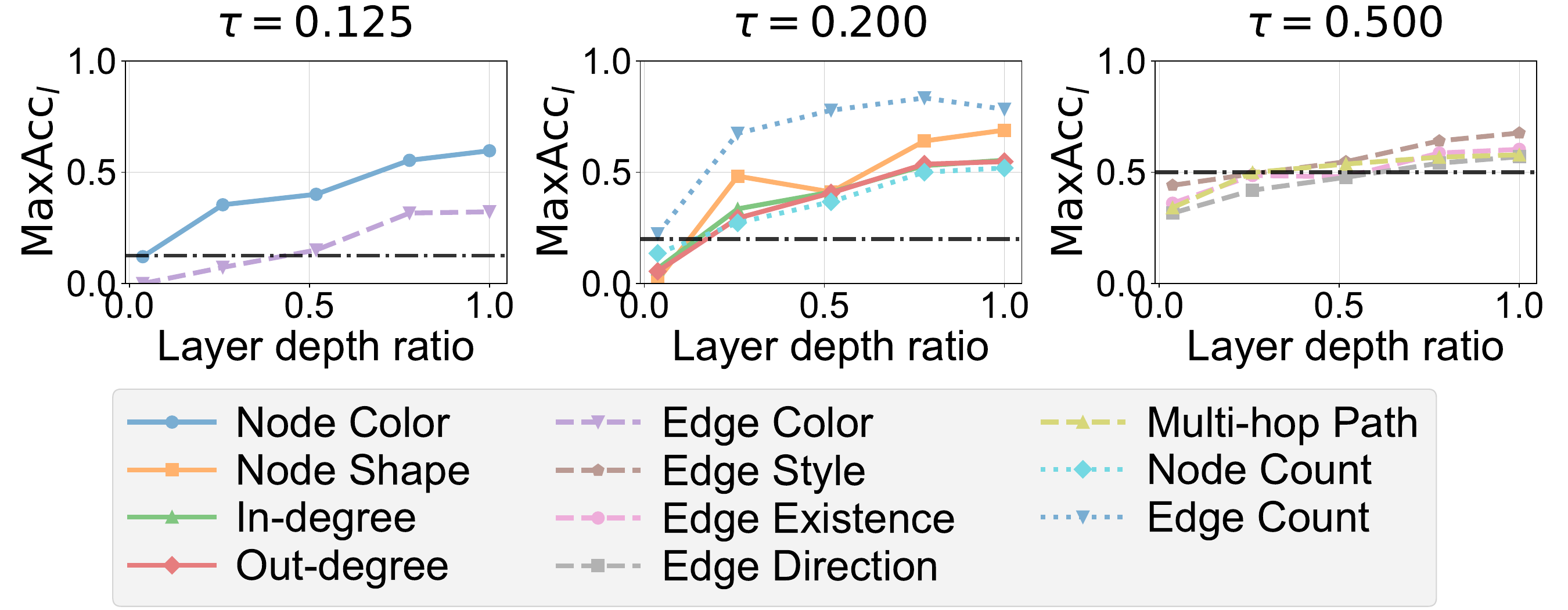}
    \caption{
        $\mathrm{MaxAcc}_{l}$ per layer in the vision encoder of Qwen3-VL 8B (FT w/VL).
        Axes and the threshold line follow Figure~\ref{fig:layerwise_image_probing_qwen3_vision_encoder}.
        Trends were largely unchanged from the pre-trained model (Figure~\ref{fig:layerwise_image_probing_qwen3_vision_encoder}).
    }
    \label{fig:layerwise_image_probing_qwen3_ftVL_vision_encoder}
\end{figure}

\begin{figure}[t]
    \centering
    \includegraphics[width=\linewidth]{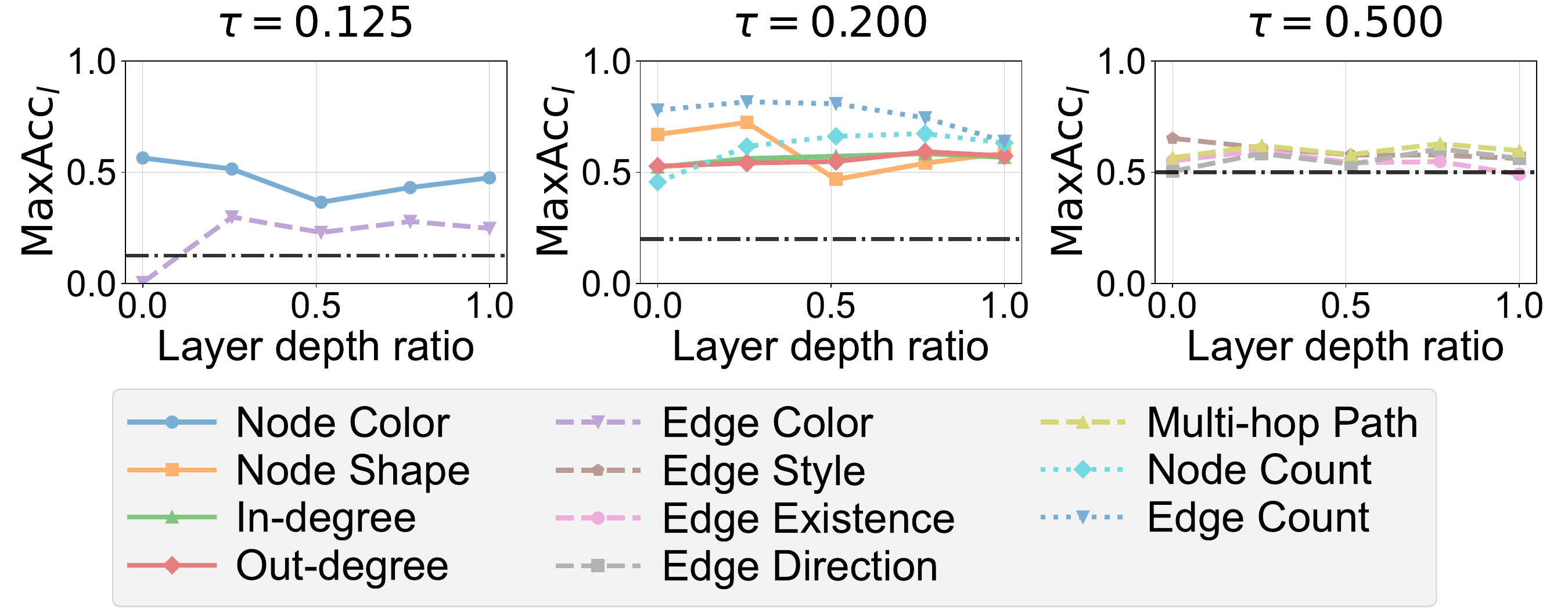}
    \caption{
        $\mathrm{MaxAcc}_{l}$ per layer in the language model of Qwen3-VL 8B (FT w/VL).
        Axes and the threshold line follow Figure~\ref{fig:layerwise_image_probing_qwen3_language_model}.
        Trends were largely unchanged from the pre-trained model (Figure~\ref{fig:layerwise_image_probing_qwen3_language_model}).
    }
    \label{fig:layerwise_image_probing_qwen3_ftVL_language_model}
\end{figure}

\begin{figure*}[t]
    \centering
    \includegraphics[width=\linewidth]{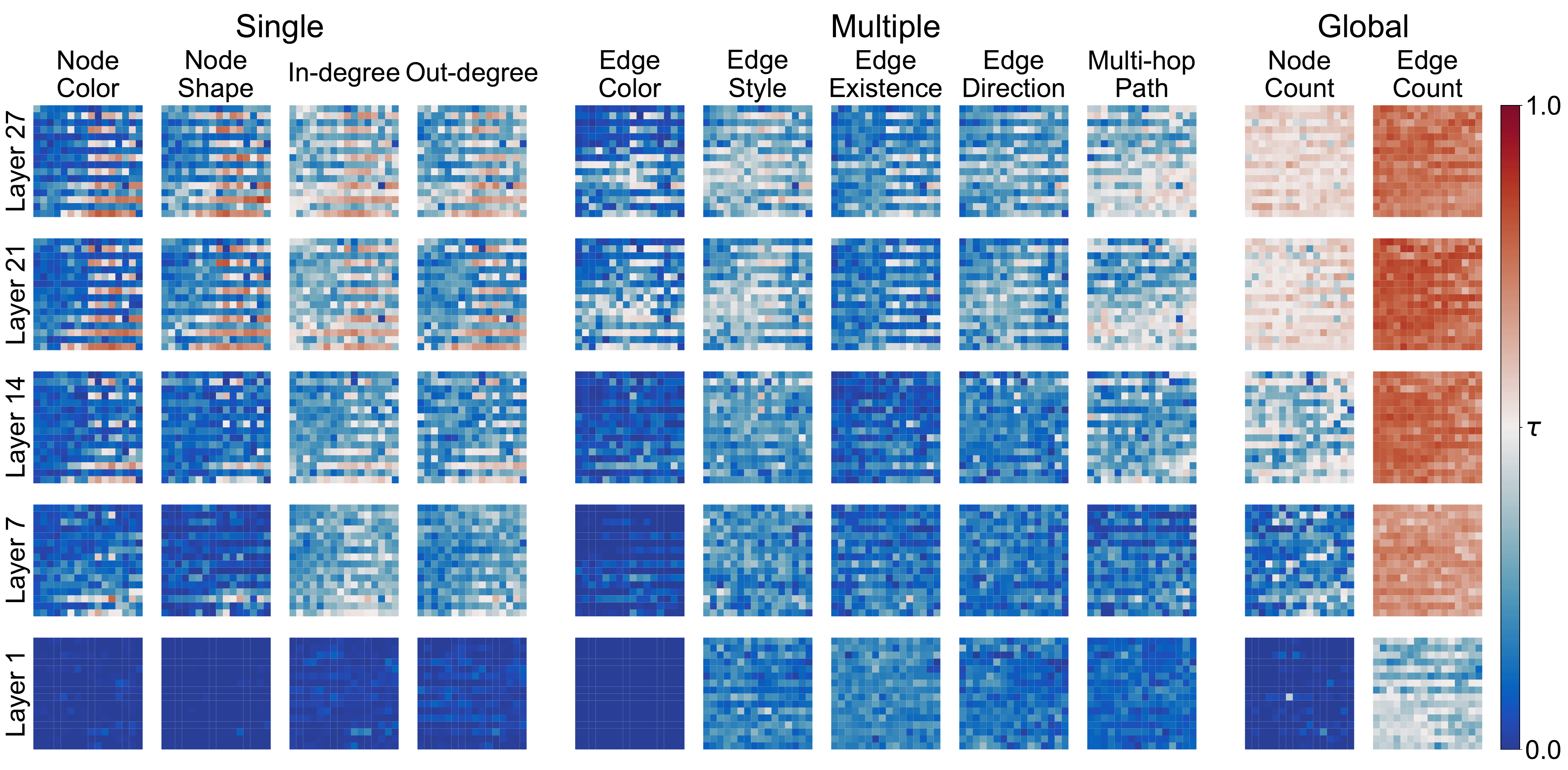}
    \caption{
        Accuracy for each hidden state in the vision encoder of Qwen3-VL 8B (FT w/VL).
        Each heatmap shows the accuracy of individual hidden states for a specific layer and aspect.
    }
    \label{fig:patchwise_image_probing_qwen3_ftVL_vision_encoder}
\end{figure*}

\begin{figure*}[t]
    \centering
    \includegraphics[width=\linewidth]{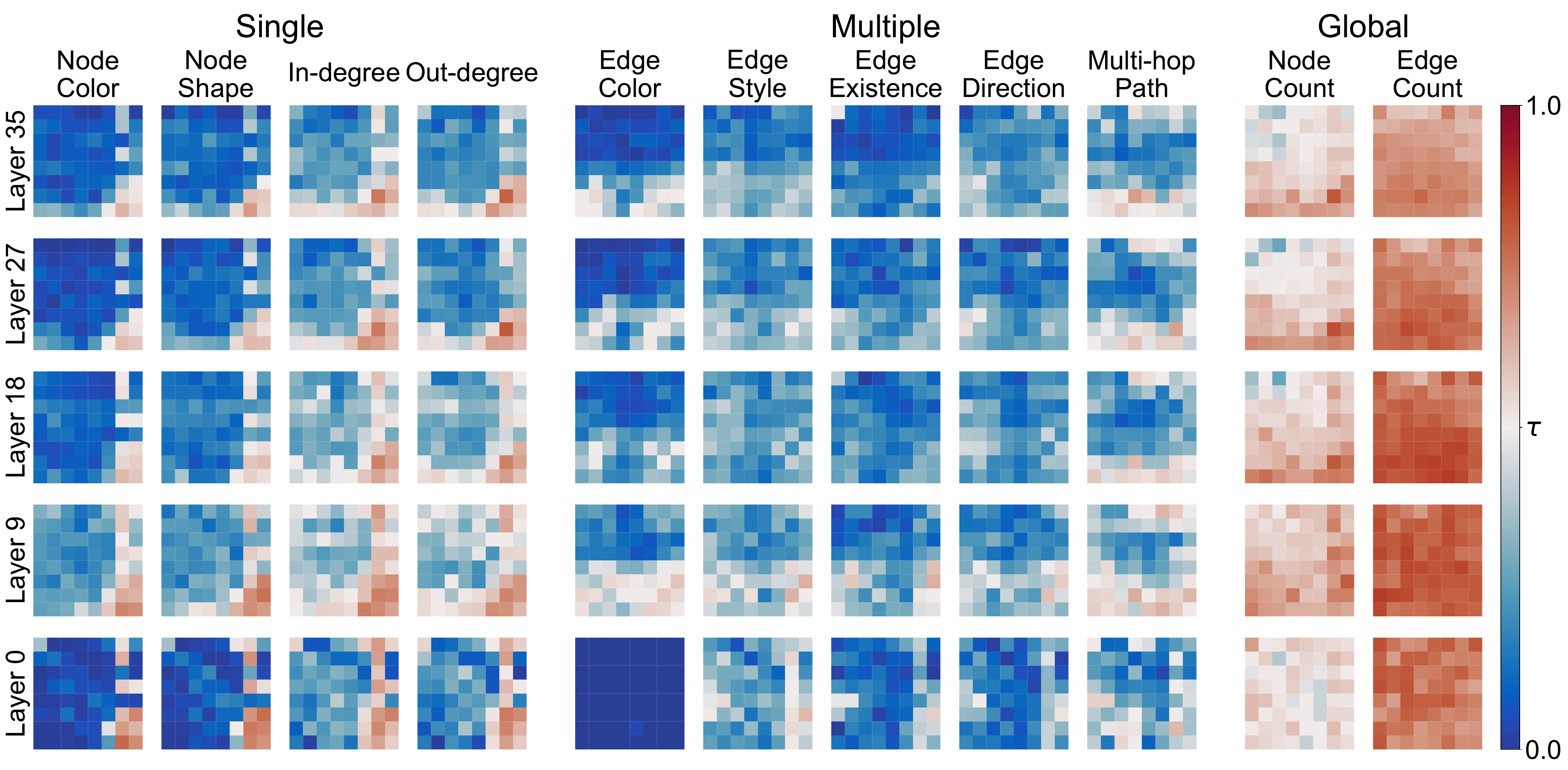}
    \caption{
        Accuracy for each hidden state in the image-input part of the Qwen3-VL 8B language model (FT w/VL).
        Each heatmap shows the accuracy of individual hidden states for a specific layer and aspect.
    }
    \label{fig:patchwise_image_probing_qwen3_ftVL_language_model}
\end{figure*}

\begin{figure*}[t]
    \centering
    \includegraphics[width=\linewidth]{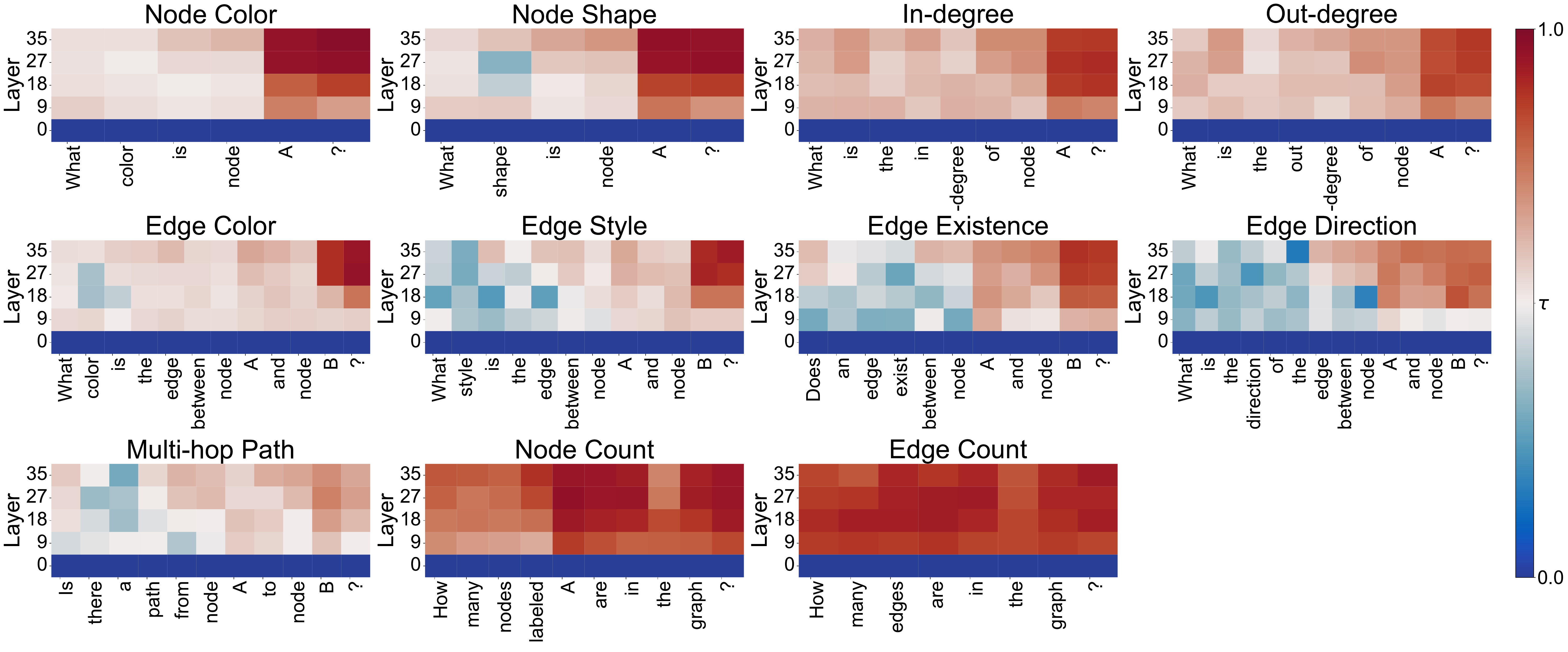}
    \caption{
        Probing results for text tokens in Qwen3-VL 8B (FT w/VL).
        The x-axis shows token position in the question, and the y-axis shows layer position.
    }
    \label{fig:text_probing_qwen3_ftVL}
\end{figure*}

We apply the same probing analysis as in \cref{sec:probing} to the fine-tuned models and compare the results with those of the pre-trained model reported in \cref{subsec:probing_results}.
Since FT w/L does not update the vision encoder, its vision encoder representations are identical to those of the pre-trained model; we therefore omit the corresponding figures and report only the language model results for this configuration.
In both configurations, the layer-wise trends and the spatial distribution of high-accuracy patches remained largely unchanged compared with the pre-trained model (Figures~\ref{fig:layerwise_image_probing_qwen3_ftL_language_model}--\ref{fig:text_probing_qwen3_ftVL}).
This suggests that, despite the improvement in VQA accuracy (Appendix~\ref{subsec:appendix_finetuning_vqa}), the linear decodability measured by our probes remained largely unchanged after fine-tuning with LoRA.

This gap between the improved VQA accuracy and the unchanged linear decodability suggests that much of the required signal was already present in the pre-trained model.
That is, fine-tuning may contribute mainly by making signals that already exist internally usable at inference time, since we would expect a corresponding change in linear decodability if the improvement came mainly from acquiring new information.
This suggests that training the language model to better exploit signals that are already present internally is a promising direction for improving diagram understanding.
It remains possible, however, that probing simply fails to capture the changes.
If LoRA updates primarily adjust nonlinear transformations or later decoding stages rather than the linear subspace, or if the low rank (rank 4) is insufficient to alter the global structure of the representation space, the representations may change without this being visible to probing.
Disentangling these factors requires further investigation, which we leave as future work.

\section{Computational Resources}
We used NVIDIA RTX 6000 Ada, RTX PRO 6000 Blackwell Max-Q, and H100 GPUs in this study.

\section{Usage of AI Assistants}
For writing this paper and developing the experimental code, we used AI assistants (e.g., ChatGPT and GitHub Copilot).
Their use was limited to tasks such as code completion, translation, text editing, and table formatting, and all technical content is based solely on the authors' ideas.

\end{document}